\documentclass[10pt,twocolumn,letterpaper]{article}

\usepackage{cvpr}
\usepackage{times}
\usepackage{epsfig}
\usepackage{graphicx}
\usepackage{amsmath}
\usepackage{amssymb}

\usepackage[USenglish]{babel}
\usepackage{subcaption}
\usepackage[skip=2pt,font=small]{caption}
\usepackage{booktabs}
\usepackage{multirow}
\usepackage{adjustbox}
\usepackage{float}
\usepackage{enumitem}
\usepackage{siunitx}
\usepackage{comment}
\usepackage{chngcntr}

\usepackage[breaklinks=true,bookmarks=false]{hyperref}

\cvprfinalcopy 

\def\cvprPaperID{218} 

\ifcvprfinal\fi

\graphicspath{{imgs/}}

\begin{document}
\title{Benchmarking the Robustness of Semantic Segmentation Models\vspace{-0.2cm}}

\author{Christoph Kamann and Carsten Rother\\
Visual Learning Lab\\
Heidelberg University (HCI/IWR)\\
{\tt\small\url{http://vislearn.de}}\vspace{-0.2cm}
}

\maketitle
\thispagestyle{empty}
\begin{abstract}
When designing a semantic segmentation module for a practical application, such as autonomous driving, it is crucial to understand the robustness of the module with respect to a wide range of image corruptions. While there are recent robustness studies for full-image classification, we are the first to present an exhaustive study for semantic segmentation, based on the state-of-the-art model DeepLabv3$+$. To increase the realism of our study, we utilize almost 400,000 images generated from Cityscapes, PASCAL VOC 2012, and ADE20K.
Based on the benchmark study, we gain several new insights. 
Firstly, contrary to full-image classification, model robustness increases with model performance, in most cases. 
Secondly, some architecture properties affect robustness significantly, such as a Dense Prediction Cell, which was designed to maximize performance on clean data only. 

\vspace{-0.5cm}
\end{abstract}
	
\section{Introduction}
\begin{figure*}[t]
	\begin{subfigure}[t!]{0.5\linewidth}
		\centering
		\includegraphics[width=0.70\linewidth]{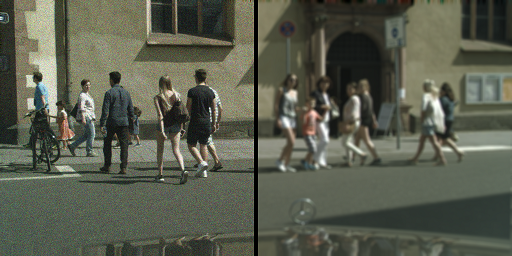}
		\caption{Corrupted validation image (left: noise, right: blur)}
	\end{subfigure}
	~ 
	\begin{subfigure}[t!]{0.5\linewidth}
		\centering
		\includegraphics[width=0.70\linewidth]{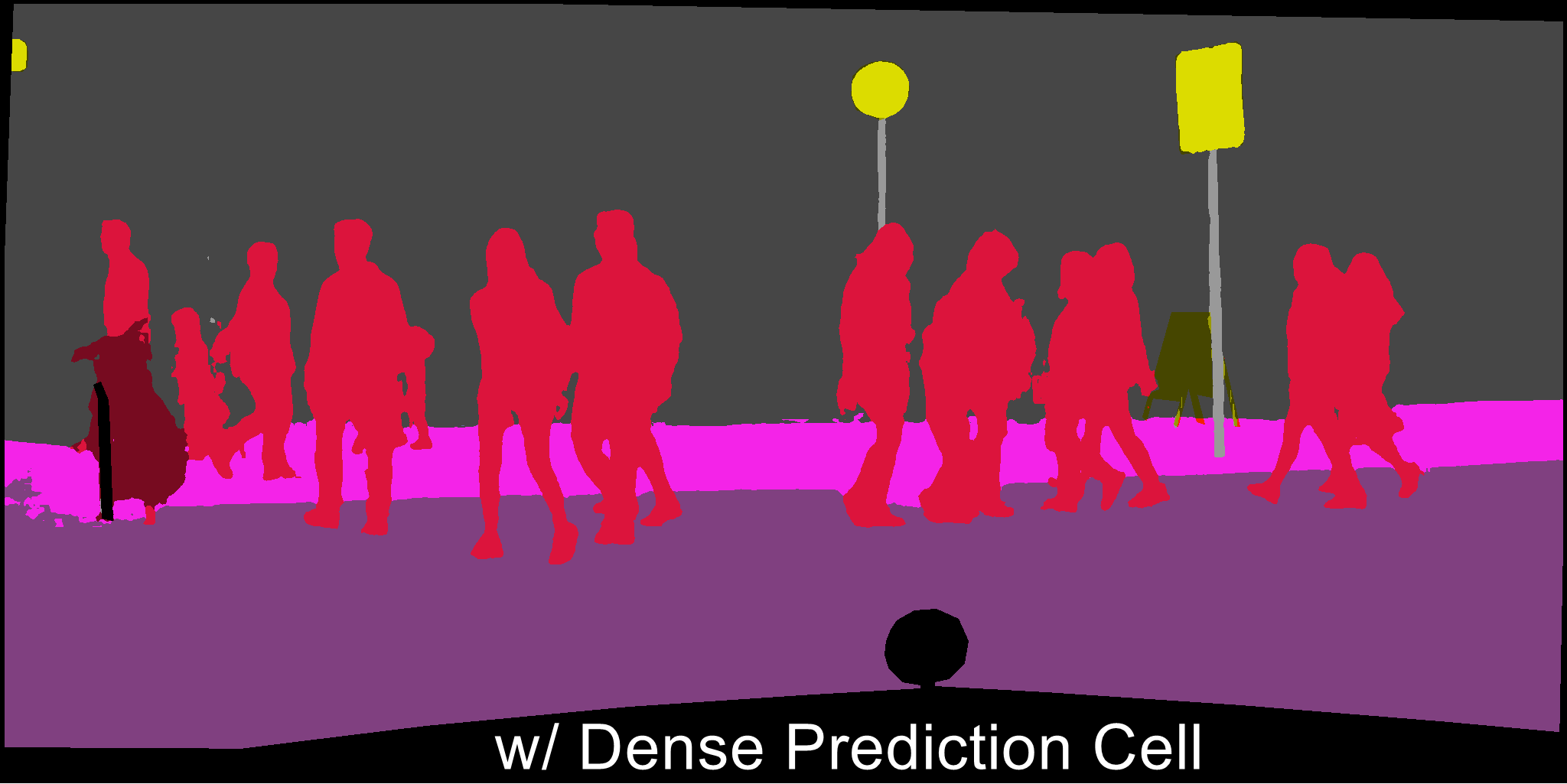}
		\caption{Prediction of best-performing architecture on clean image}
	\end{subfigure}
	
	\begin{subfigure}[t!]{0.5\linewidth}
		\centering
		\includegraphics[width=0.70\linewidth]{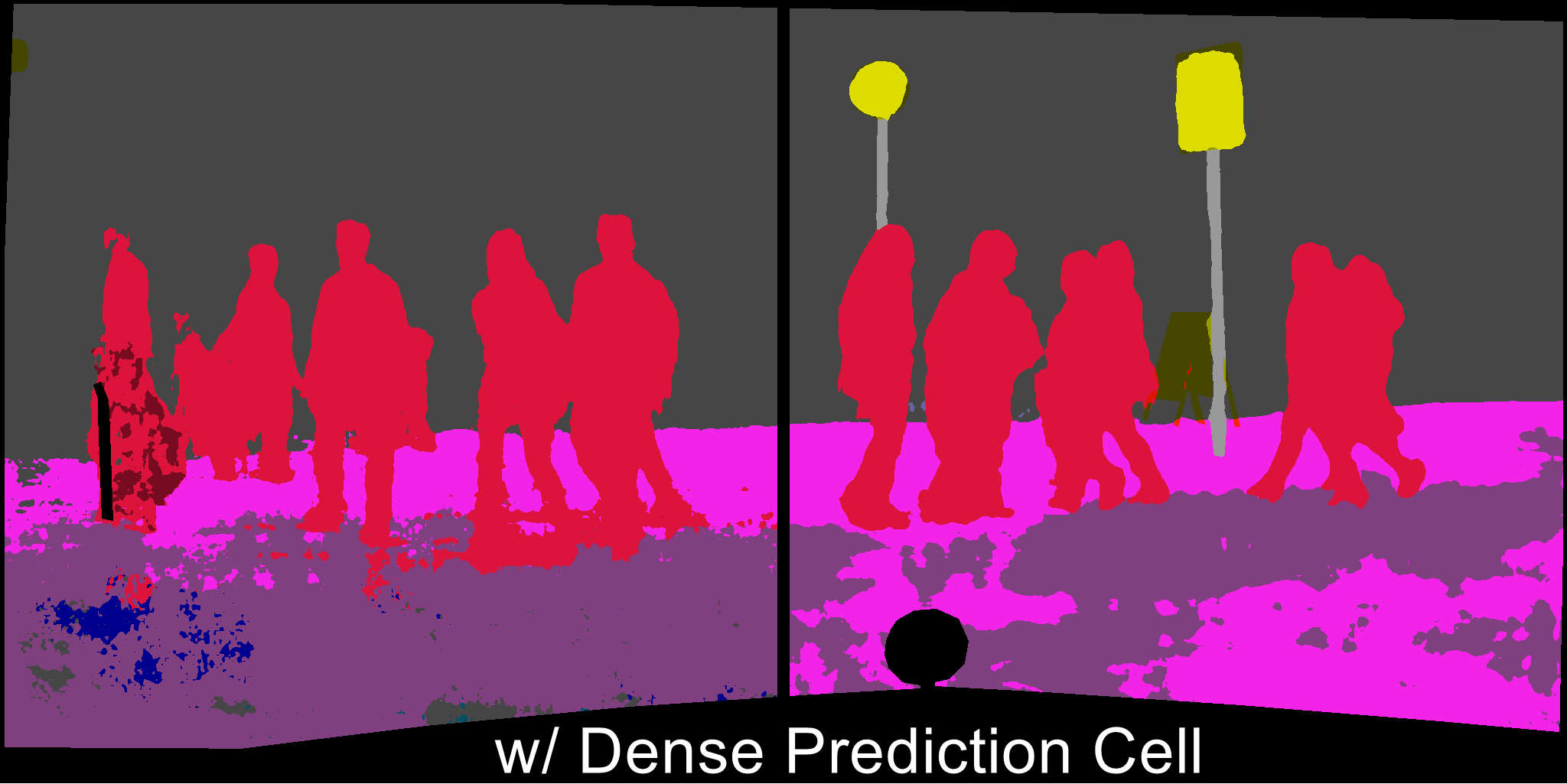}
		\caption{Prediction of best-performing architecture on corrupted image}
	\end{subfigure}
	~ 	
	\begin{subfigure}[t!]{0.5\linewidth}
		\centering
		\includegraphics[width=0.70\linewidth]{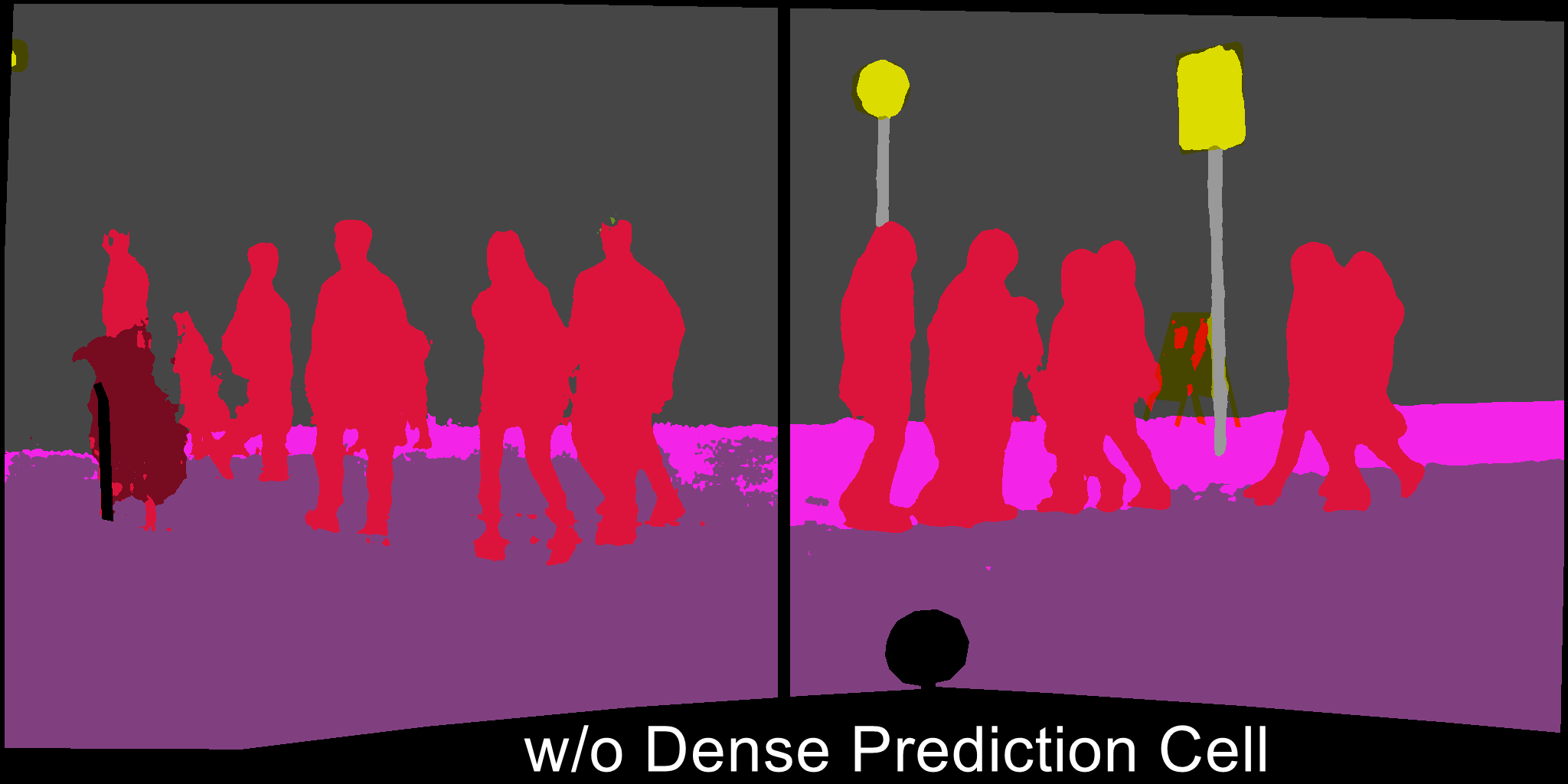}
		\caption{Prediction of ablated architecture on corrupted image}
	\end{subfigure}
	\caption{Results of our ablation study. 
	Here we train the state-of-the-art semantic segmentation model DeepLabv3$+$ on clean Cityscapes data and test it on corrupted data. 
	(a) A validation image from Cityscapes,	where the left-hand side is corrupted by {\it shot noise} and the right-hand side by {\it defocus blur}. 
	(b) Prediction of the best-performing model-variant on the corresponding clean image. 
	(c) Prediction of the same architecture on the corrupted image (a). 
	(d) Prediction of an ablated architecture on the corrupted image (a). 
	We clearly see that prediction (d) is superior to (c), hence the corresponding model is more robust with respect to this image corruption. 
	We present a study of various architectural choices and various image corruptions for the three datasets Cityscapes, PASCAL VOC 2012, and ADE20K.\vspace{-0.5cm}}
	\label{fig:Fig1}	
\end{figure*}

In recent years, Deep Convolutional Neural Networks (DCNN) have set the state-of-the-art on a broad range of computer vision tasks~\cite{krizhevsky2012imagenet,he_deep_2016,simonyan_very_2015,szegedy_going_2015,lecun_gradient-based_1998,redmon_you_2016,chen_semantic_2015,goodfellow_deep_2016,he_delving_2015,lecun_deep_2015}. 
The performance of DCNN models is generally measured using benchmarks of publicly available datasets, which often consist of clean and post-processed images~\cite{cordts_cityscapes_2016,everingham_pascal_2010}. 
However, it has been shown that model performance is prone to image corruptions~\cite{zhou_classification_2017,vasiljevic_examining_2016,hendrycks_benchmarking_2019,geirhos_generalisation_2018,dodge_understanding_2016,ford_adversarial_2019,azulay_why_2018}, 
especially image noise decreases the performance significantly.
\\
\indent 
Image quality depends on environmental factors such as illumination and weather conditions, ambient temperature, and camera motion since they directly affect the optical and electrical properties of a camera. 
Image quality is also affected by optical aberrations of the camera lenses, causing, \eg, image blur.
Thus, in safety-critical applications, such as autonomous driving, models must be robust towards such inherently present image corruptions~\cite{hasirlioglu_test_2016,kamann_test_2017,janai_computer_2017}.
\\
\indent 
In this work, we present an extensive evaluation of the robustness of semantic segmentation models towards a broad range of real-world image corruptions. 
Here, the term {\it robustness} refers to training a model on clean data and then validating it on corrupted data. We choose the task of semantic image segmentation for two reasons. 
Firstly, image segmentation is often applied in safety-critical applications, where robustness is essential. Secondly, a rigorous evaluation for real-world image corruptions has, in recent years, only been conducted for full-image classification and object detection, \eg, most recently~\cite{geirhos_generalisation_2018,hendrycks_benchmarking_2019,michaelis_benchmarking_2019}.
\\
\indent 
When conducting an evaluation of semantic segmentation models, there are, in general, different choices such as: i) comparing different architectures, or ii) conducting a detailed ablation study of a state-of-the-art architecture.
In contrast to \cite{geirhos_generalisation_2018,hendrycks_benchmarking_2019}, which focused on aspect i), we perform both options. We believe that an ablation study (option ii) is important since knowledge about architectural choices are likely helpful when designing a practical system, where types of image corruptions are known beforehand. 
For example,~\cite{geirhos_generalisation_2018} showed that ResNet-152~\cite{he_deep_2016} is more robust to image noise than GoogLeNet~\cite{szegedy_going_2015}. Is the latter architecture more prone to noise due to missing skip-connections, shallower architecture, or other architectural design choices? 
When the overarching goal is to develop robust DCNN models, we believe that it is important to learn about the robustness capabilities of architectural properties.
\\
\indent 
We conduct our study on three popular datasets: Cityscapes~\cite{cordts_cityscapes_2016}, PASCAL VOC 2012~\cite{everingham_pascal_2010}, and ADE20K~\cite{zhou_scene_2017,zhou_semantic_2016}. To generate a wide-range of image corruptions, we utilize the image transformations presented by Hendrycks \etal ~\cite{hendrycks_benchmarking_2019}. While they give a great selection of image transformations, the level of realism is rather lacking, in our view. Hence we augment their image transformations by additional ones, in particular, intensity-dependent camera noise, PSF blur, and geometric distortions. In total, we employ 19 different image corruptions from the categories of blur, noise, weather, digital, and geometric distortion. We are thus able to validate each DCNN model on almost 400,000 images.
\\
\indent 
We use the state-of-the-art DeepLabv3$+$ architecture~\cite{chen_encoder-decoder_2018} with multiple network backbones as reference and consider many ablations of it. 
Based on our evaluation, we are able to conclude two main findings: 
1) Contrary to the task of full-image classification, we observe that the robustness of semantic segmentation models of DeepLabv3$+$ increases often with model performance. 
2) Architectural properties can affect the robustness of a model significantly. 
Our results show that atrous (i.e., dilated) convolutions and long-range link naturally aid the robustness against many types of image corruptions. 
However, an architecture with a Dense Prediction Cell~\cite{chen_searching_2018}, which was designed to maximize performance on clean data, hampers the performance for corrupted images significantly (see Fig.~\ref{fig:Fig1}). 
\\ \indent In summary, we give the following contributions:
\begin{itemize}[noitemsep,nolistsep]
    \item[--] We benchmark the robustness of many architectural properties of the state-of-the-art semantic segmentation model DeepLabv3$+$ for a wide range of real-world image corruptions. We utilize almost 400,000 images generated from the Cityscapes dataset, PASCAL VOC 2012, and ADE20K. 
    \item[--] Besides DeepLabv3$+$, we have also benchmarked a wealth of other semantic segmentation models.
    \item[--] We develop a more realistic noise model than previous approaches.
    \item[--] Based on the benchmark study, we have several new insights: 
    1) contrary to full-image classification, model robustness of DeepLabv3$+$ increases with model performance, in most cases; 
    2) Some architecture properties affect robustness significantly. 
\end{itemize}

\section{Related Work}
Robustness studies~\cite{zhou_classification_2017,vasiljevic_examining_2016,hendrycks_benchmarking_2019,geirhos_generalisation_2018,dodge_study_2017,dodge_understanding_2016,neyshabur_exploring_2017,michaelis_benchmarking_2019} and robustness enhancement~\cite{yim_enhancing_2017,zheng_improving_2016,goodfellow_deep_2016,henriques_warped_2017,borkar_deepcorrect:_2017,sun_feature_2018,geirhos_imagenet-trained_2019} of DCNN architectures~\cite{krizhevsky2012imagenet,szegedy_going_2015,simonyan_very_2015,sermanet_overfeat:_2014,long_fully_2015,chen_semantic_2015,chen_deeplab:_2017,chen_encoder-decoder_2018,chen_rethinking_2017,mukherjee_visual_2018,bahnsen_learning_2019} have been addressed in various benchmarks~\cite{everingham_pascal_2010,cordts_cityscapes_2016,deng_imagenet:_2009}. 
Recent work also dealt with evaluating and increasing the robustness of CNNs against various weather conditions~\cite{sakaridis_semantic_2018,volk_towards_2019,dai_dark_2018,chen_domain_2018,sakaridis_guided_2019}.
Vasiljevic \etal~\cite{vasiljevic_examining_2016} examined the impact of blur on full-image classification and semantic segmentation using VGG-16~\cite{simonyan_very_2015}.
Model performance decreases with an increased degree of blur for both tasks. 
We also focus in this work on semantic segmentation but evaluate on a much wider range of real-world image corruptions.
\\ \indent Geirhos \etal~\cite{geirhos_generalisation_2018} compared the generalization capabilities of humans and Deep Neural Networks (DNNs).
The ImageNet dataset~\cite{deng_imagenet:_2009} is modified in terms of color variations, noise, blur, and rotation. 
\\ \indent Hendrycks \etal~\cite{hendrycks_benchmarking_2019} introduce the ``ImageNet-C dataset''. 
The authors corrupted the ImageNet dataset by common image corruptions.
Although the absolute performance scores increase from AlexNet~\cite{krizhevsky2012imagenet} to ResNet~\cite{he_deep_2016}, the robustness of the respective models does barely change. 
They further show that Multigrid and DenseNet architectures~\cite{ke_multigrid_2017,huang_densely_2017} are less prone to noise corruption than ResNet architectures. 
In this work, we use most of the proposed image transformations and apply them to the Cityscapes dataset, PASCAL VOC 2012, and ADE20K~\cite{cordts_cityscapes_2016,everingham_pascal_2010,zhou_scene_2017,zhou_semantic_2016}.
\\ \indent Geirhos \etal \cite{geirhos_imagenet-trained_2019} showed that humans and DNNs classify images with different strategies. Unlike humans, DNNs trained on ImageNet seem to rely more on local texture instead of global object shape. 
The authors then show that model robustness \wrt image corruptions increases, when CNNs rely more on object shape than on object texture. 
\\ \indent Robustness of models with respect to adversarial examples is an active field of research~\cite{huang_safety_2017,boopathy_cnn-cert:_2019,cisse_parseval_2017,gu_towards_2014,carlini_towards_2017,metzen_detecting_2017,carlini_adversarial_2017}. 
Arnab \etal~\cite{arnab_robustness_2018} evaluate the robustness of semantic segmentation models for adversarial attacks of a wide variety of network architectures (e.g.~\cite{zhao_pyramid_2017,badrinarayanan_segnet:_2017,paszke_enet:_2016,zhao_icnet_2018,yu_multi-scale_2016}).  
In this work, we adopt a similar evaluation procedure, but we do not focus on the robustness \wrt adversarial attacks, which are typically not realistic, but rather on physically realistic image corruptions. 
We further rate robustness \wrt many architectural properties instead of solely comparing CNN architectures. 
Our approach modifies a single property per model at a time, which allows for an accurate evaluation. 
\\ \indent Ford \etal~\cite{ford_adversarial_2019} connect adversarial robustness and robustness with respect to image corruption of Gaussian noise. 
The authors showed that training procedures that increase adversarial robustness also improve robustness with respect to many image corruptions.
\section{Image Corruption Models}
We evaluate the robustness of semantic segmentation models towards a broad range of image corruptions. Besides using image corruptions from the ImageNet-C dataset, we propose new and more realistic image corruptions. 
\subsection{ImageNet-C}
We employ many image corruptions from the ImageNet-C dataset~\cite{hendrycks_benchmarking_2019}.
These consist of several types of {\it blur:} motion, defocus, frosted glass and Gaussian; {\it Noise:} Gaussian, impulse, shot and speckle; {\it Weather:} snow, spatter, fog, and frost; and {\it Digital:} brightness, contrast, and JPEG compression. 
Each corruption is parameterized with five severity levels. We refer to the supplemental material for an illustration of these corruptions.
\subsection{Additional Image Corruptions}
\label{subsec:realistic_corruptions}
\begin{figure}[t]
	\begin{subfigure}[t!]{0.24\linewidth}
		\centering
		\includegraphics[width=0.95\linewidth]{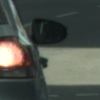}
		\caption{Clean image}
	\end{subfigure}%
	~
	\begin{subfigure}[t!]{0.24\linewidth}
		\centering
		\includegraphics[width=0.95\linewidth]{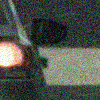}
		\caption{Gaussian}
	\end{subfigure}%
	~
	\begin{subfigure}[t!]{0.24\linewidth}
		\centering
		\includegraphics[width=0.95\linewidth]{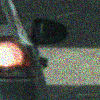}
		\caption{Shot}
	\end{subfigure}%
	~
	\begin{subfigure}[t!]{0.24\linewidth}
		\centering
		\includegraphics[width=0.95\linewidth]{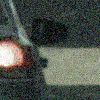}
		\caption{Proposed}
	\end{subfigure}%
	\caption{A crop of a validation image from Cityscapes corrupted by various noise models. (a) Clean image. (b) Gaussian noise. (c) Shot noise. (d) Our proposed noise model. The amount of noise is high in regions with low pixel intensity.
	\vspace{-0.5cm}}
	\label{fig:rgbnoise}
\end{figure}

\textbf{Intensity-Dependent Noise Model.} 
DCNNs are prone to noise. 
Previous noise models are often simplistic, \eg, images are evenly distorted with Gaussian noise. However, \textit{real} image noise significantly differs from the noise generated by these simple models.
Real image noise is a combination of multiple types of noise (\eg, photon noise, kTC noise, dark current noise as described in~\cite{healey_radiometric_1994,young_fundamentals_1998,lukas_digital_2006,liu_automatic_2008}).
\\ \indent We propose a noise model that incorporates commonly observable behavior of cameras. 
Our noise model consists of two noise components:
i) a chrominance and luminance noise component, which are both added to original pixel intensities in linear color space. 
ii) an intensity-level dependent behavior.
In accordance with image noise observed from real-world cameras, pixels with low intensities are noisier than pixels with high intensities.
Fig. \ref{fig:rgbnoise} illustrates noisy variants of a Cityscapes image-crop.  
In contrast to the other, simpler noise models, the amount of noise generated by our noise model depends clearly on pixel intensity. 
\\ \indent \textbf{PSF blur.} Every optical system of a camera exhibits aberrations, which mostly result in image blur.
A point-spread-function (PSF) aggregates all optical aberrations that result in image blur~\cite{joshi_psf_2008}. 
We denote this type of corruption as \textit{PSF blur}.
Unlike simple blur models, such as Gaussian blur, real-world PSF functions are spatially varying. We corrupt the Cityscapes dataset with three different PSF functions that we have generated with the optical design program {\it Zemax}, for which the amount of blur increases with a larger distance to the image center.
\\ \indent \textbf{Geometric distortion.} Every camera lens exhibits geometric distortions~\cite{fitzgibbon_simultaneous_2001}. We applied several radially-symmetric barrel distortions~\cite{willson_modeling_1994} as a polynomial of grade 4~\cite{shah_intrinsic_1996} to both the RGB-image and respective ground truth.

\section{Models}
We employ DeepLabv3$+$~\cite{chen_encoder-decoder_2018} as the reference architecture. 
We chose DeepLabv3$+$ for several reasons. 
It supports numerous network backbones, ranging from novel state-of-art models (\eg, modified aligned Xception~\cite{chollet_xception:_2017,chen_encoder-decoder_2018,qi_deformable_2017}, denoted by \textit{Xception}) and established ones (\eg, ResNets~\cite{he_deep_2016}). 
For semantic segmentation, DeepLabv3$+$ utilizes popular architectural properties, making it a highly suitable candidate for an ablation study.
Please note that the range of network backbones, offered by DeepLabv3$+$, represents different execution times since different applications have different demands.
\\ \indent Besides DeepLabv3$+$, we have also benchmarked a wealth of other semantic segmentation models, such as FCN8s~\cite{long_fully_2015}, VGG-16~\cite{simonyan_very_2015}, ICNet~\cite{zhao_icnet_2018}, DilatedNet~\cite{yu_multi-scale_2016}, ResNet-38~\cite{wu_wider_2019}, PSPNet~\cite{zhao_pyramid_2017}, and the recent Gated-ShapeCNN (GSCNN)~\cite{takikawa_gated-scnn:_2019}.
\subsection{DeepLabv3$+$}
\label{subsec:deeplab}
\begin{figure}
	\centering
	\includegraphics[width=0.60\linewidth]{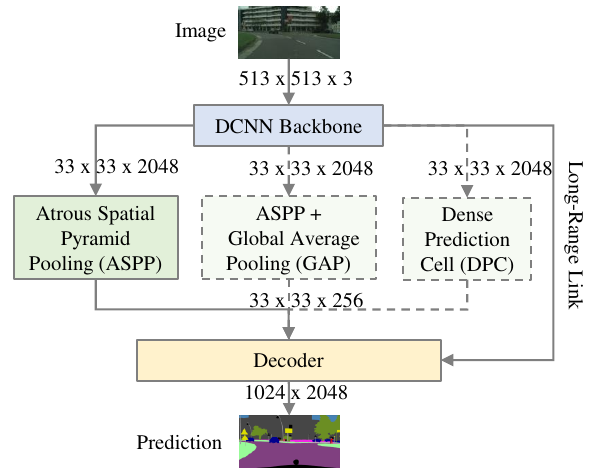}
	\caption{Building blocks of DeepLabv3$+$. Input images are firstly processed by a network backbone, containing atrous convolutions. The backbone output is further processed by a multi-scale processing module (ASPP or DPC). A long-range link concatenates early features of the network backbone with encoder output. Finally, the decoder outputs estimates of semantic labels. Our reference model is shown by regular arrows (\ie, without DPC and GAP). The dimension of activation volumes is shown after each block.\vspace{-0.5cm}}
	\label{fig:deeplab}
\end{figure}

Fig. \ref{fig:deeplab} illustrates important elements of the DeepLabv3$+$ architecture. A network backbone (ResNet, Xception or MobileNet-V2) processes an input image~\cite{he_deep_2016,sandler_mobilenetv2:_2018,howard_mobilenets:_2017}.
Its output is subsequently processed by a multi-scale processing module, extracting dense feature maps. 
This module is either Dense Prediction Cell~\cite{chen_searching_2018} (DPC) or Atrous Spatial Pyramid Pooling (ASPP, with or without global average pooling (GAP)). 
We consider the variant with ASPP and without GAP as reference architecture.
A long-range link concatenates early features from the network backbone with features extracted by the respective multi-scale processing module.
Finally, the decoder outputs estimates of the semantic labels. 
\\ \indent \textbf{Atrous convolution.} Atrous (\ie, dilated) convolution~\cite{chen_deeplab:_2017,holschneider_real-time_1989,papandreou_modeling_2015} is a type of convolution that integrates spacing between kernel parameters and thus increases the kernel field of view. DeepLabv3$+$ incorporates atrous convolutions in the network backbone. 
\\ \indent \textbf{Atrous Spatial Pyramid Pooling.} To extract features at different scales, several semantic segmentation architectures~\cite{chen_deeplab:_2017,chen_semantic_2015,zhao_pyramid_2017} perform Spatial Pyramid Pooling~\cite{he_spatial_2014,grauman_pyramid_2005,lazebnik_beyond_2006}. DeepLabv3$+$ applies \textit{Atrous} Spatial Pyramid Pooling (ASPP), where three atrous convolutions with large atrous rates (6,12 and 18) process the DCNN output. 
\\ \indent \textbf{Dense Prediction Cell.}~\cite{chen_searching_2018} is an efficient multi-scale architecture for dense image prediction, constituting an alternative to ASPP. It is the result of a neural-architecture-search with the objective to maximize the performance for clean images. In this work, we analyze whether this objective leads to overfitting.
\\ \indent \textbf{Long-Range link.} 
A long-range link concatenates early features of the encoder with features extracted by the respective multi-scale processing module~\cite{hariharan_hypercolumns_2015}. In more detail, for Xception (MobileNet-V2) based models, the long-range link connects the output of the second or the third Xception block (inverted residual block) with ASPP or DPC output. Regarding ResNet architectures, the long-range link connects the output of the second residual block with the ASPP or DPC output.
\\ \indent \textbf{Global Average Pooling.} 
A global average pooling (GAP) layer~\cite{lin_network_2014} averages the feature maps of an activation volume. DeepLabv3$+$ incorporates GAP in parallel to the ASPP.
	
\subsection{Architectural Ablations}
\label{subsec:ablations}
In the next section, we evaluate various ablations of the DeepLabv3$+$ reference architecture. In detail, we remove atrous convolutions (AC) from the network backbone by transforming them into regular convolutions. 
We denote this ablation in the remaining sections as w\textbackslash o AC. We further removed the long-range link (LRL, \ie, w\textbackslash o LRL) and Atrous Spatial Pyramid Pooling (ASPP) module (w\textbackslash o ASPP). The removal of ASPP is additionally replaced by Dense Prediction Cell (DPC) and denoted as w\textbackslash o ASPP$+$w\textbackslash DPC. We also examined the effect of global average pooling (w\textbackslash GAP).

\section{Experiments}
We present the experimental setup (sec.~\ref{subsec:expsetup}) and then the results of two different experiments. 
We firstly benchmark multiple neural network backbone architectures of DeepLabv3$+$ and other semantic segmentation models (sec.~\ref{subsec:benchmark}).
While this procedure gives an overview of the robustness across several architectures, no conclusions about which architectural properties affect the robustness can be drawn.
Hence, we modify multiple architectural properties of DeepLabv3$+$ (sec.~\ref{subsec:ablations}) and evaluate the robustness for re-trained ablated models \wrt image corruptions (sec.~\ref{subsec:as_cs}, ~\ref{subsec:as_pascal}, ~\ref{subsec:as_ade20k}). 
Our findings show that architectural properties can have a substantial impact on the robustness of a semantic segmentation model \wrt image corruptions. 

    \begin{table*}[h!]
	\begin{adjustbox}{width=\textwidth}
		\begin{tabular}{@{}ccccccccccccccccccccc@{}}
			\toprule
            &  & \multicolumn{5}{c}{\textbf{Blur}} & \multicolumn{5}{c}{\textbf{Noise}} & \multicolumn{4}{c}{\textbf{Digital}} & \multicolumn{4}{c}{\textbf{Weather}} &  \\ \midrule
			\multicolumn{1}{c|}{Architecture} & \multicolumn{1}{c|}{Clean} & Motion & Defocus & \begin{tabular}[c]{@{}c@{}}Frosted\\ Glass\end{tabular} & Gaussian & \multicolumn{1}{l|}{PSF} & Gaussian & Impulse & Shot & Speckle & \multicolumn{1}{l|}{Intensity} & Brightness & Contrast & Saturate & \multicolumn{1}{l|}{JPEG} & \multicolumn{1}{l}{Snow} & \multicolumn{1}{l}{Spatter} & \multicolumn{1}{l}{Fog} & \multicolumn{1}{l|}{Frost} & \begin{tabular}[c]{@{}c@{}}Geometric\\ Distortion\end{tabular} \\ \midrule
			\multicolumn{1}{l|}{MobileNet-V2} & \multicolumn{1}{c|}{72.0} & 53.5 & 49.0 & 45.3 & 49.1 & \multicolumn{1}{c|}{70.5} & 6.4 & 7.0 & 6.6 & 16.6 & \multicolumn{1}{c|}{26.9} & 51.7 & 46.7 & 32.4 & \multicolumn{1}{c|}{27.2} & 13.7 & 38.9 & 47.4 & \multicolumn{1}{c|}{17.3} & 65.5 \\			
			\multicolumn{1}{c|}{ResNet-50} & \multicolumn{1}{c|}{76.6} & 58.5 & 56.6 & 47.2 & 57.7 & \multicolumn{1}{c|}{74.8} & 6.5 & 7.2 & 10.0 & 31.1 & \multicolumn{1}{c|}{30.9} & 58.2 & 54.7 & 41.3 & \multicolumn{1}{c|}{27.4} & 12.0 & 42.0 & 55.9 & \multicolumn{1}{c|}{22.8} & 69.5 \\
			\multicolumn{1}{c|}{ResNet-101} & \multicolumn{1}{c|}{77.1} & 59.1 & 56.3 & 47.7 & 57.3 & \multicolumn{1}{c|}{75.2} & 13.2 & 13.9 & 16.3 & 36.9 & \multicolumn{1}{c|}{39.9} & 59.2 & 54.5 & 41.5 & \multicolumn{1}{c|}{37.4} & 11.9 & 47.8 & 55.1 & \multicolumn{1}{c|}{22.7} & 69.7 \\
			\multicolumn{1}{c|}{Xception-41} & \multicolumn{1}{c|}{77.8} & 61.6 & 54.9 & 51.0 & 54.7 & \multicolumn{1}{c|}{76.1} & \textbf{17.0} & \textbf{17.3} & \textbf{21.6} & \textbf{43.7} & \multicolumn{1}{c|}{48.6} & 63.6 & 56.9 & \textbf{51.7} & \multicolumn{1}{c|}{38.5} & 18.2 & 46.6 & 57.6 & \multicolumn{1}{c|}{20.6} & \textbf{73.0} \\
			\multicolumn{1}{c|}{Xception-65} & \multicolumn{1}{c|}{78.4} & 63.9 & 59.1 & \textbf{52.8} & 59.2 & \multicolumn{1}{c|}{\textbf{76.8}} & 15.0 & 10.6 & 19.8 & 42.4 & \multicolumn{1}{c|}{46.5} & 65.9 & \textbf{59.1} & 46.1 & \multicolumn{1}{c|}{31.4} & \textbf{19.3} & \textbf{50.7} & 63.6 & \multicolumn{1}{c|}{\textbf{23.8}} & 72.7 \\
			\multicolumn{1}{c|}{Xception-71} & \multicolumn{1}{c|}{\textbf{78.6}} & \textbf{64.1} & \textbf{60.9} & 52.0 & \textbf{60.4} & \multicolumn{1}{c|}{76.4} & 14.9 & 10.8 & 19.4 & 41.2 & \multicolumn{1}{c|}{\textbf{50.2}} & \textbf{68.0} & 58.7 & 47.1 & \multicolumn{1}{c|}{\textbf{40.2}} & 18.8 & 50.4 & \textbf{64.1} & \multicolumn{1}{c|}{20.2} & 71.0 \\ \bottomrule
			\multicolumn{1}{c|}{ICNet} & \multicolumn{1}{c|}{65.9} & 45.8 & 44.6 & \textbf{47.4} & 44.7 & \multicolumn{1}{c|}{65.2} & 8.4 & 8.4 & 10.6 & 27.9 & \multicolumn{1}{c|}{29.7} & 41.0 & 33.1 & 27.5 & \multicolumn{1}{c|}{\textbf{34.0}} & 6.3 & 30.5 & 27.3 & \multicolumn{1}{c|}{11.0} & 35.7 \\ 
			\multicolumn{1}{c|}{FCN8s-VGG16} & \multicolumn{1}{c|}{66.7} & 42.7 & 31.1 & 37.0 & 34.1 & \multicolumn{1}{c|}{61.4} & 6.7 & 5.7 & 7.8 & 24.9 & \multicolumn{1}{c|}{18.8} & 53.3 & 39.0 & 36.0 & \multicolumn{1}{c|}{21.2} & 11.3 & 31.6 & 37.6 & \multicolumn{1}{c|}{19.7} & 36.9 \\ 
			\multicolumn{1}{c|}{DilatedNet} & \multicolumn{1}{c|}{68.6} & 44.4 & 36.3 & 32.5 & 38.4 & \multicolumn{1}{c|}{61.1} & \textbf{15.6} & 14.0 & \textbf{18.4} & 32.7 & \multicolumn{1}{c|}{35.4} & 52.7 & 32.6 & 38.1 & \multicolumn{1}{c|}{29.1} & 12.5 & 32.3 & 34.7 & \multicolumn{1}{c|}{19.2} & 38.9 \\ 
			\multicolumn{1}{c|}{ResNet-38} & \multicolumn{1}{c|}{77.5} & 54.6 & 45.1 & 43.3 & 47.2 & \multicolumn{1}{c|}{74.9} & 13.7 & \textbf{16.0} & 18.2 & \textbf{38.3} & \multicolumn{1}{c|}{\textbf{35.9}} & 60.0 & 50.6 & 46.9 & \multicolumn{1}{c|}{14.7} & \textbf{13.5} & 45.9 & 52.9 & \multicolumn{1}{c|}{22.2} & 43.2 \\ 
			\multicolumn{1}{c|}{PSPNet} & \multicolumn{1}{c|}{78.8} & \textbf{59.8} & 53.2 & 44.4 & 53.9 & \multicolumn{1}{c|}{76.9} & 11.0 & 15.4 & 15.4 & 34.2 & \multicolumn{1}{c|}{32.4} & 60.4 & 51.8 & 30.6 & \multicolumn{1}{c|}{21.4} & 8.4 & 42.7 & 34.4 & \multicolumn{1}{c|}{16.2} & \textbf{43.4} \\ 
			\multicolumn{1}{c|}{GSCNN} & \multicolumn{1}{c|}{\textbf{80.9}} & 58.9 & \textbf{58.4} & 41.9 & \textbf{60.1} & \multicolumn{1}{c|}{\textbf{80.3}} & 5.5 & 2.6 & 6.8 & 24.7 & \multicolumn{1}{c|}{29.7} & \textbf{75.9} & \textbf{61.9} & \textbf{70.7} & \multicolumn{1}{c|}{12.0} & 12.4 & \textbf{47.3} & \textbf{67.9} & \multicolumn{1}{c|}{\textbf{32.6}} & 42.7 \\ \bottomrule
		\end{tabular}
	\end{adjustbox}
	\caption{Average mIoU for clean and corrupted variants of the Cityscapes validation set for several network backbones of the DeepLabv3$+$ architecture (\textit{top}) and non-DeepLab based models (\textit{bottom}). Every mIoU is averaged over all available severity levels, except for corruptions of category noise where only the first three (of five) severity levels are considered. Xception based network backbones are usually most robust against each corruption. Most models are robust against our realistic PSF blur. 
	Highest mIoU per corruption is bold.\vspace{-0.5cm}}
	\label{tab:benchmark_cityscapes}
\end{table*}

	\subsection{Experimental Setup}
	\label{subsec:expsetup}
	\textbf{Network backbones.} 
	We trained DeepLabv3$+$ with several network backbones on clean and corrupted data using TensorFlow~\cite{abadi_tensorflow:_2016}. 
	We utilized MobileNet-V2, ResNet-50, ResNet-101, Xception-41, Xception-65 and Xception-71 as network backbones. Every model has been trained with batch size $16$, crop-size $513 \times 513$, fine-tuning batch normalization parameters \cite{ioffe_sergey_batch_2015}, initial learning rate $ 0.01$ or $0.007$, and random scale data augmentation. 
	\\ \indent \textbf{Datasets.} We use PASCAL VOC 2012, the Cityscapes dataset, and ADE20K for training and validation. The training set of PASCAL VOC consists of $1,464$ train and $1,449$ validation images. We use the high-quality pixel-level annotations of Cityscapes, comprising of $2975$ train and $500$ validation images. We evaluated all models on original image dimensions. ADE20K consists of $20,210$ train, $2000$ validation images, and $150$ semantic classes.
	\\ \indent \textbf{Evaluation metrics.} We apply mean Intersection-over-Union as performance metric (mIoU) for every model and average over severity levels. 
	In addition, we use, and slightly modify, the concept of Corruption Error and relative Corruption Error from~\cite{hendrycks_benchmarking_2019} as follows.
    \\ \indent We use the term \textit{Degradation D}, where $D=1 - \mathit{mIoU}$ in place of \textit{Error}. 
    Degradations across severity levels, which are defined by the ImageNet-C corruptions~\cite{hendrycks_benchmarking_2019}, are often aggregated. 
    To make models mutually comparable, we divide the degradation $D$ of a trained model $f$ through the degradation of a reference model $\mathit{ref}$. 
    With this, the \textit{Corruption Degradation} (CD) of a trained model is defined as 
	\vspace{-0.1cm}
	\begin{equation}
	\vspace{-0.1cm}
	\mathit{CD_{c}^{f}} = \left(\sum_{s=1}^{5}D_{s,c}^{f}\right)\bigg/\left(\sum_{s=1}^{5}D_{s,c}^{\mathit{ref}}\right)
	\label{eq:CD}
	\end{equation}
	where $c$ denotes the corruption type (\eg, Gaussian blur) and $s$ its severity level. 
	Please note that for {\it category noise}, only the first three severity levels are taken into account. 
	While we predominately use CD for comparing the robustness of model architectures, we also consider the degradation of models relative to clean data, measured by the \textit{relative Corruption Degradation} (rCD).
	We highlight the difference between CD and rCD in more detail in the supplement.
	\vspace{-0.1cm}
	\begin{equation}
	\vspace{-0.1cm}
	\mathit{rCD_{c}^{f}} = \left(\sum_{s=1}^{5}D_{s,c}^{f} - D_{\mathit{clean}}^{f}\right)\bigg/\left(\sum_{s=1}^{5}D_{s,c}^{\mathit{ref}}-D_{\mathit{clean}}^{\mathit{ref}}\right)
	\label{eq:rCD}
	\end{equation}

	\subsection{Benchmarking Network Backbones}
	\label{subsec:benchmark}
    
    \begin{figure*}[!htbp]
	\centering
	\includegraphics[width=1\textwidth]{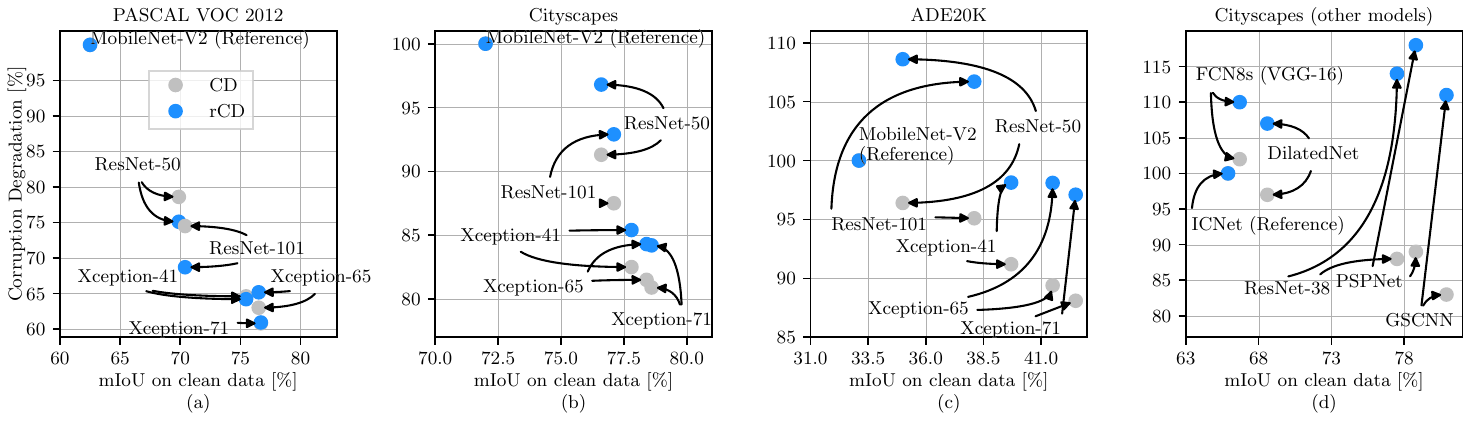}
	\caption{(a$-$c) CD and rCD for several network backbones of the DeepLabv3$+$ architecture evaluated on PASCAL VOC 2012, the Cityscapes dataset, and ADE20K. 
	MobileNet-V2 is the reference model in each case.
    rCD and CD values below \SI{100}{\%} represent higher robustness than the reference model. In almost every case, model robustness increases with model performance (\ie mIoU on clean data). Xception-71 is the most robust network backbone on each dataset. (d) CD and rCD for non-DeepLabv3$+$ based models evaluated on Cityscapes. While CD decreases with increasing performance on clean data, rCD is larger than \SI{100}{\%}. \vspace{-0.5cm}}
	\label{fig:benchmarks}
	\end{figure*}
	
	We trained various network backbones (MobileNet-V2, ResNets, Xceptions) on the original, clean training-sets of PASCAL VOC 2012, the Cityscapes dataset, and ADE20K. 
	Table \ref{tab:benchmark_cityscapes} shows the average mIoU for the Cityscapes dataset, and each corruption type averaged over all severity levels. 
	We refer to the supplement for the respective results for other datasets and individual severity levels.
	\\ 
	\indent 
	As expected, for DeepLabv3$+$, Xception-71 exhibits the best performance for clean data with an mIoU of \SI{78.6}{\%}\footnote{Note that we were not able to reproduce the results from~\cite{chen_encoder-decoder_2018}. We conjecture that this is due to hardware limitations, as we could not set the suggested crop-size of $769 \times 769$ for Cityscapes.}. 
	The bottom part of Table \ref{tab:benchmark_cityscapes} shows the benchmark results of non-DeepLab based models.
	\\ 
	\indent 
	\textbf{Network backbone performance}. Most Xception based models perform significantly better than ResNets and MobileNet-V2. GSCNN is the best performing architecture on clean data of this benchmark.
	\\ 
	\indent 
	\textbf{Performance \wrt blur.} 
	Interestingly, all models (except DilatedNet and VGG16) handle PSF blur well, as the respective mIoU decreases only by roughly \SI{2}{\%}. 
	Thus, even a lightweight network backbone such as MobileNet-V2 is hardly vulnerable against this realistic type of blur. 
	The number of both false positive and false negative pixel-level classifications increases, especially for far-distant objects. 
	With respect to Cityscapes this means that persons are simply overlooked or confused with similar classes, such as rider. 
	Please find some result images in the supplement. 
	\\ 
	\indent 
	\textbf{Performance \wrt noise.} 
	Noise has a substantial impact on model performance. 
	Hence we only averaged over the first three severity levels. 
	Xception-based network backbones of DeepLabv3$+$ often perform similar or better than non-DeepLabv3$+$ models. 
	MobileNet-V2, ICNet, VGG-16, and GSCNN handle the severe impact of image noise significantly worse than the other models.
	\\ 
	\indent 
	\textbf{Performance \wrt digital.} 
	The first severity levels of corruption types contrast, brightness, and saturation are handled well. 
	However, JPEG compression decreases performance by a large margin. 
	Notably, PSPNet and GSCNN have for this corruption halved or less mIoU than Xception-41 and -71, though their mIoU on clean data is similar.
	\\ 
	\indent 
	\textbf{Performance \wrt weather.} 
	Texture-corrupting distortions as snow and frost degrade mIoU of each model significantly. 
	\\ 
	\indent 
	\textbf{Performance \wrt Geometric distortion.} Models of DeepLabv3$+$ handle geometric distortion significantly better than non-DeepLabv3$+$ based models.
	\\ 
	\indent	
	This benchmark indicates, in general, a similar result as in~\cite{geirhos_imagenet-trained_2019}, that is image distortions corrupting the texture of an image (\eg, image noise, snow, frost, JPEG), often have distinct negative effect on model performance compared to image corruptions preserving texture to a certain point (\eg, blur, brightness, contrast, geometric distortion). 
    To evaluate the robustness \wrt image corruptions of proposed network backbones, it is also interesting to consider Corruption Degradation (CD) and relative Corruption Degradation (rCD). 
    Fig. \ref{fig:benchmarks} illustrates the mean CD and rCD with respect to the mIoU for {\it clean} images (lower values correspond to higher robustness regarding the reference model). 
    Each dot depicts the performance of one network backbone, averaged over all corruptions except for PSF blur\footnote{Due to the considerably smaller impact of PSF blur on model performance, small changes in mIoU of only tenths percentage can have a significant impact on the corresponding rCD.}.
    Subplot a$-$c illustrates respective results for PASCAL VOC 2012, Cityscapes, and ADE20K.
	On each dataset, Xception-71 is the most robust network backbone for DeepLabv3$+$ architecture. 
	Interestingly, rCD decreases often with increasing model performance, except for Xception-65 on PASCAL VOC 2012 (Fig. \ref{fig:benchmarks} a) and ResNets on ADE20K (Fig. \ref{fig:benchmarks} c). 
	The latter result indicates that ResNet-based backbones are vulnerable when applied for a large-scale dataset as ADE20K.
	Fig. \ref{fig:benchmarks} d presents the respective result for several non-DeepLabv3$+$ based segmentation models.
	The rCD for these models increases slightly.
	On the other hand, CD decreases mostly with increasing model performance on clean data. 
	The authors of~\cite{hendrycks_benchmarking_2019} report the same result for the task of full-image classification: 
	The rCD for established networks stays relatively constant, even though model performance on clean data differs significantly, as Fig. \ref{fig:benchmarks} d indicate. 
	When we, however, evaluate within a semantic segmentation architecture, as DeepLabv3$+$, the contrary result (\ie, decreasing rCD) is generally observed.
	The following speculation may also give further insights. 
	Geirhos \etal~\cite{geirhos_imagenet-trained_2019} stated recently that (i) DCNNs for full-image classification examine local textures, rather than global shapes of an object, to solve the task at hand, and (ii) model performance \wrt image corruption increases when the model relies more on object shape (rather than object texture).
	Transferring these results to the task of semantic segmentation, Xception-based backbones might have a more pronounced shape bias than others (\eg, ResNets), resulting hence in a higher rCD score \wrt image corruption. 
	This may be an interesting topic for future work, however, beyond the scope of this paper. 

\subsection{Ablation Study on Cityscapes}
	\label{subsec:as_cs}
\begin{table*}[!htbp]
\begin{adjustbox}{width=\textwidth}
\begin{tabular}{@{}ccccccccccccccccccccc@{}}
\toprule
&  & \multicolumn{5}{c}{\textbf{Blur}} & \multicolumn{5}{c}{\textbf{Noise}} & \multicolumn{4}{c}{\textbf{Digital}} & \multicolumn{4}{c}{\textbf{Weather}} &  \\ \midrule
\begin{tabular}[c]{@{}c@{}}Deeplab-v3+ \\ Backbone\end{tabular} & \multicolumn{1}{c|}{Clean} & Motion & Defocus & \begin{tabular}[c]{@{}c@{}}Frosted \\ Glass\end{tabular} & Gaussian & \multicolumn{1}{c|}{PSF} & Gaussian & Impulse & Shot & Speckle & \multicolumn{1}{c|}{Intensity} & Brightness & Contrast & Saturate & \multicolumn{1}{c|}{JPEG} & Snow & Spatter & Fog & \multicolumn{1}{c|}{Frost} & \begin{tabular}[c]{@{}c@{}}Geometric\\ Distortion\end{tabular} \\ \midrule
\textbf{Xception-71} & \multicolumn{1}{c|}{78.6} & 64.1 & 60.9 & 52.0 & 60.4 & \multicolumn{1}{c|}{76.4} & 14.9 & 10.8 & 19.4 & 41.2 & \multicolumn{1}{c|}{\textbf{50.2}} & \textbf{68.0} & 58.7 & 47.1 & \multicolumn{1}{c|}{40.2} & \textbf{18.8} & 50.4 & 64.1 & \multicolumn{1}{c|}{20.2} & 71.0 \\
w/o ASPP & \multicolumn{1}{c|}{73.9} & 60.7 & 59.5 & 51.5 & 58.4 & \multicolumn{1}{c|}{72.8} & \textbf{18.5} & \textbf{14.7} & \textbf{22.3} & 39.8 & \multicolumn{1}{c|}{44.7} & 63.4 & 56.2 & 42.7 & \multicolumn{1}{c|}{39.9} & 17.6 & 49.0 & 58.3 & \multicolumn{1}{c|}{21.8} & 69.3 \\
w/o AC & \multicolumn{1}{c|}{77.9} & 62.2 & 57.9 & 51.8 & 58.2 & \multicolumn{1}{c|}{76.1} & 7.7 & 5.7 & 11.2 & 32.8 & \multicolumn{1}{c|}{43.2} & 67.6 & 55.6 & 46.0 & \multicolumn{1}{c|}{\textbf{40.7}} & 18.2 & 50.1 & 61.1 & \multicolumn{1}{c|}{\textbf{21.6}} & 71.1 \\
w/o ASPP$+$w/ DPC & \multicolumn{1}{c|}{\textbf{78.8}} & 62.8 & 59.4 & 52.6 & 58.2 & \multicolumn{1}{c|}{76.9} & 7.3 & 2.8 & 10.7 & 33.0 & \multicolumn{1}{c|}{42.4} & 64.8 & 59.4 & 45.3 & \multicolumn{1}{c|}{32.0} & 14.4 & 48.6 & 64.0 & \multicolumn{1}{c|}{20.8} & 72.1 \\
w/o LRL & \multicolumn{1}{c|}{77.9} & 64.2 & \textbf{63.2} & 50.7 & \textbf{62.2} & \multicolumn{1}{c|}{76.7} & 13.9 & 9.3 & 18.2 & \textbf{41.3} & \multicolumn{1}{c|}{49.9} & 64.5 & 59.2 & 44.3 & \multicolumn{1}{c|}{36.1} & 16.9 & 48.7 & \textbf{64.3} & \multicolumn{1}{c|}{21.3} & 71.3 \\
w/ GAP & \multicolumn{1}{c|}{78.6} & \textbf{64.2} & 61.7 & \textbf{55.9} & 60.7 & \multicolumn{1}{c|}{\textbf{77.8}} & 9.7 & 8.4 & 13.9 & 36.9 & \multicolumn{1}{c|}{45.6} & 68.0 & \textbf{60.2} & \textbf{48.4} & \multicolumn{1}{c|}{40.6} & 16.8 & \textbf{51.0} & 62.1 & \multicolumn{1}{c|}{20.9} & \textbf{73.6} \\ \bottomrule
\end{tabular}
\end{adjustbox}
\caption{
Average mIoU for clean and corrupted variants of the Cityscapes validation dataset for Xception-71 and five  corresponding architectural ablations. 
Based on DeepLabv3$+$ we evaluate the removal of atrous spatial pyramid pooling (\textbf{ASPP}), atrous convolutions (\textbf{AC}), and long-range link (\textbf{LRL}). 
We further replaced ASPP by Dense Prediction Cell (\textbf{DPC}) and utilized global average pooling (\textbf{GAP}). Mean-IoU is averaged over severity levels. 
The standard deviation for image noise is $0.2$ or less.
Highest mIoU per corruption is bold.\vspace{-0.2cm}}
\label{tab:ablationstudy_cityscapes}
\end{table*}

\begin{figure*}
	\centering
	\includegraphics[width=\textwidth]{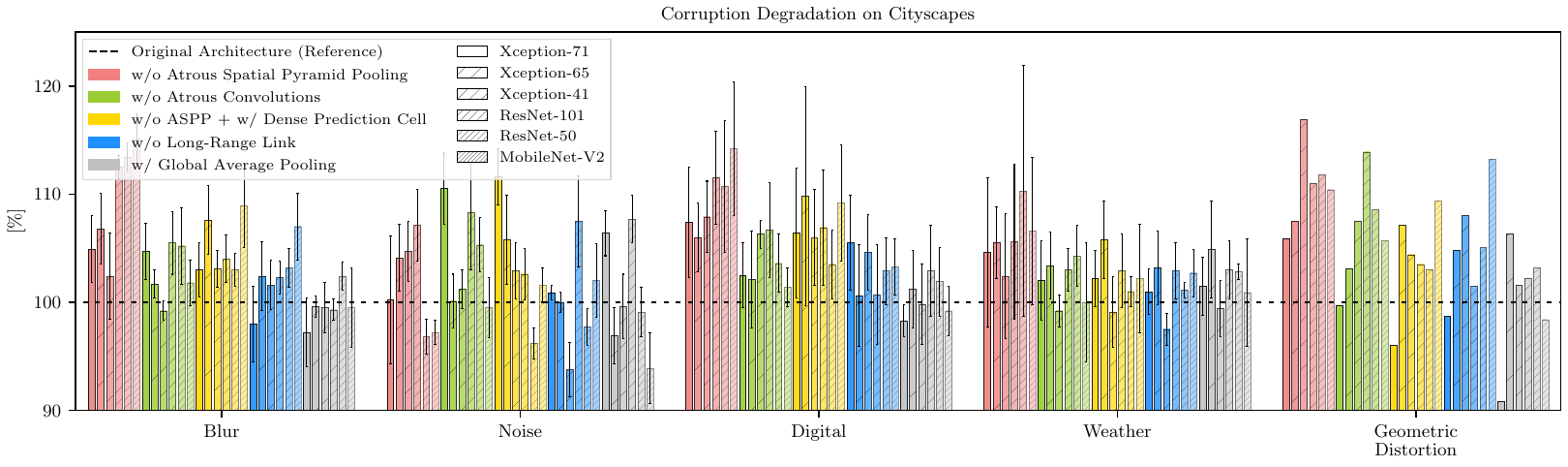}
	\caption{CD evaluated on Cityscapes for the proposed ablated variants of the DeepLabv3$+$ architecture \wrt image corruptions, employing six different network backbones. 
	Bars above \SI{100}{\%} represent a decrease in performance compared to the respective reference architecture.
	Each ablated architecture is re-trained on the original training dataset. 
	Removing ASPP reduces the model performance significantly. Atrous convolutions increase robustness against blur. The model becomes vulnerable against most effects when Dense Prediction Cell is used. 
	Each bar is the average CD of a corruption category, except for geometric distortion (error bars indicate the standard deviation).
		\vspace{-0.5cm}	}
	\label{fig:barplotcs}
\end{figure*}

Instead of solely comparing robustness across network backbones, we now conduct an extensive ablation study for DeepLabv3$+$. 
We employ the state-of-the-art performing Xception-71 (XC-71), Xception-65 (XC-65), Xception-41 (XC-41), ResNet-101, ResNet-50 and, their lightweight counterpart, MobileNet-V2 (MN-V2) (width multiplier 1, $224 \times 224$), as network backbones. 
XC-71 is the best performing backbone on clean data, but at the same time, computationally most expensive. 
The efficient MN-V2, on the other hand, requires roughly ten times less storage space. We ablated for each network backbone of the DeepLabv3$+$ architecture the same architectural properties as listed in section~\ref{subsec:ablations}. 
Each ablated variant has been re-trained on clean data of Cityscapes, PASCAL VOC 2012, and ADE20K, summing up to over 100 trainings.
Table \ref{tab:ablationstudy_cityscapes} shows the averaged mIoU for XC-71, evaluated on Cityscapes. 
We refer to the supplement for the results of the remaining backbones.
In the following sections, we discuss the most distinct, statistically significant results. 
\\ \indent 
We see that with Dense Prediction Cell (DPC), we achieve the highest mIoU on clean data followed by the reference model. 
We also see that removing ASPP reduces mIoU significantly. 
\\ To better understand the robustness of each ablated model, we illustrate the average CD within corruption categories (\eg, blur) in Fig.~\ref{fig:barplotcs} (bars above \SI{100}{\%} indicate reduced robustness compared to the respective reference model).
\\ 
\indent 
\textbf{Effect of ASPP.} Removal of ASPP reduces model performance significantly (Table \ref{tab:ablationstudy_cityscapes} first column). We refer to the supplement for an evaluation.
\\ 
\indent 
\textbf{Effect of AC.} 
Atrous convolutions (AC) generally show a positive effect \wrt corruptions of type blur for most network backbones, especially for XC-71 and ResNets. 
For example, without AC, the average mIoU for defocus blur decreases by \SI{3.8}{\%} for ResNet-101 (CD$\ =\SI{109}{\%}$). 
Blur reduces high-frequency information of an image, leading to similar signals stored in consecutive pixels. 
Applying AC can hence increase the amount of information per convolution filter, by skipping direct neighbors with similar signals.
Regarding XC-71 and ResNets, AC clearly enhance robustness on noise-based corruptions.
The mIoU for the first severity level of Gaussian noise are \SI{12.2}{\%} (XC-71), \SI{10.8}{\%} (ResNet-101), \SI{8.0}{\%} (ResNet-50) less than respective reference. 
AC generally exhibit also a positive effect \wrt geometric distortion.
For MN-V2 and ResNets, the averaged mIoU reduces by up to \SI{4.2}{\%} 
(CD\textsuperscript{ResNet-50}$=\SI{109}{\%}$, CD\textsuperscript{ResNet-101}$=\SI{114}{\%}$, CD\textsuperscript{MN-V2}$=\SI{106}{\%}$).
In summary, AC often increase robustness against a broad range of network backbones and image corruptions. 
\\ 
\indent 
\textbf{Effect of DPC.} When employing Dense Prediction Cell (DPC) instead of ASPP, the models become clearly vulnerable against corruptions of most categories. While this ablated architecture 
reaches the highest mIoU on clean data for XC-71, it is less robust to a broad range of corruptions. 
For example, CD for defocus blur on MN-V2 and XC-65 are \SI{113}{\%} and \SI{110}{\%}, respectively. 
Average mIoU decreases by \SI{6.8}{\%} and by \SI{4.1}{\%}. 
For XC-71, CD for all corruptions of category noise are within \SI{109}{\%} and \SI{115}{\%}.
The average mIoU of this ablated variant is least for all, but one type of noise (Table~\ref{tab:ablationstudy_cityscapes}). 
Similar behavior can be observed for other corruptions and backbones.
DPC has been found through a neural-architecture-search (NAS, \eg,~\cite{zoph_learning_2018,zoph_neural_2017,pham_efficient_2018}) with the objective of maximizing performance on clean data.
This result indicates that such architectures tend to over-fit on this objective, \ie clean data.
It may be an interesting topic to evaluate robustness \wrt image corruptions for other NAS-based architectures as future work, however, is beyond the scope of this paper.
Consequently, performing NAS on corrupted data might deliver interesting findings of robust architectural properties--similar as in~\cite{cubuk_intriguing_2018} \wrt adversarial examples.
We further hypothesize that DPC learns less multi-scale representations than ASPP, which may be useful against common corruptions.
We discuss this hypothesis in the supplement.
\begin{table*}
\centering
\begin{adjustbox}{width=1\textwidth}
\begin{tabular}{@{}ccc|ccc|cc|ccc|cc|ccc|cc|ccc|ccccc@{}}
\toprule
\textbf{Ablation} & \multicolumn{5}{c|}{\textbf{w/o ASPP}} & \multicolumn{5}{c|}{\textbf{w/o AC}} & \multicolumn{5}{c|}{\begin{tabular}[c]{@{}c@{}}\textbf{w/o ASPP}\\ \textbf{w/ DPC}\end{tabular}} & \multicolumn{5}{c|}{\textbf{w/o LRL}} & \multicolumn{5}{c}{\textbf{w/ GAP}} \\ \midrule
\multirow{2}{*}{\begin{tabular}[c]{@{}c@{}}\textbf{Network}\\ \textbf{Backbone}\end{tabular}} & \multicolumn{2}{c|}{ResNet$-$} & \multicolumn{3}{c|}{Xception$-$} & \multicolumn{2}{c|}{ResNet$-$} & \multicolumn{3}{c|}{Xception$-$} & \multicolumn{2}{c|}{ResNet$-$} & \multicolumn{3}{c|}{Xception$-$} & \multicolumn{2}{c|}{ResNet$-$} & \multicolumn{3}{c|}{Xception$-$} & \multicolumn{2}{c|}{ResNet$-$} & \multicolumn{3}{c}{Xception$-$} \\
 & 50 & 101 & 41 & 65 & 71 & 50 & 101 & 41 & 65 & 71 & 50 & 101 & 41 & 65 & 71 & 50 & 101 & 41 & 65 & 71 & 50 & \multicolumn{1}{c|}{101} & 41 & 65 & 71 \\ \midrule
\textbf{Blur} & \textbf{120} & 117 & 115 & 118 & \textbf{119} & \textbf{102} & 99 & 99 & 98 & \textbf{100} & \textbf{103} & 101 & 100 & 104 & \textbf{109} & \textbf{102} & 101 & 97 & 102 & \textbf{104} & \textbf{98} & \multicolumn{1}{c|}{98} & 98 & 95 & \textbf{101} \\
\textbf{Noise} & 124 & \textbf{127} & 122 & \textbf{126} & 123 & 100 & \textbf{106} & 103 & 100 & 101 & 99 & \textbf{102} & 98 & 103 & \textbf{105} & 100 & \textbf{103} & 96 & \textbf{101} & 95 & 94 & \multicolumn{1}{c|}{\textbf{97}} & \textbf{99} & 97 & 98 \\
\textbf{Digital} & \textbf{133} & 128 & \textbf{127} & 124 & 124 & \textbf{103} & 101 & 102 & 101 & \textbf{103} & \textbf{104} & 102 & 101 & 103 & \textbf{105} & \textbf{103} & 102 & 98 & \textbf{103} & 103 & 95 & \multicolumn{1}{c|}{\textbf{96}} & \textbf{98} & 97 & 94 \\
\textbf{Weather} & \textbf{121} & 119 & \textbf{120} & 114 & 118 & \textbf{101} & 100 & 101 & 99 & \textbf{104} & \textbf{102} & 100 & 103 & 102 & \textbf{105} & \textbf{101} & 100 & 100 & 101 & \textbf{103} & \textbf{94} & \multicolumn{1}{c|}{93} & \textbf{98} & 95 & 96 \\
\textbf{\begin{tabular}[c]{@{}c@{}}Geometric\\ Distortion\end{tabular}} & \textbf{133} & 124 & \textbf{128} & 118 & 117 & \textbf{104} & 102 & \textbf{104} & 100 & 102 & \textbf{107} & 106 & \textbf{104} & 100 & 101 & \textbf{105} & 105 & 100 & \textbf{102} & 102 & \textbf{99} & \multicolumn{1}{c|}{98} & \textbf{102} & 101 & 101 \\ \bottomrule
\end{tabular}
\end{adjustbox}
\caption{CD evaluated on PASCAL VOC 2012 for ablated network backbones of the DeepLabv3$+$ architecture \wrt image corruptions. 
\vspace{-0.5cm}
}
\label{tab:rmcd_pascal}
\end{table*}
\\ 
\indent 
\textbf{Effect of LRL.} 
A long-range link (LRL) appears to be very beneficial for ResNet-101 against image noise. 
The model struggles especially for our noise model, as its CD equals \SI{116}{\%}.
For XC-71, corruptions of category digital as \textit{brightness} have considerably high CDs (\eg, CD\textsuperscript{XC-71}$=\SI{111}{\%}$).
For MN-V2, removing LRL decreases robustness \wrt defocus blur and geometric distortion as average mIoU reduces by \SI{5.1}{\%} (CD$\ =\SI{110}{\%}$) and \SI{4.6}{\%} (CD$\ =\SI{113}{\%}$).
\\ 
\indent 
\textbf{Effect of GAP.} 
Global average pooling (GAP) increases slightly robustness \wrt blur for most Xceptions. 
Interestingly, when applied in XC-71 (ResNet-101), the model is vulnerable to image noise.
Corresponding CD values range between \SI{103}{\%} and \SI{109}{\%} (\SI{106}{\%} and \SI{112}{\%}). 



\subsection{Ablation Study on Pascal VOC 2012}
\label{subsec:as_pascal}
We generally observe that the effect of the architectural ablations for DeepLabv3$+$ trained on PASCAL VOC 2012 is not always similar to previous results on Cityscapes (see Table~\ref{tab:rmcd_pascal}). Since this dataset is less complex than Cityscapes, the mIoU of ablated architectures differ less.
\\ 
\indent 
We do not evaluate results on MN-V2, as the model is not capable of giving a comparable performance.
Please see the supplement corresponding mIoU scores.
\\
\indent 
\textbf{Effect of ASPP.} 
Similar to the results on Cityscapes, removal of ASPP reduces model performance of each network backbone significantly. 
\\ 
\indent 
\textbf{Effect of AC.} 
Unlike on Cityscapes, atrous convolutions show no positive effect against blur. We explain this with the fundamentally different datasets. On Cityscapes, a model without AC often overlooks classes covering small image-regions, especially when far away. Such images are hardly present in PASCAL VOC 2012. 
As on Cityscapes, AC slightly helps performance for most models \wrt geometric distortion. For XC-41 and ResNet-101, we see a positive effect of AC against image noise.
\\ 
\indent 
\textbf{Effect of DPC.} 
As on Cityscapes, DPC decreases robustness for many corruptions. Generally, CD increases from XC-41 to XC-71. The impact on XC-71 is especially strong as indicated by the CD score, averaged over all corruptions, is \SI{106}{\%}. 
A possible explanation might be that the neural-architecture-search (NAS) \eg,~\cite{zoph_learning_2018,zoph_neural_2017,pham_efficient_2018}
 has been performed on XC-71 and enhances, therefore, the over-fitting effect additionally, as discussed in section~\ref{subsec:as_cs}.
\\ 
\indent 
\textbf{Effect of LRL.} 
Removing LRL increases robustness against noise for XC-71 and XC-41, probably due to discarding early features (we refer to the supplement for discussion).
However, this finding does not hold for XC-65. 
As reported in section~\ref{subsec:benchmark}, on PASCAL VOC 2012, XC-65 is also the most robust model against noise. 
Regarding ResNets, the LRL affects the image corruption of category geometric distortion the most.
\\ 
\indent 
\textbf{Effect of GAP.} 
When global average pooling is applied, the overall robustness of every network backbone increases particularly significant. 
The mIoU on clean data increases for every model (up to \SI{2.2}{\%} for ResNet-101, probably due to the difference between PASCAL VOC 2012 and the remaining dataset (we refer to supplement).
\subsection{Ablation Study on ADE20K}
\label{subsec:as_ade20k}
The performance on clean data ranges from MN-V2 (mIoU of \SI{33.1}{\%}) to XC-71 using DPC, as best-performing model, achieving an mIoU of \SI{42.5}{\%} (detailed results listed in the supplement). 
The performance on clean data for most Xception-based backbones (ResNets) is highest when Dense Prediction Cell (global average pooling) is used. 
Our evaluation shows that the mean CD for each ablated architecture is often close to \SI{100.0}{\%}.
The impact of proposed architectural properties on model performance is thus on the large-scale dataset ADE20K hardly present. 
A possible explanation is probably that the effect of architectural design choices becomes more decisive, and respective impacts are more pronounced when models perform well, \ie have large mIoU. 
DeepLabv3$+$ performs much poorer on ADE20K than, \eg, on the Cityscapes dataset. 
\\ \indent The tendencies of the previous findings are nevertheless present. 
Regarding XC-71, for example, the corresponding means of both CD and rCD for DPC are respectively \SI{101}{\%} and \SI{107}{\%}, showing its robustness is again less than the reference model. 
ASPP, on the other hand, affects segmentation performance also significantly. 

\section{Conclusion}
\label{subsec:conclusion}
We have presented a detailed, large-scale evaluation of state-of-the-art semantic segmentation models with respect to real-world image corruptions. 
Based on the study, we can introduce robust model design rules:
Atrous convolutions are generally recommended since they increase robustness against many corruptions.
The vulnerability of Dense Prediction Cell to many corruptions must be considered, especially in low-light and safety-critical applications. 
The ASPP module is important for decent model performance, especially for digitally and geometrically distorted input. Global average pooling should always be used on PASCAL VOC 2012.
Our detailed study may help to improve on the state-of-the-art for robust semantic segmentation models.

{\small
	\bibliographystyle{ieee_fullname}
	\bibliography{literatur}
}

\setcounter{section}{0}
\renewcommand{\thesection}{\Alph{section}}
\counterwithin{table}{section}
\counterwithin{figure}{section}

\clearpage
{\large \textbf{Supplemental Material}}
\\
\\
\indent 
\\
We provide further information about the utilized image corruptions and the conducted experiments. 
In more detail, we first show examples of every image corruption, and we give further details of our proposed image corruptions (section~\ref{sec:secA}). 
To make the image corruptions mutually comparable, we provide the Signal-to-Noise ratio for image corruptions of category noise (section~\ref{sec:rebuttal_snr_table}).
\indent 
\\
In section~\ref{sec:B}, we provide supplementary information about the experimental setup (section~\ref{sec:architectures}, \ref{sec:expdetails}), we explain the difference of the utilized evaluation metrics (\ie, CD and rCD) in more detail (section~\ref{sec:rebuttal_diff_metrics}), we discuss possible causes of the effect of architectural design choices (section~\ref{sec:rebuttal_theory}), and we show qualitative results (section~\ref{sec:qualitativeresults}).
\indent 
\\
In addition, we report the individual evaluation metric scores (\ie, mIoU, CD, and rCD) for Cityscapes (section~\ref{sec:experimentalresults_cityscapes}), PASCAL VOC 2012 (section~\ref{sec:experimentalresults_pascal}), and ADE20K (section~\ref{sec:experimentalresults_ade}). 
We further show the performance of many models for different severity levels of many image corruptions (section~\ref{sec:rebuttal_degrade_severity_levels}).

\section{Image Corruption Models}
\label{sec:secA}
\subsection{ImageNet-C}
\begin{figure*}[h]
\vspace{+3cm}
\centering
    \begin{subfigure}[t!]{0.30\linewidth}
		\centering
		\includegraphics[width=1\linewidth]{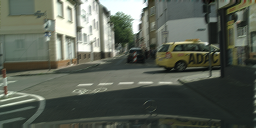}
	\end{subfigure}~
	\begin{subfigure}[t!]{0.30\linewidth}
		\centering
		\includegraphics[width=1\linewidth]{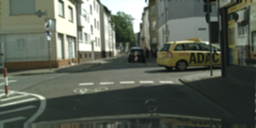}
	\end{subfigure}~
	\begin{subfigure}[t!]{0.30\linewidth}
		\centering
		\includegraphics[width=1\linewidth]{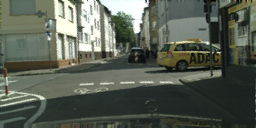}
	\end{subfigure}~
	
    \begin{subfigure}[t!]{0.30\linewidth}
		\centering
		\includegraphics[width=1\linewidth]{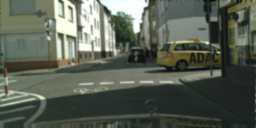}
	\end{subfigure}~
	\begin{subfigure}[t!]{0.30\linewidth}
		\centering
		\includegraphics[width=1\linewidth]{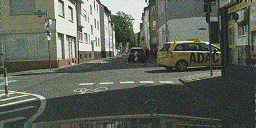}
	\end{subfigure}~
	\begin{subfigure}[t!]{0.30\linewidth}
		\centering
		\includegraphics[width=1\linewidth]{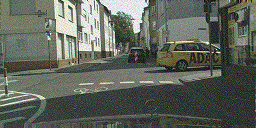}
	\end{subfigure}~
	
    \begin{subfigure}[t!]{0.30\linewidth}
		\centering
		\includegraphics[width=1\linewidth]{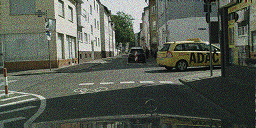}
	\end{subfigure}~
	\begin{subfigure}[t!]{0.30\linewidth}
		\centering
		\includegraphics[width=1\linewidth]{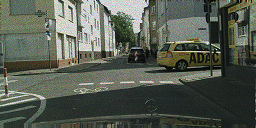}
	\end{subfigure}~
	\begin{subfigure}[t!]{0.30\linewidth}
		\centering
		\includegraphics[width=1\linewidth]{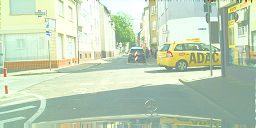}
	\end{subfigure}~
	
    \begin{subfigure}[t!]{0.30\linewidth}
		\centering
		\includegraphics[width=1\linewidth]{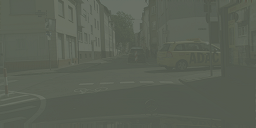}
	\end{subfigure}~
	\begin{subfigure}[t!]{0.30\linewidth}
		\centering
		\includegraphics[width=1\linewidth]{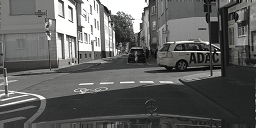}
	\end{subfigure}~
	\begin{subfigure}[t!]{0.30\linewidth}
		\centering
		\includegraphics[width=1\linewidth]{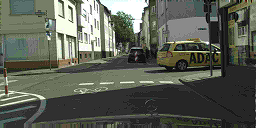}
	\end{subfigure}~
	
    \begin{subfigure}[t!]{0.30\linewidth}
		\centering
		\includegraphics[width=1\linewidth]{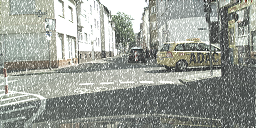}
	\end{subfigure}~
	\begin{subfigure}[t!]{0.30\linewidth}
		\centering
		\includegraphics[width=1\linewidth]{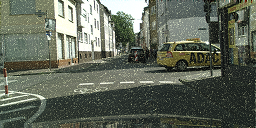}
	\end{subfigure}~
	\begin{subfigure}[t!]{0.30\linewidth}
		\centering
		\includegraphics[width=1\linewidth]{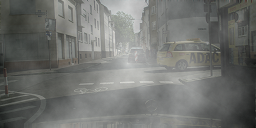}
	\end{subfigure}~
	
    \begin{subfigure}[t!]{0.30\linewidth}
	\includegraphics[width=1\linewidth]{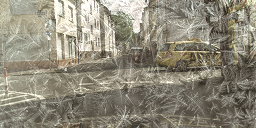}
	\end{subfigure}
	\caption{Illustration of  utilized image corruptions of ImageNet-C. First row: Motion blur, defocus blur, frosted glass blur. Second row: Gaussian blur, Gaussian noise, impulse noise. Third row: Shot noise, speckle noise, brightness. Fourth row: Contrast, saturate, JPEG. Fifth row: Snow, spatter, fog. Sixth row: frost.\vspace{+1.5cm}}
	\label{fig:imagenetc_corruptions}	
\end{figure*}

In this section, we illustrate the used image corruptions of ImageNet-C\footnote{\url{https://github.com/hendrycks/robustness/tree/master/ImageNet-C}}. 
Figure~\ref{fig:imagenetc_corruptions} shows the image corruptions of the categories blur, noise, digital, and weather of ImageNet-C. 
To make the image corruption clearly visible, we selected each example of severity level three or higher. 
The Figure is best viewed in color.

\begin{table*}[h]
\vspace{+2cm}
\centering
\begin{tabular}{cccc}
Angle of Incidence & Severity Level 1 & Severity Level 2 & Severity Level 3 \\
\\
\multicolumn{1}{c}{$0^\circ$} & \raisebox{-.5\height}{\includegraphics[width=0.24\linewidth]{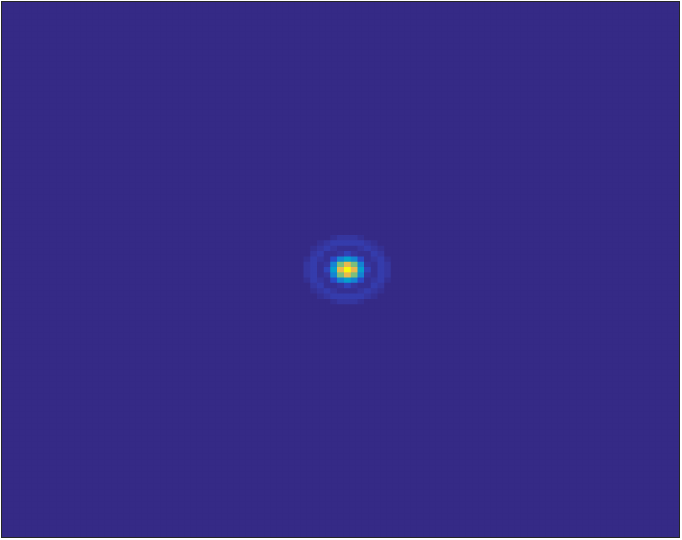}} & \raisebox{-.5\height}{\includegraphics[width=0.24\linewidth]{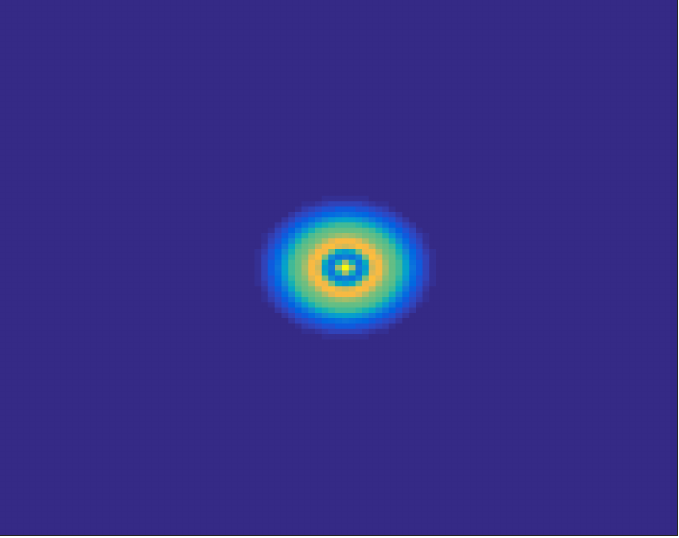}} &  \raisebox{-.5\height}{\includegraphics[width=0.24\linewidth]{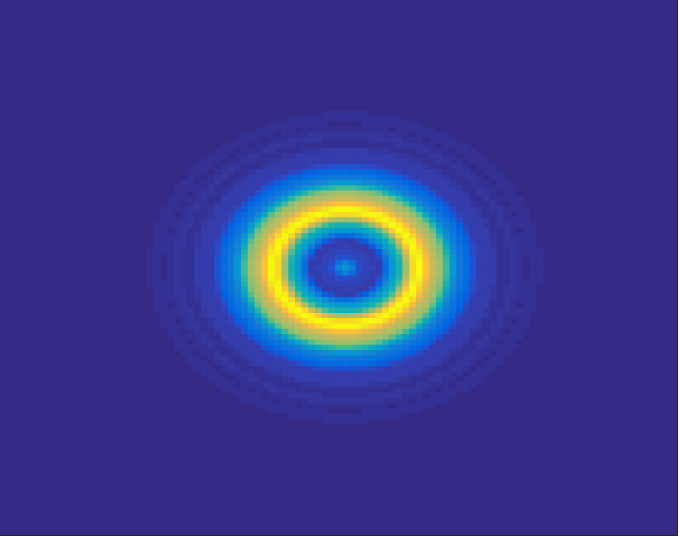}} \\

\multicolumn{1}{c}{$15^\circ$} & \raisebox{-.5\height}{\includegraphics[width=0.24\linewidth]{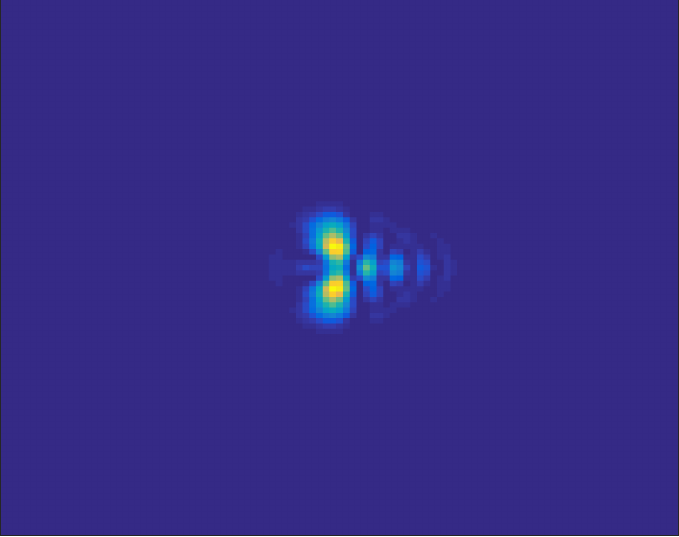}} & \raisebox{-.5\height}{\includegraphics[width=0.24\linewidth]{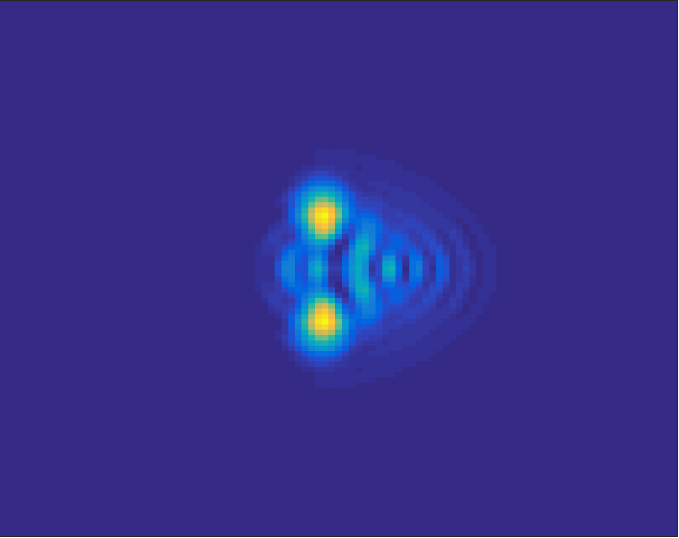}} &  \raisebox{-.5\height}{\includegraphics[width=0.24\linewidth]{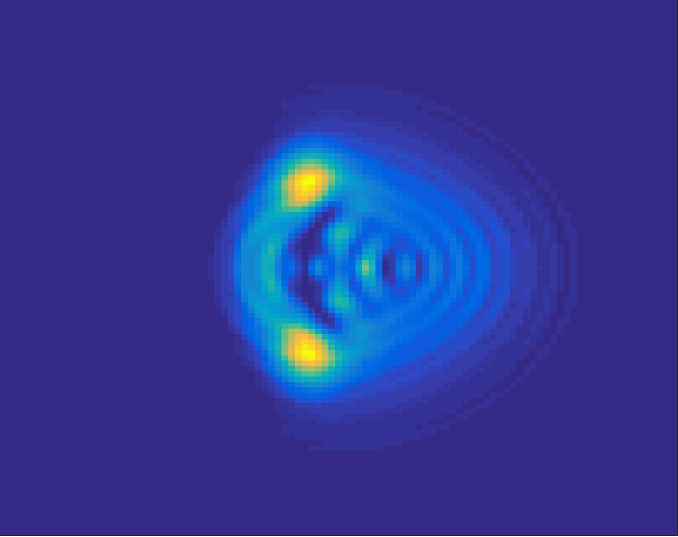}} \\

\multicolumn{1}{c}{$30^\circ$} & \raisebox{-.5\height}{\includegraphics[width=0.24\linewidth]{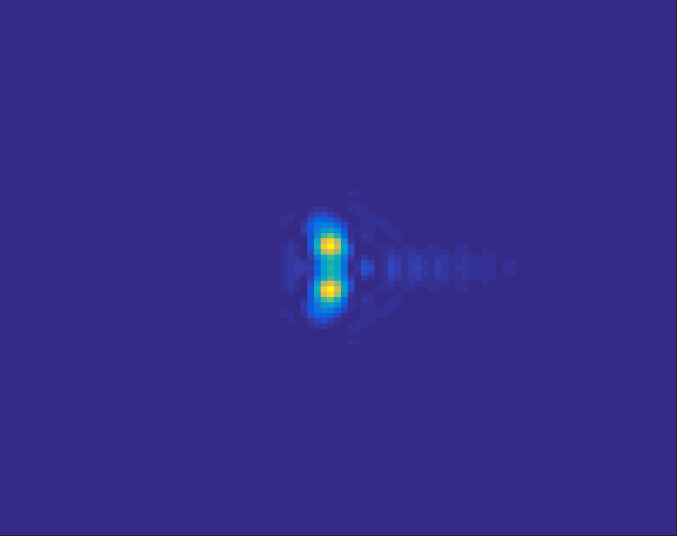}} & \raisebox{-.5\height}{\includegraphics[width=0.24\linewidth]{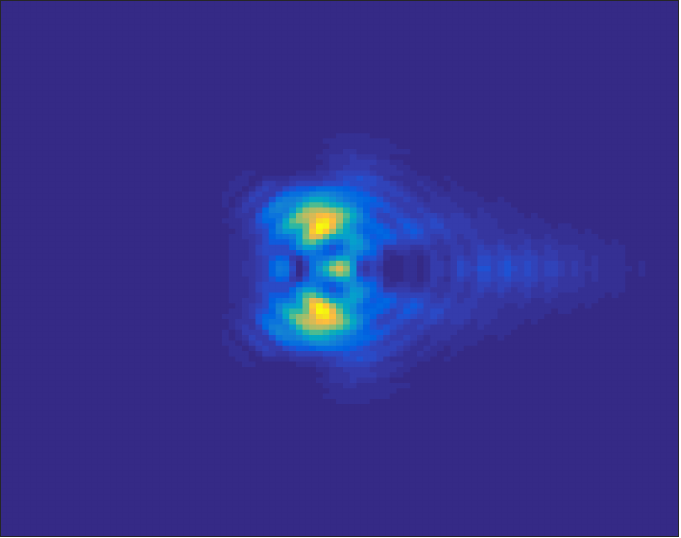}} &  \raisebox{-.5\height}{\includegraphics[width=0.24\linewidth]{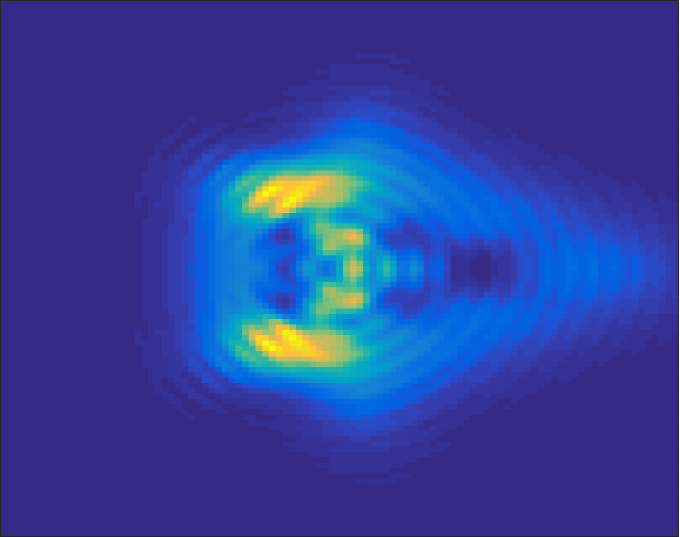}} \\

\multicolumn{1}{c}{$45^\circ$} & \raisebox{-.5\height}{\includegraphics[width=0.24\linewidth]{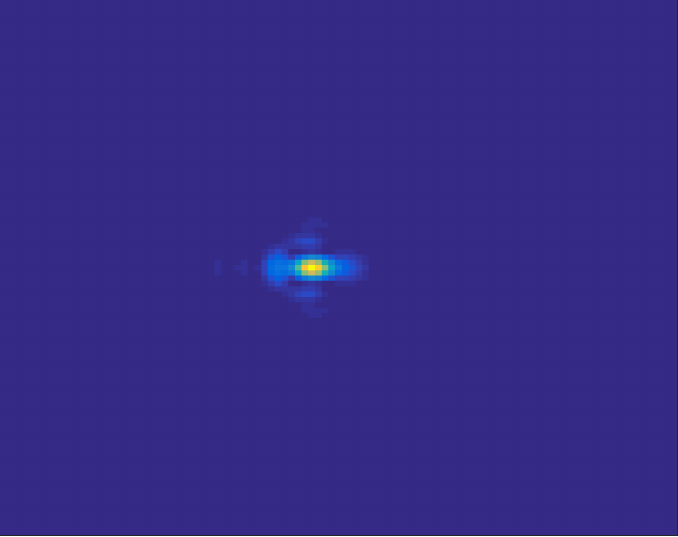}} & \raisebox{-.5\height}{\includegraphics[width=0.24\linewidth]{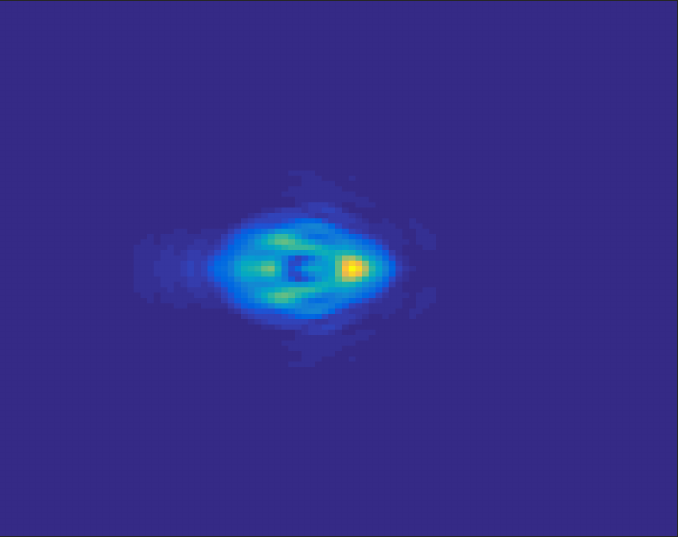}} &  \raisebox{-.5\height}{\includegraphics[width=0.24\linewidth]{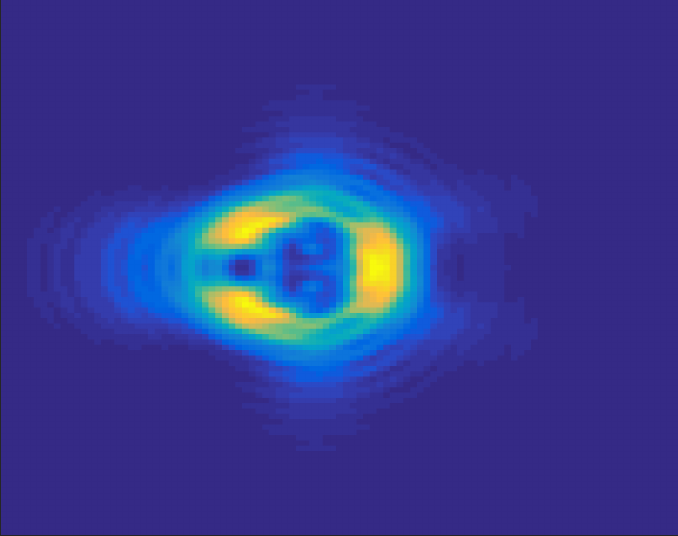}} \\
\end{tabular}
\captionof{figure}{The intensity distribution of used PSF kernels. 
The degree of the spatial distribution of intensity increases with the severity level. 
The shape of the PSF kernel depends on the image region, \ie, the angle of incidence.\vspace{+0.5cm}}
\label{tab:psfen}
\end{table*}

\begin{figure*}[h]
\centering
    \begin{subfigure}[t!]{0.30\linewidth}
		\centering
		\includegraphics[width=1\linewidth]{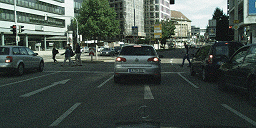}
	\end{subfigure}~
    \begin{subfigure}[t!]{0.30\linewidth}
		\centering
		\includegraphics[width=1\linewidth]{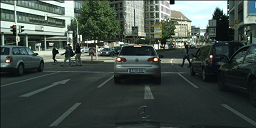}
	\end{subfigure}~
    \begin{subfigure}[t!]{0.30\linewidth}
		\centering
		\includegraphics[width=1\linewidth]{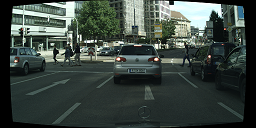}
	\end{subfigure}~
	\caption{Illustration of our proposed image corruptions. From left to right: Proposed noise model, PSF blur, and geometric distortion. Best viewed in color.\vspace{+0.0cm}}
	\label{fig:realisticcorruptions}	
	
\end{figure*}

\subsection{Proposed Image Corruptions}
\begin{figure}[H]
	\centering
	\includegraphics[width=0.45\linewidth]{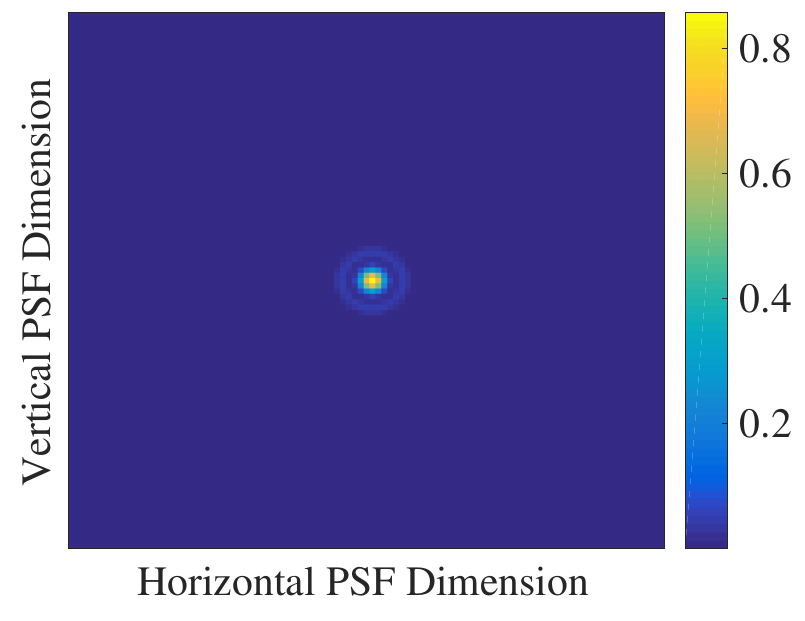}
	\caption{The normalized intensity distribution of a PSF kernel of our proposed PSF blur.}
	\label{fig:psf0}
\end{figure}

In this section, we provide more details about our image corruptions, \ie, the proposed image noise model, PSF blur, and geometric distortion. 
Figure~\ref{fig:realisticcorruptions} shows examples of our proposed image corruptions. 
\\ \indent \textbf{Intensity-Dependent Noise Model.} 
In the main paper, we proposed a noise model that incorporates intensity-dependent chrominance und luminance noise components that are both added to original pixel intensities in linear color space. 
Here, the term \textit{chrominance noise} means that a random noise component for an image pixel is drawn for each color channel independently, resulting thus in color noise. 
\textit{Luminance noise}, on the other hand, refers to a random noise value that is added to each channel of a pixel equally, resulting hence in gray-scale noise.
We model the noisy pixel intensity for a color channel $c$ as a random variable $I_{\mathit{noise},c}$:
\begin{equation}
\begin{aligned}
I_{\mathit{noise},c}(\Phi_{c},N_{\mathit{luminance}},N_{\mathit{chrominance,c}};w_{s}) =
\\log_2(2^{\Phi_{c}}+w_{s}\cdot (N_{\mathit{luminance}} + N_{\mathit{chrominance,c}}))
\end{aligned}
\end{equation}    
where $\Phi_{c}$ is the normalized pixel intensity of color channel $c$, $N_{\mathit{luminance}}$ and $N_{\mathit{chrominance}}$ are random variables following a Normal distribution with mean $\mu=0$ and standard deviation $\sigma=1$, $w_{s}$ is a weight factor, parameterized by severity level $s$.
\\
\indent 
\textbf{PSF blur.} 
Every optical system, \eg, the lens array of a camera, exhibits optical aberrations. 
Many of them cause image blur. 
Point-spread-functions aggregate every optical aberration that results in blur. 
The point-spread-functions of an optical system are typically spatially-varying, meaning, for example, that the degree of blur is at the image edge more pronounced than in the center of the image.
Figure~\ref{fig:psf0} illustrates the intensity distribution of a PSF kernel, where most of its energy is punctually centered.

Figure~\ref{tab:psfen} illustrates the intensity distribution of several PSF kernels utilized in the main paper. 
Each row corresponds to a specific PSF blur kernel at the respective angle of incidence, \ie, the higher the angle of incidence, the higher the distance to the image center. 
Note that the PSF kernel varies its shape within a severity level (\ie, column). 
The intensity of a PSF kernel is spatially more distributed for higher severity levels.
\\ \indent \textbf{Geometric distortion.} 
Distortion parameters of an optical system vary over time, are affected by environmental influences, differ from calibration stages, and thus, may never be fully compensated. 
Additionally, image warping may introduce re-sampling artifacts, degrading the informational content of an image. 
It can hence be preferable to utilize the original (i.e., geometrically distorted) image~\cite[p.192f]{hartley_multiple_2003}.
We used the command-line tool \textit{ImageMagick} to apply a radially-symmetric barrel distortion as a polynomial of grade 4 to both the RGB and ground-truth images. 
It is essential to use the nearest-neighbor filter for color determination of the ground truth, as otherwise, the class labels are corrupted. 

\subsection{Signal-to-Noise Ratio for Image Corruptions}
\label{sec:rebuttal_snr_table}
To make the severity levels of corruptions of category image noise mutually comparable, we provide in Tab.~\ref{tab:snr} the SNR for this image corruption category.

\begin{table}
\centering
\begin{adjustbox}{width=1\linewidth}
\begin{tabular}{@{}ccccc@{}}
\toprule
\textbf{} & \textbf{\begin{tabular}[c]{@{}c@{}}Severity\\ Level\end{tabular}} & \textbf{Cityscapes} & \textbf{\begin{tabular}[c]{@{}c@{}}PASCAL \\ VOC 2012\end{tabular}} & \textbf{ADE20K} \\ \midrule
\textbf{Gaussian Noise} & 1 & 13.2 & 18.6 & 18.3 \\
 & 2 & 9.9 & 15.5 & 15.2 \\
 & 3 & 6.8 & 12.4 & 12.1 \\
 & 4 & 4.1 & 9.8 & 9.6 \\
 & 5 & 1.7 & 7.3 & 7.3 \\ \midrule
\textbf{Impulse Noise} & 1 & 11.2 & 16.7 & 16.5 \\
 & 2 & 8.1 & 13.8 & 13.6 \\
 & 3 & 6.4 & 12.1 & 12.0 \\
 & 4 & 3.6 & 9.3 & 9.3 \\
 & 5 & 1.6 & 7.2 & 7.4 \\ \midrule
\textbf{Shot Noise} & 1 & 14.2 & 18.2 & 17.8 \\
 & 2 & 10.5 & 14.9 & 14.3 \\
 & 3 & 7.4 & 12.1 & 11.5 \\
 & 4 & 3.9 & 8.9 & 8.2 \\
 & 5 & 2.0 & 7.2 & 6.4 \\ \midrule
\textbf{Speckle Noise} & 1 & 17.0 & 19.3 & 18.9 \\
 & 2 & 14.5 & 17.1 & 16.6 \\
 & 3 & 9.8 & 12.8 & 12.2 \\
 & 4 & 7.9 & 11.0 & 10.3 \\
 & 5 & 5.8 & 9.2 & 8.4 \\ \midrule
\textbf{Intensity Noise} & 1 & 20.5 & 28.4 & 24.9 \\
 & 2 & 18.6 & 22.1 & 23.1 \\
 & 3 & 14.4 & 18.7 & 20.3 \\
 & 4 & 10.8 & 15.1 & 15.5 \\
 & 5 & 7.1 & 11.2 & 11.6 \\ \bottomrule
\end{tabular}
\end{adjustbox}
\caption{Averaged Signal-to-Noise ratios for corrupted variants of category image noise of the validation sets of Cityscapes, PASCAL VOC 2012, and ADE20K.
}
\label{tab:snr}
\end{table}

\section{Experiments}
\label{sec:B}
This section contains qualitative results and the remaining evaluation metric scores of the main paper.
Based on DeepLabv3+, we evaluate the removal of atrous spatial pyramid pooling (ASPP), atrous convolutions (AC), and long-range link (LRL). 
We further replaced ASPP by Dense Prediction Cell (DPC) or applied global average pooling (GAP). 
Each ablated variant has been re-trained on the corresponding clean training data of Cityscapes, PASCAL VOC 2012, or ADE20K. 
To guarantee comparable results, we also re-trained the original, non-ablated models, though several publicly available checkpoints were available.
\\ \indent Besides DeepLabv3+, we have benchmarked several other semantic segmentation models as well. 

\subsection{Architectures}
\label{sec:architectures}
The DeepLabv3$+$ architecture functions\footnote{\url{https://github.com/tensorflow/models/tree/master/research/deeplab}} as reference model in our ablation study.
We have benchmarked many other semantic segmentation models, such as FCN8s-VGG16 (trained by us), ICNet\footnote{\url{https://github.com/hszhao/ICNet}}, DilatedNet\footnote{\url{https://github.com/fyu/dilation}}, ResNet-38\footnote{\url{https://github.com/itijyou/ademxapp}}, PSPNet\footnote{\url{https://github.com/hszhao/PSPNet}}, and the recent Gated-ShapeCNN (GSCNN)\footnote{\url{https://github.com/nv-tlabs/GSCNN}}, using mostly publicly available model checkpoints. 
Regarding PSPNet and ICNet, we conducted the benchmark using a PyTorch~\cite{paszke_automatic_2017,shah_semantic_2017} implementation\footnote{\url{https://github.com/meetshah1995/pytorch-semseg}}.


\subsection{Experimental Details}
\label{sec:expdetails}
\indent
\textbf{Hardware Setup.} 
We have trained the models on machines equipped with four GTX 1080 Ti, each having \SI{11}{GB} of memory or on a machine with two Titan RTX, each having \SI{24}{GB} of memory.
\\
\indent
\textbf{Training Details.} 
We set the crop size in every training for every dataset to $513$, used a batch size of $16$, as we always fine-tuned the batch normalization parameters. 
We applied the original training protocol of the developers of DeepLabv3$+$. 
We applied a polynomial learning rate with an initial learning rate of $0.007$ or $0.01$.
\\ \indent We re-trained 102 models for this benchmark:
On ADE20K and the Cityscapes dataset, we re-trained 36 models each (six architectural ablations per network backbone; six network backbones in total). 
On PASCAL VOC, we re-trained 30 models, as we have evaluated on one network backbone less than on Cityscapes and ADE20K.

\subsection{Evaluation Metrics}
\label{sec:rebuttal_diff_metrics}
In the main paper, we predominantly use the Corruption Degradation (CD) to rate model robustness with respect to image corruptions, since the CD rates model robustness in terms of absolute performance.
The relative Corruption Degradation (rCD), on the other hand, incorporates the respective model performance on clean data.
The degradation on clean data is for both models (i.e., the model for which the robustness is to be rated, and the reference model) subtracted, resulting hence in a measure that gives a ratio of the absolute performance decrease in the presence of image corruption.

\subsection{Discussion of Architectural Properties}
\label{sec:rebuttal_theory}
As mentioned in the main paper, we now discuss possible causes of the robustness of architectural design choices with respect to image corruptions in more detail.
\\
\indent 
\textbf{Dense Prediction Cell.} 
As mentioned in the main paper, a model with Dense Prediction Cell (DPC) might learn less multi-scale representations than a model with the Atrous Spatial Pyramid Pooling (ASPP) module.
Whereas ASPP processes its input in parallel by three atrous convolution (AC) layers with large symmetric rates ($6$, $12$, $18$), DPC firstly processes the input by a single AC layer with small rate ($1 \times 6$)~\cite[Fig. 5]{chen_searching_2018}. 
We hypothesize that DPC might learn less multi-scale representations than ASPP, which may be useful for common image corruptions (e.g.,~\cite{geirhos_imagenet-trained_2019} shows that classification models are more robust to common corruption if the shape bias of a model is increased). 
When we test DPC on corrupted data, it cannot hence apply the same beneficial multi-scale cues (due to the comparable small atrous convolution with rate $1 \times 6$) as ASPP and may, therefore, perform worse. 
\\
\indent
\textbf{Global Average Pooling.} 
Global average pooling (GAP) increases performance on clean data on PASCAL VOC 2012, but not on the Cityscapes dataset or ADE20K.
GAP averages $2048$ activations of size $33 \times 33$ for our utilized training parameters.
A possible explanation for the effectiveness of GAP on PASCAL VOC 2012 might be, that the Cityscapes dataset and ADE20K consist of both a notably larger number and spatial distribution of instances per image.
Using GAP on these datasets might therefore not aid performance since important features may be lost due to averaging.
\\
\indent
\textbf{Long-Range Link.} 
Removing the Long-Range Link (LRL) discards early representations.
The degree of, \eg, image noise is more pronounced on early CNN levels.
Removing LRL tends hence to increase the robustness for a more shallow backbone as Xception-41 on PASCAL VOC 2012 and Cityscapes, as less corrupted features are linked from encoder to decoder.
For a deeper backbone like ResNet-101, this behavior cannot be observed. 

\subsection{Qualitative Results}
\label{sec:qualitativeresults}
\begin{figure*}[h]
\centering
    \begin{subfigure}[t!]{0.25\linewidth}
		\includegraphics[width=1\linewidth]{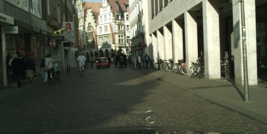}
		\caption{Blurred validation image}
	\end{subfigure}~
	\begin{subfigure}[t!]{0.25\linewidth}
		\centering
		\includegraphics[width=1\linewidth]{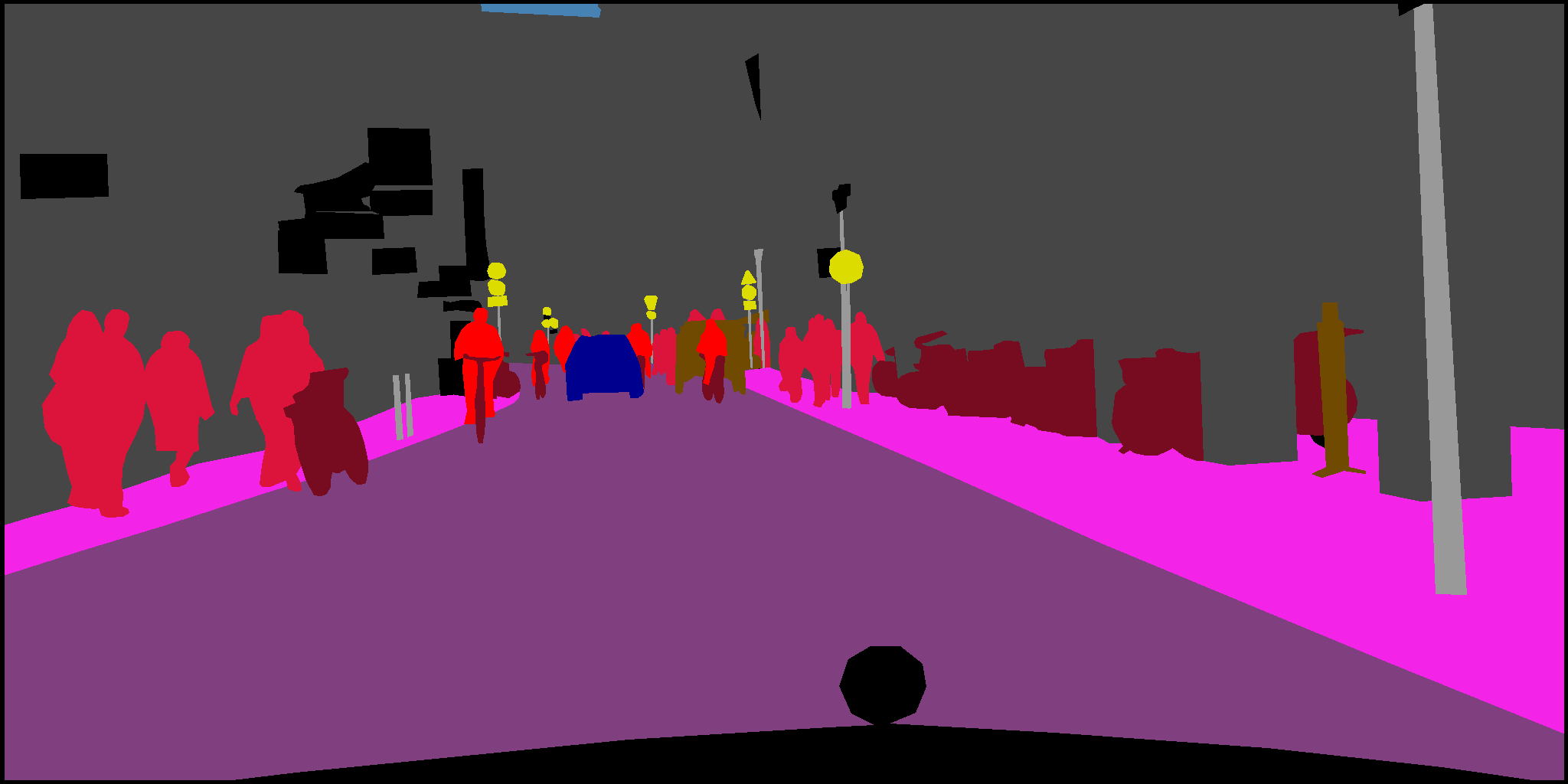}
		\caption{ground truth (gt)}
	\end{subfigure}~
	\begin{subfigure}[t!]{0.25\linewidth}
		\centering
		\includegraphics[width=1\linewidth]{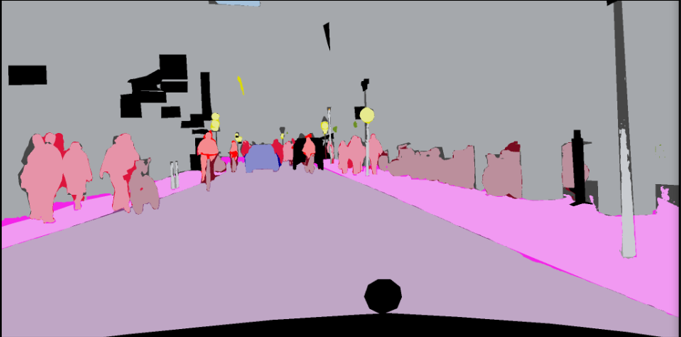}
		\caption{Overlay clean estimate + gt}
	\end{subfigure}~
	\begin{subfigure}[t!]{0.25\linewidth}
		\centering
		\includegraphics[width=1\linewidth]{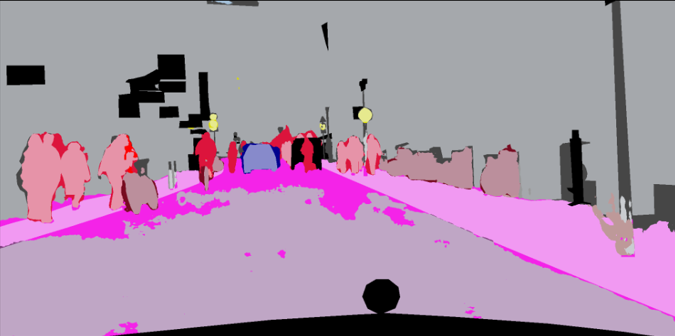}
		\caption{Overlay blurred estimate + gt}
	\end{subfigure}
	\caption{Prediction of the reference architecture (\ie original DeepLabv3$+$) on blurred input, using Xception-71 as network backbone. 
	(a) A blurred validation image of the Cityscapes dataset and corresponding ground truth (b). 
	(c) Prediction on the  clean image overlaid with the ground truth. True-positives are alpha-blended, false-positives and false-negatives remain unchanged. Hence, wrongly classified pixels can be easier spotted. 
	(d) Prediction on the blurred image overlaid with the ground truth (b). 
	Whereas the \textit{riders} are mostly correctly classified in (c), they are in (d) miss-classified as \textit{person}. 
	Extensive areas of \textit{road} are miss-classified as \textit{sidewalk}.\vspace{-0.2cm}}
	\label{fig:blur_missclassify}	
\end{figure*}

We provide qualitative results in this subsection. 
As mentioned in the main paper, blurred images cause the models to miss-classify pixels of classes covering small image regions, especially when far away. 
Please see Figure~\ref{fig:blur_missclassify} for an example. 
(a) A blurred validation image of the Cityscapes dataset and the corresponding ground truth in (b). 
(c) The prediction on the clean image overlaid with the ground truth (b). 
In this visualization, true-positives are alpha-blended, and false-positives, as well as false-negatives, remain unchanged. 
Hence, wrongly classified pixels can be easier identified. 
(d) The prediction on the blurred image overlaid with the ground truth (b).
Whereas the \textit{riders} and \textit{persons} are mostly correctly classified in (c), \textit{riders} in (d) are miss-classified as \textit{persons}, and extensive areas of \textit{road} are miss-classified as \textit{sidewalk}. 
We used the reference model along with Xception-71 as network backbone to produce these predictions.
\begin{figure*}[h]
\centering
    \begin{subfigure}[t!]{0.25\linewidth}
		\includegraphics[width=1\linewidth]{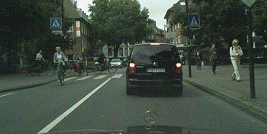}
	\caption{Corrupted validation image}
	\end{subfigure}~
	\begin{subfigure}[t!]{0.25\linewidth}
		\centering
		\includegraphics[width=1\linewidth]{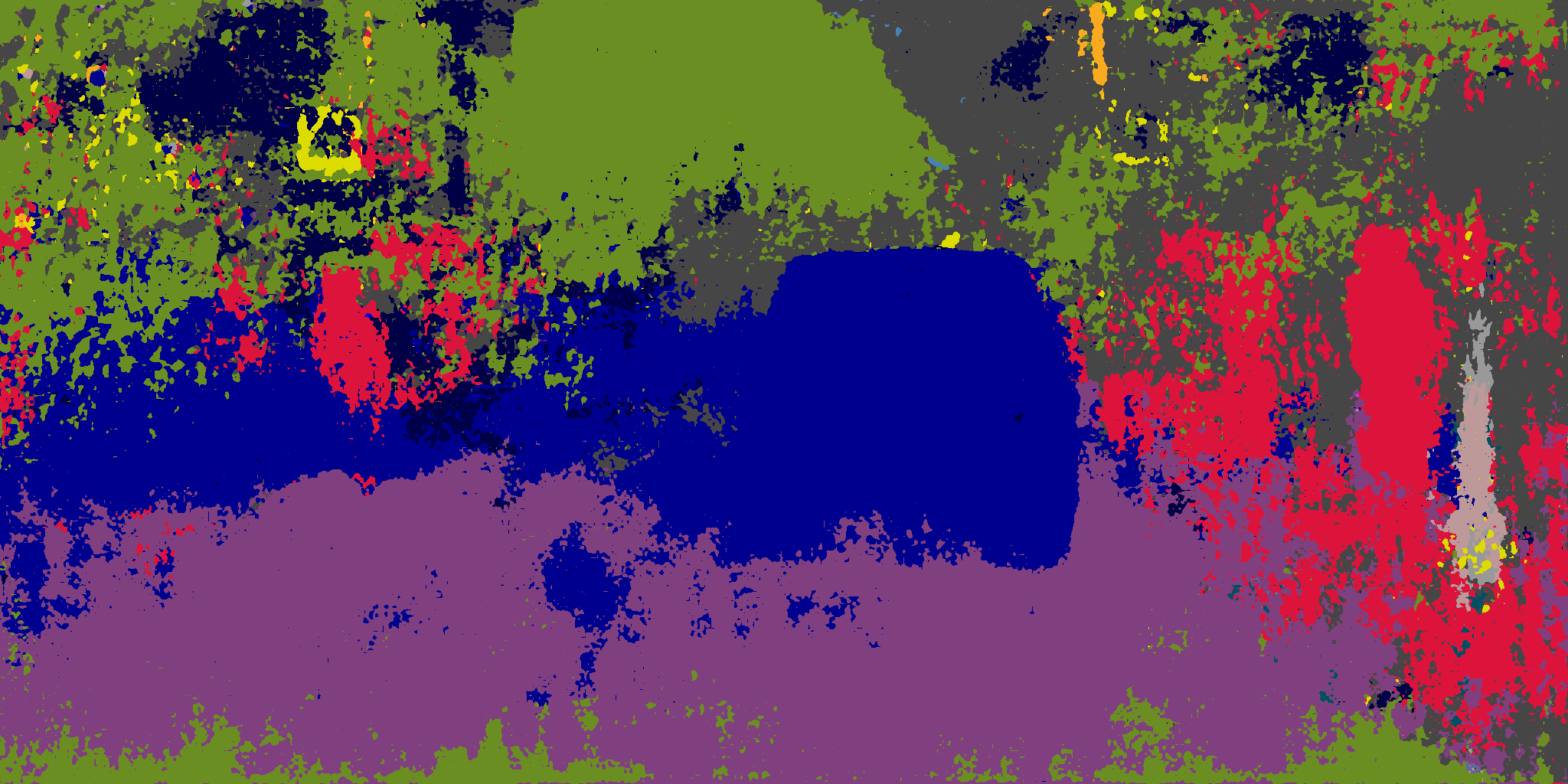}
	\caption{Prediction on (a)}
	\end{subfigure}~
	\begin{subfigure}[t!]{0.25\linewidth}
		\centering
		\includegraphics[width=1\linewidth]{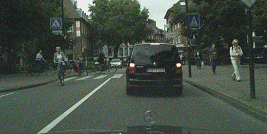}
		\caption{Corrupted validation image}
	\end{subfigure}~
	\begin{subfigure}[t!]{0.25\linewidth}
		\centering
		\includegraphics[width=1\linewidth]{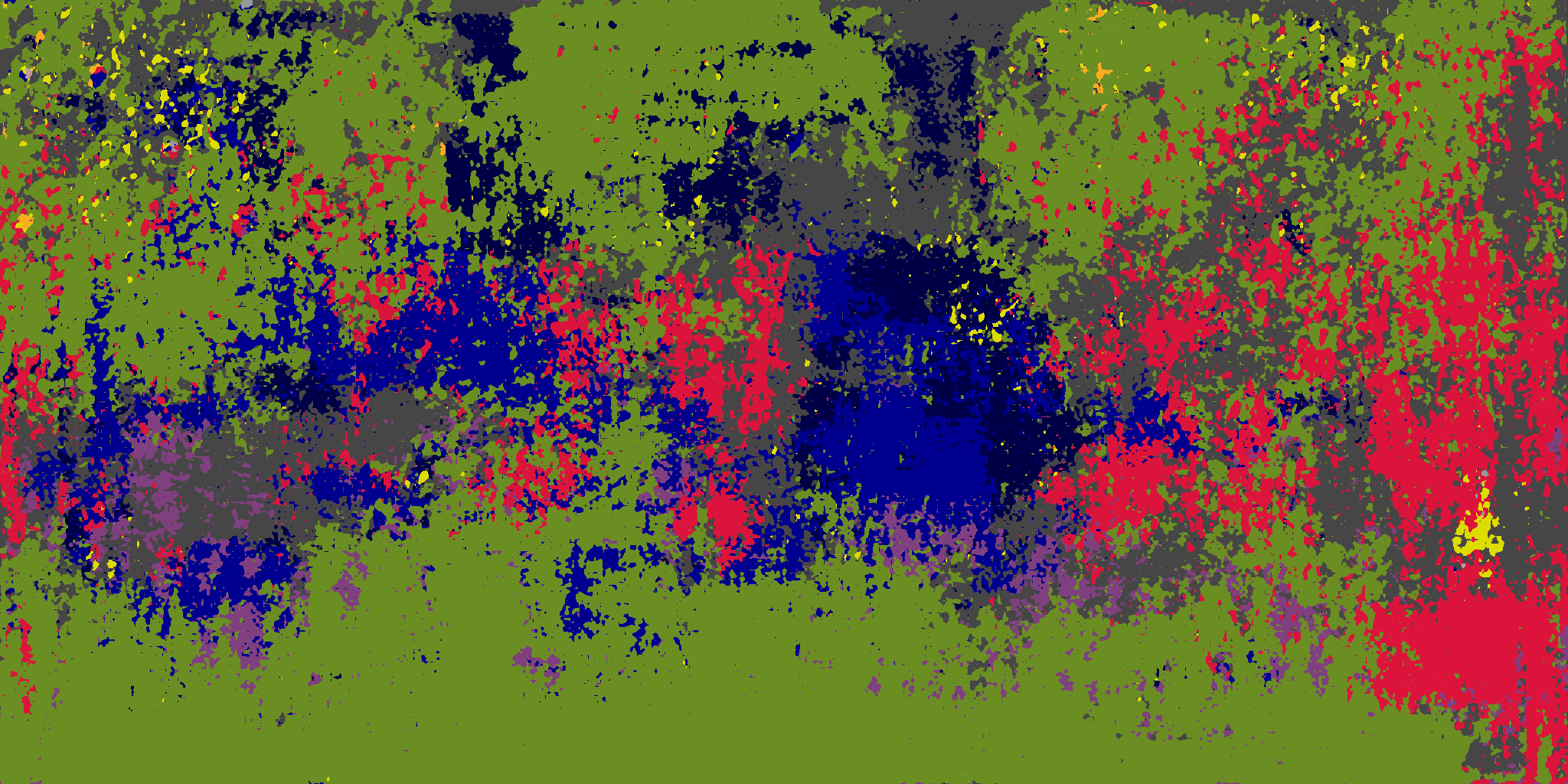}
		\caption{Prediction on (c)}
	\end{subfigure}
	\caption{
		Drastic influence of image noise on model performance. 
		(a) A validation image of Cityscapes is corrupted by the second severity level of Gaussian noise and respective prediction (b). 
		(c) A validation image of Cityscapes is corrupted by the third severity level of Gaussian Noise and respective prediction (d). 
		Predictions are produced by the reference model, using Xception-71 as the backbone.\vspace{-0.2cm}}
	\label{fig:noise_influence}	
\end{figure*}
\begin{figure*}[h!]
\centering
    \begin{subfigure}[t!]{0.25\linewidth}
		\includegraphics[width=1\linewidth]{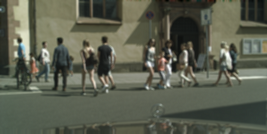}
   		\caption{corrupted image}
	\end{subfigure}~
	\begin{subfigure}[t!]{0.25\linewidth}
		\centering
		\includegraphics[width=1\linewidth]{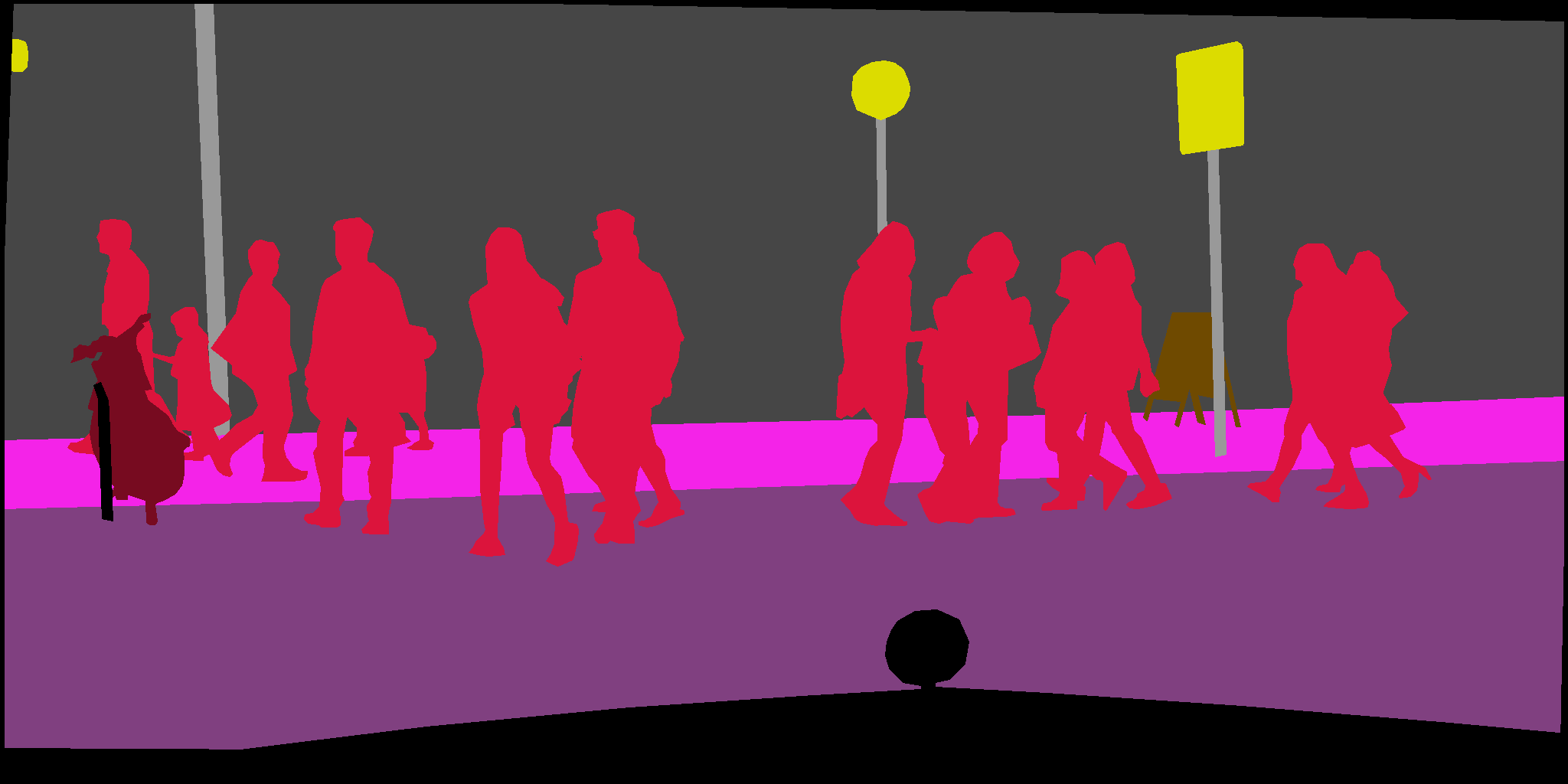}
		\caption{ground truth}
	\end{subfigure}~
	\begin{subfigure}[t!]{0.25\linewidth}
		\centering
		\includegraphics[width=1\linewidth]{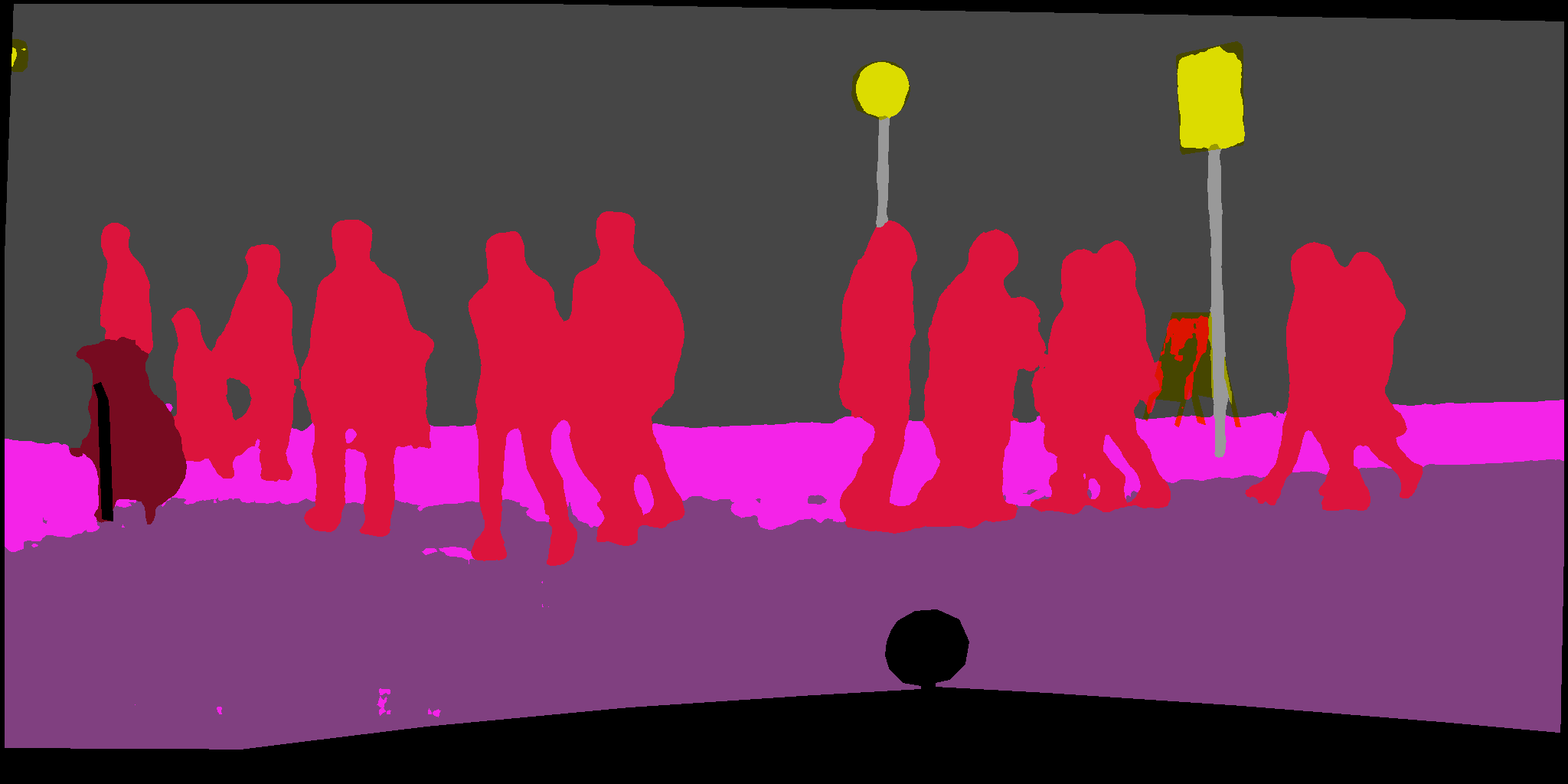}
		\caption{prediction of ref. model}
	\end{subfigure}~
	\begin{subfigure}[t!]{0.25\linewidth}
		\centering
		\includegraphics[width=1\linewidth]{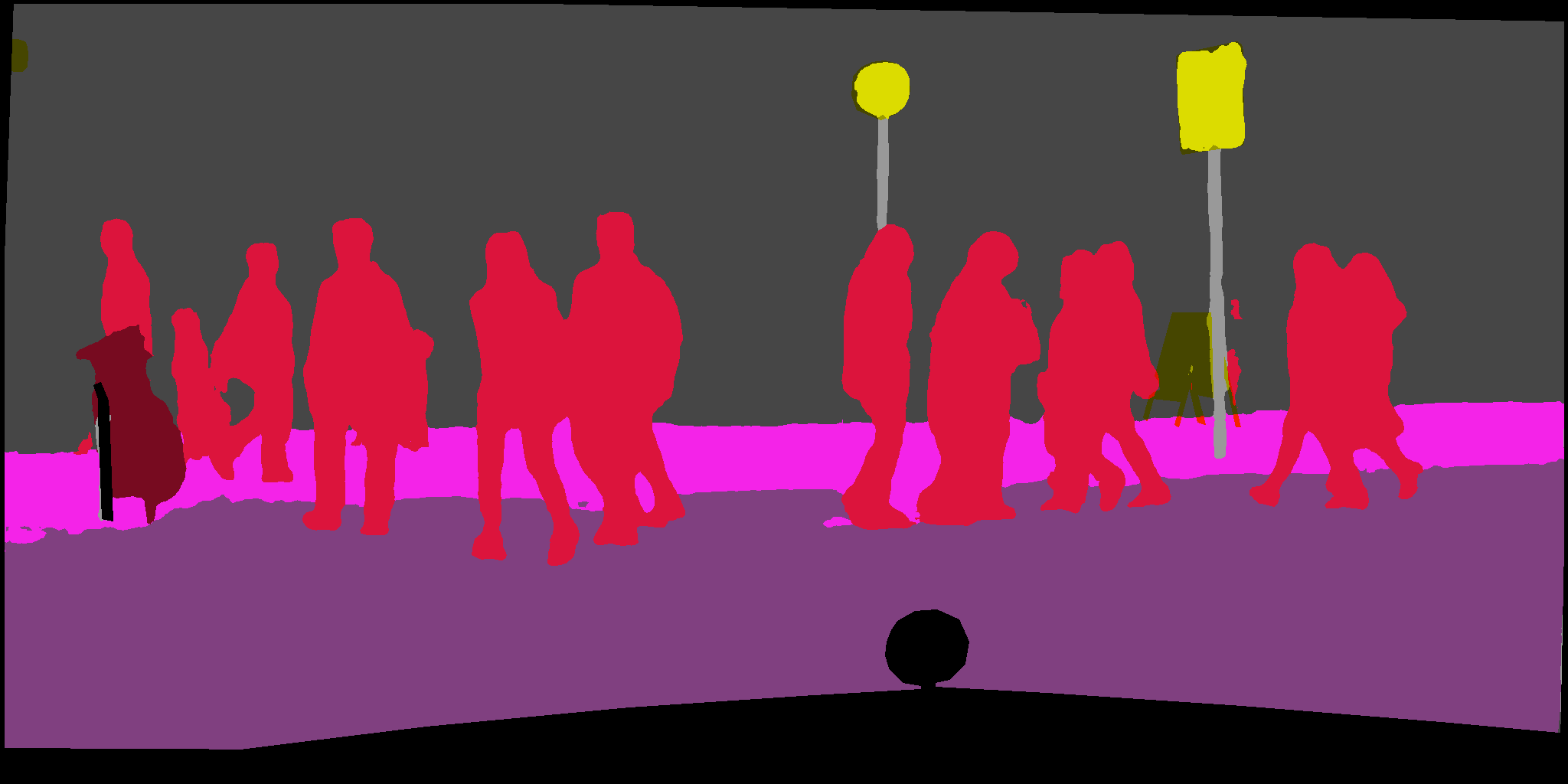}
    	\caption{prediction w/o ASPP}
	\end{subfigure}
	
    \begin{subfigure}[t!]{0.25\linewidth}
		\includegraphics[width=1\linewidth]{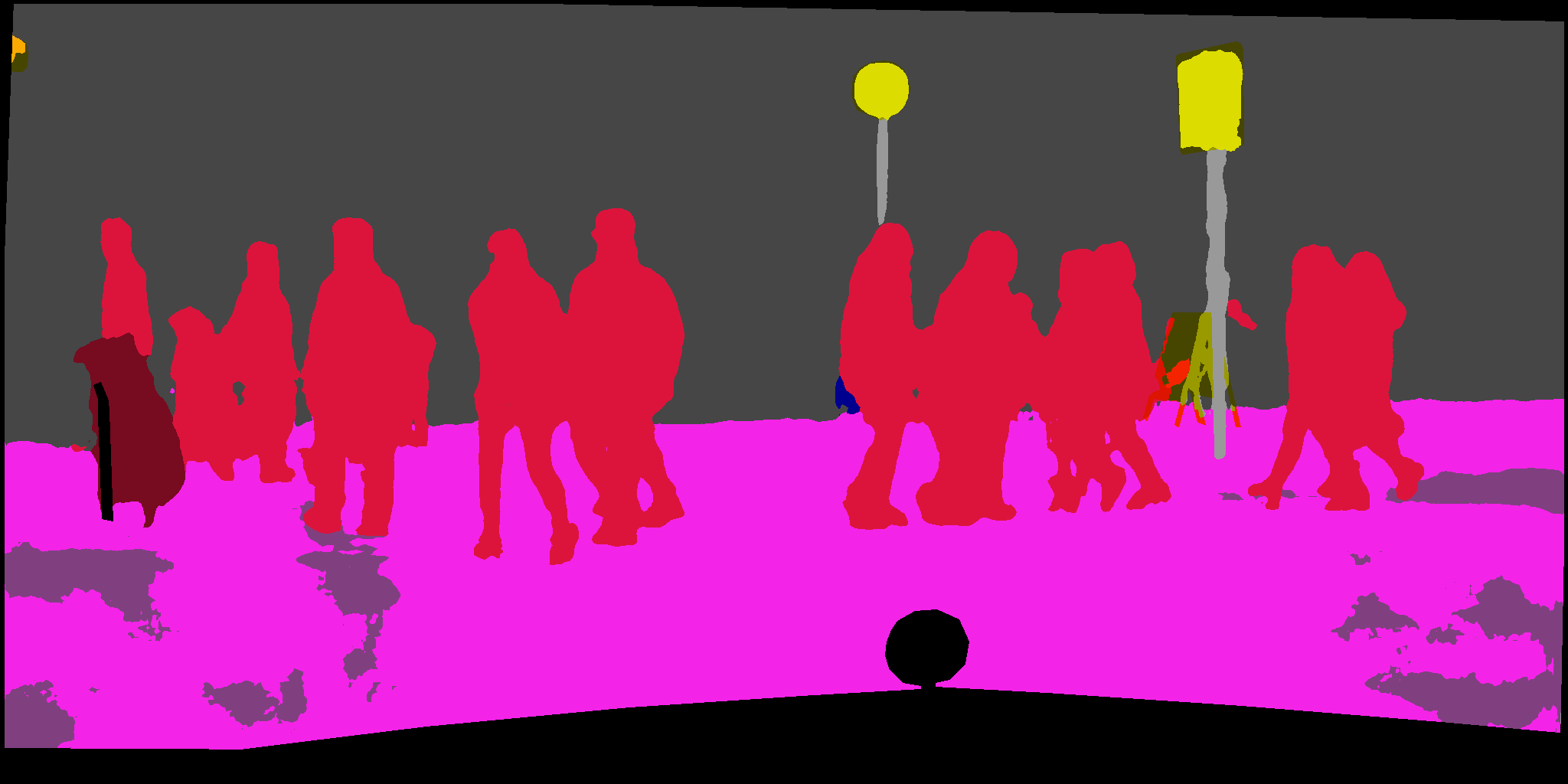}
		\caption{prediction w/o AC}
	\end{subfigure}~
	\begin{subfigure}[t!]{0.25\linewidth}
		\centering
		\includegraphics[width=1\linewidth]{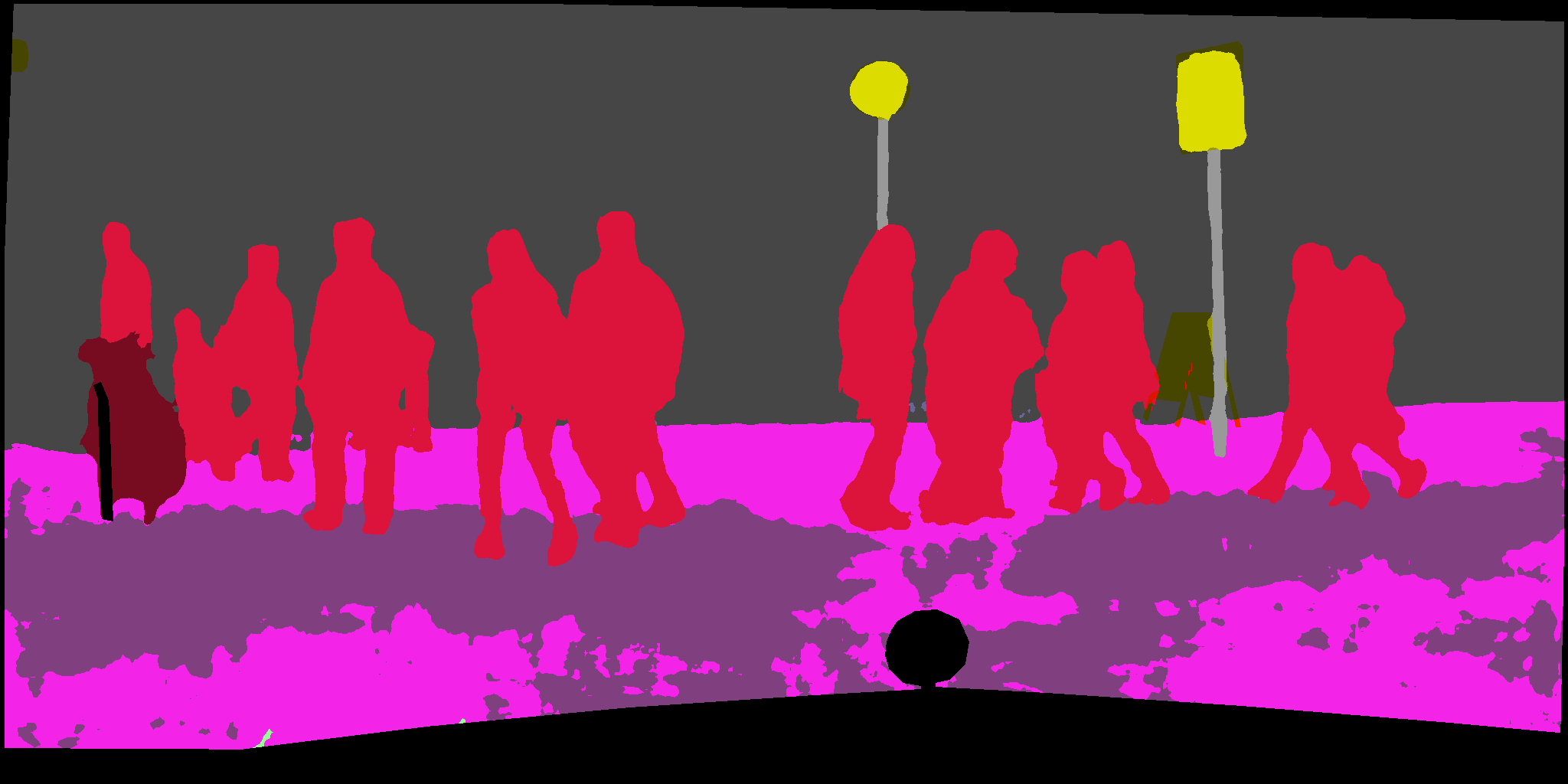}
		\caption{prediction w/ DPC}
	\end{subfigure}~
	\begin{subfigure}[t!]{0.25\linewidth}
		\centering
		\includegraphics[width=1\linewidth]{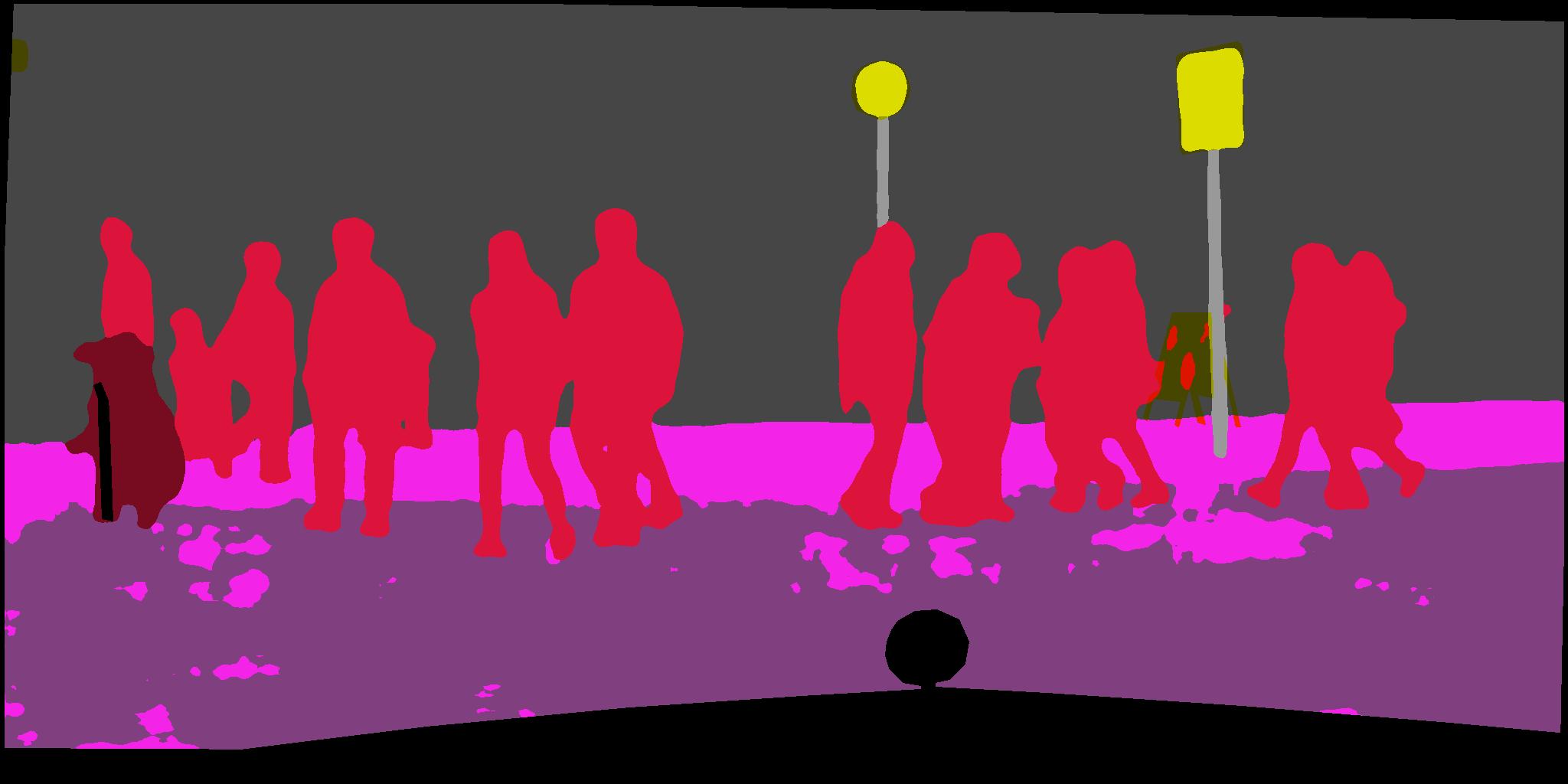}
		\caption{prediction w/o LRL}
	\end{subfigure}~
	\begin{subfigure}[t!]{0.25\linewidth}
		\centering
		\includegraphics[width=1\linewidth]{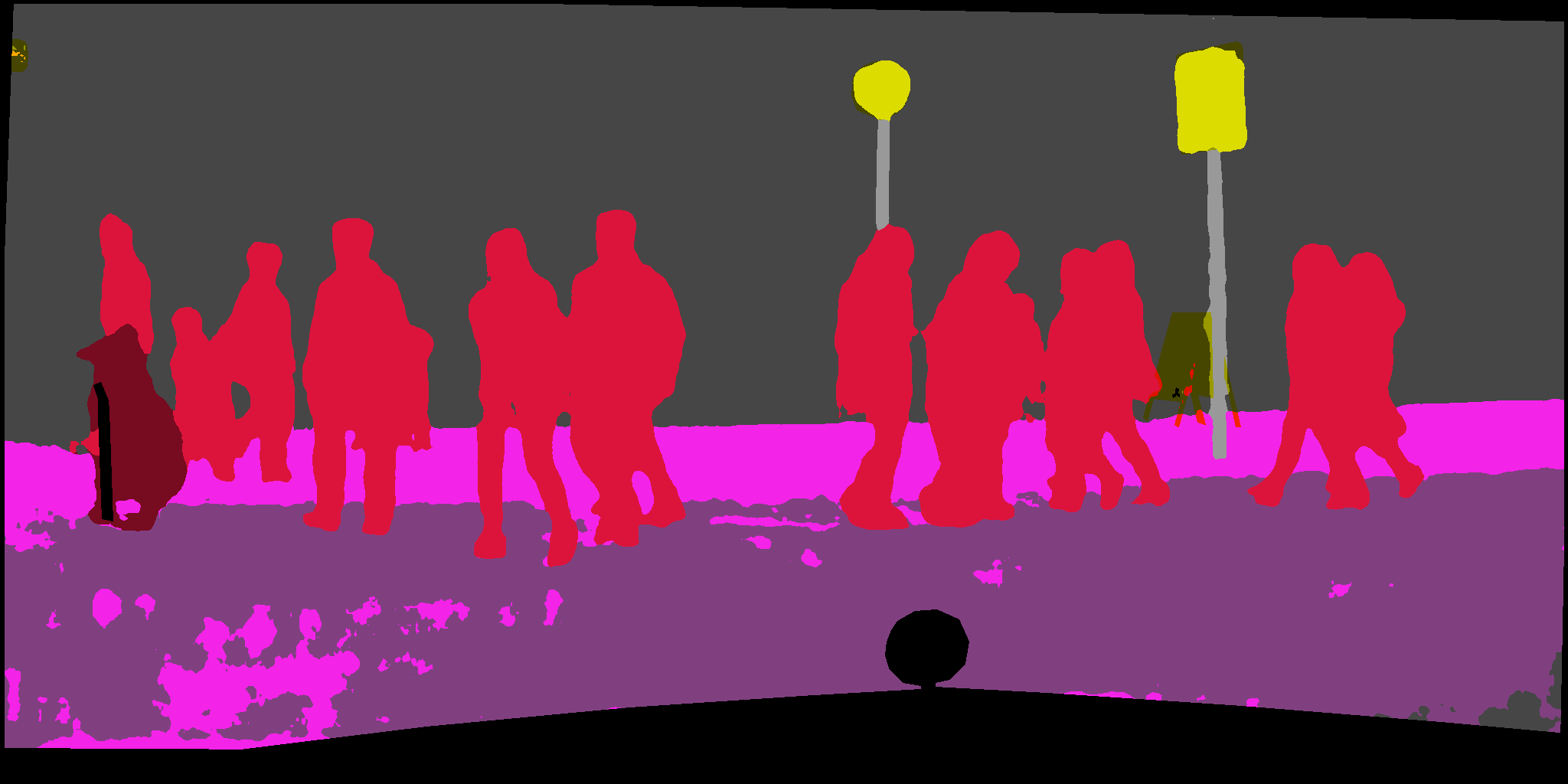}
		\caption{prediction w/ GAP}
	\end{subfigure}~

	\caption{Predictions of reference architecture and ablations on a blurred image. The ablated variants w/o AC, and w/ DPC are especially vulnerable to blur.\vspace{-0.2cm}}
	\label{fig:ex1_blur}	
\end{figure*}
\begin{figure*}[h!]
\centering
    \begin{subfigure}[t!]{0.25\linewidth}
		\includegraphics[width=1\linewidth]{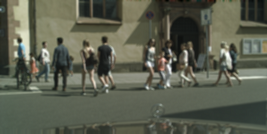}
		\caption{corrupted image}
	\end{subfigure}~
	\begin{subfigure}[t!]{0.25\linewidth}
		\centering
		\includegraphics[width=1\linewidth]{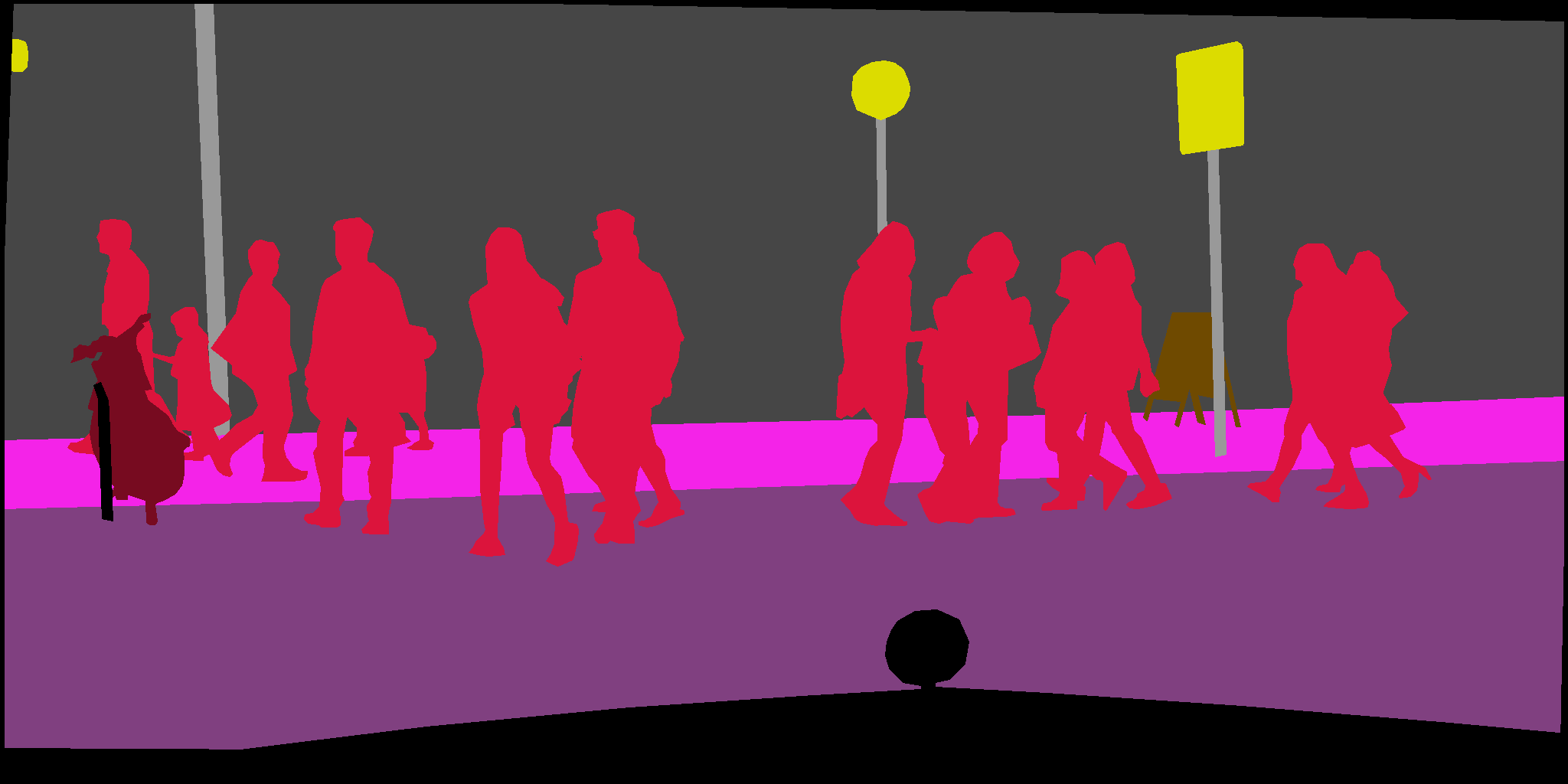}
		\caption{ground truth}
	\end{subfigure}~
	\begin{subfigure}[t!]{0.25\linewidth}
		\centering
		\includegraphics[width=1\linewidth]{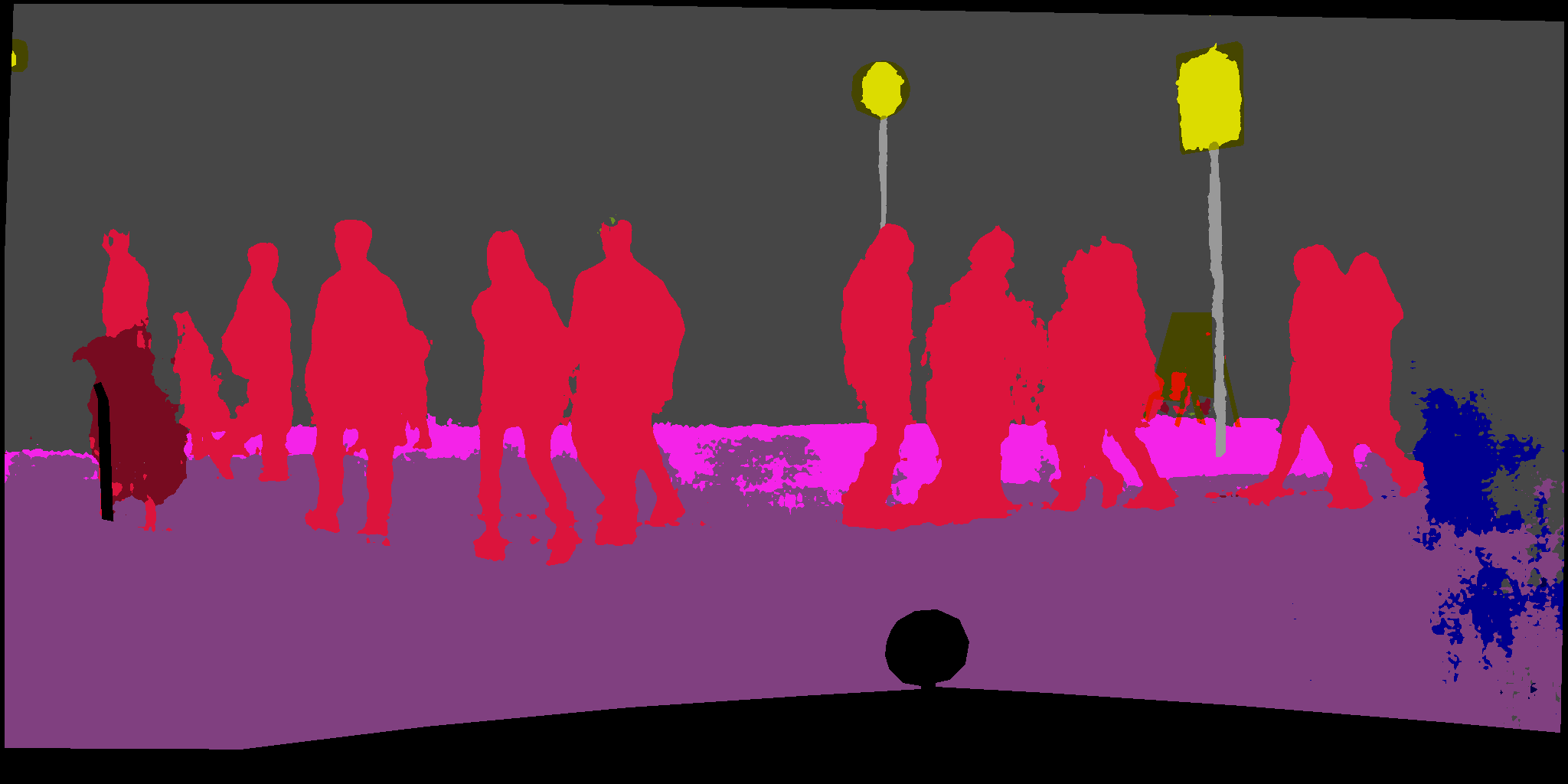}
		\caption{prediction of ref. model}
	\end{subfigure}~
	\begin{subfigure}[t!]{0.25\linewidth}
		\centering
		\includegraphics[width=1\linewidth]{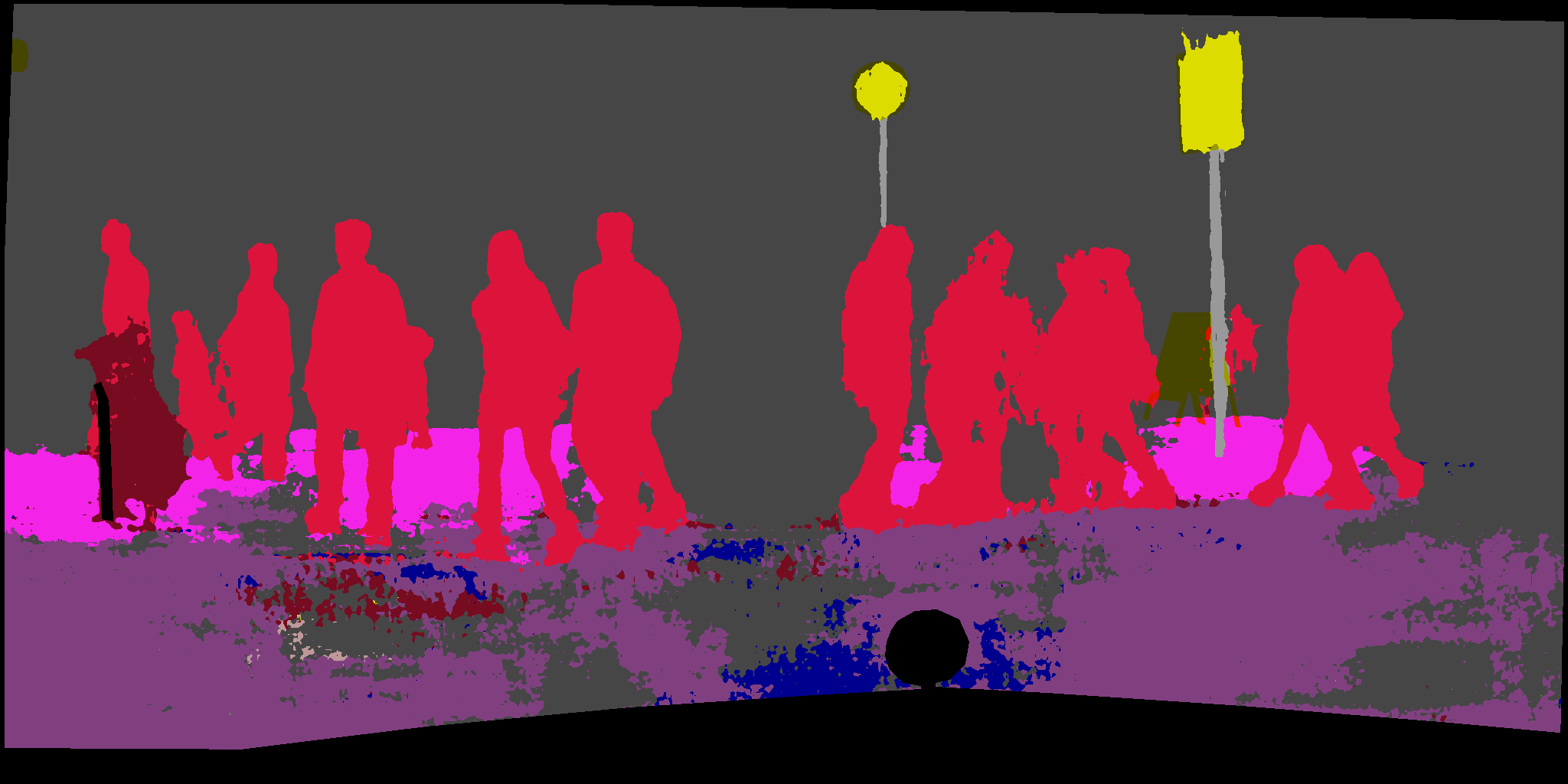}
		\caption{prediction w/o ASPP}
	\end{subfigure}
	
    \begin{subfigure}[t!]{0.25\linewidth}
		\includegraphics[width=1\linewidth]{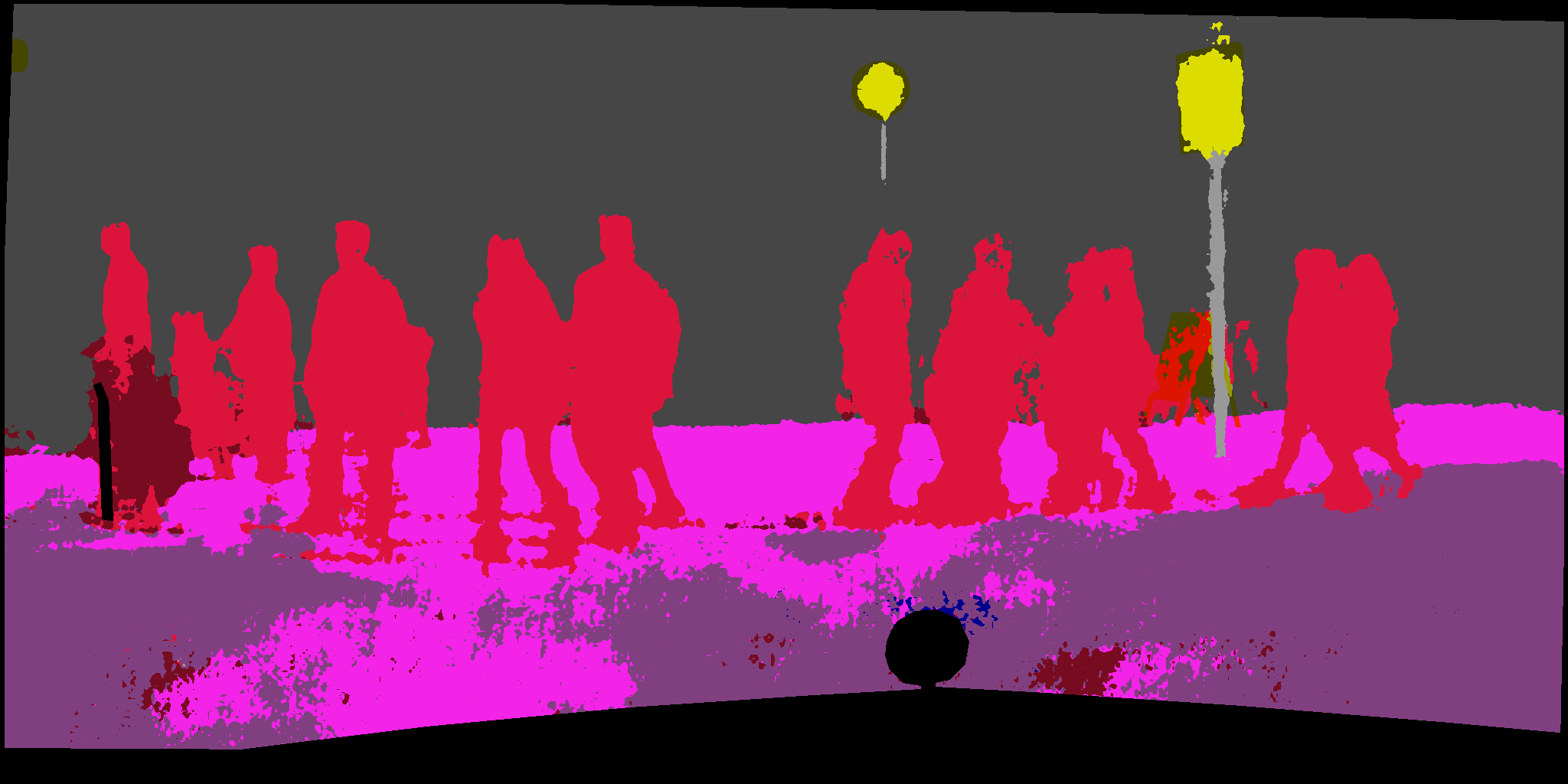}
		\caption{prediction w/o AC}
	\end{subfigure}~
	\begin{subfigure}[t!]{0.25\linewidth}
		\centering
		\includegraphics[width=1\linewidth]{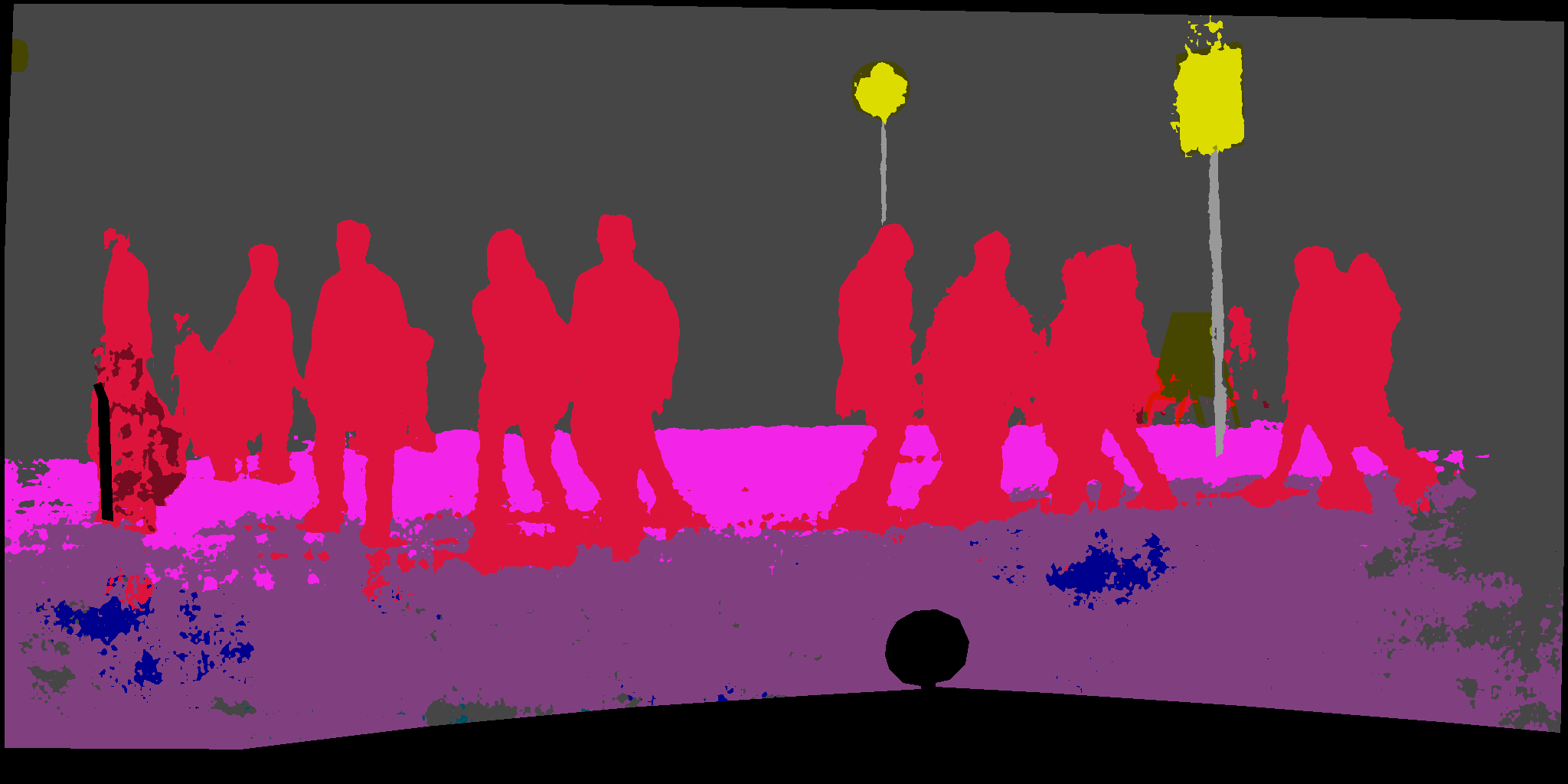}
		\caption{prediction w/ DPC}
	\end{subfigure}~
	\begin{subfigure}[t!]{0.25\linewidth}
		\centering
		\includegraphics[width=1\linewidth]{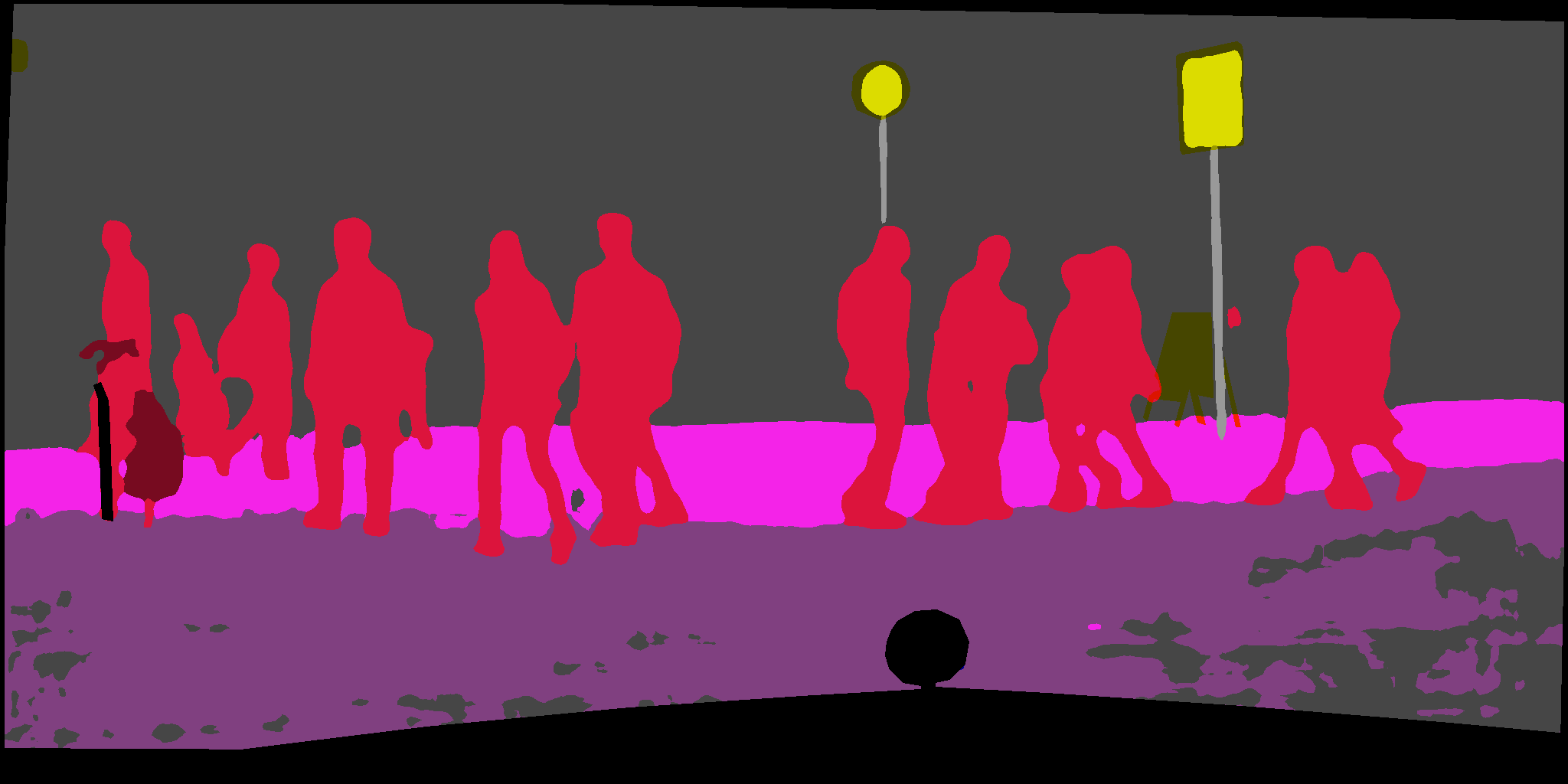}
		\caption{prediction w/o LRL}
	\end{subfigure}~
	\begin{subfigure}[t!]{0.25\linewidth}
		\centering
		\includegraphics[width=1\linewidth]{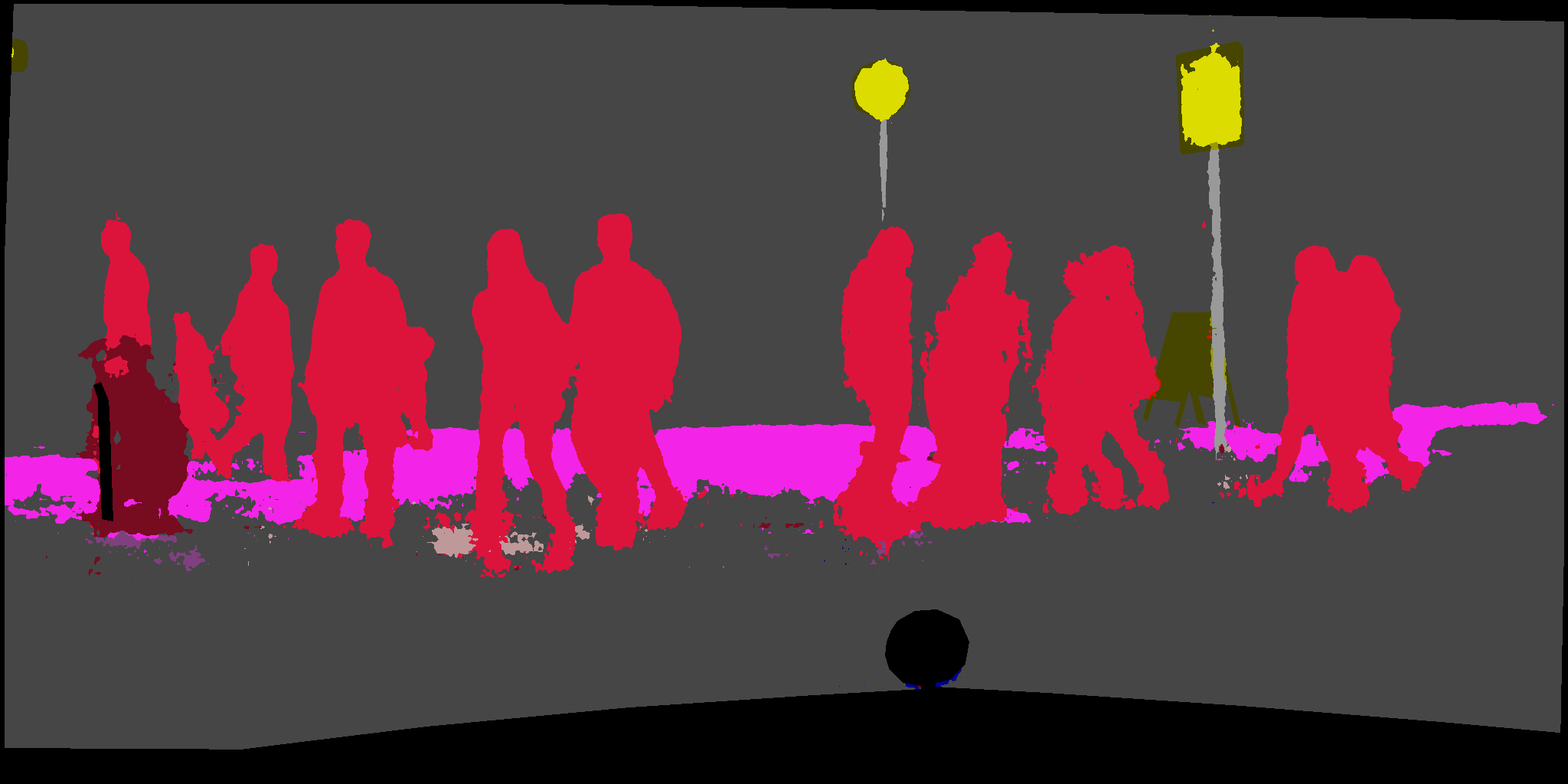}
		\caption{prediction w/ GAP}
	\end{subfigure}~
	\caption{Predictions of reference architecture and ablations on a noisy image. The ablated variants w/o AC, ASPP, and w/ DPC, GAP are especially vulnerable.\vspace{-0.2cm}}
	\label{fig:ex2_noise}	
\end{figure*}
\begin{figure*}[h!]
\vspace{+4.5cm}
\centering
    \begin{subfigure}[t!]{0.20\linewidth}
		\includegraphics[width=1\linewidth]{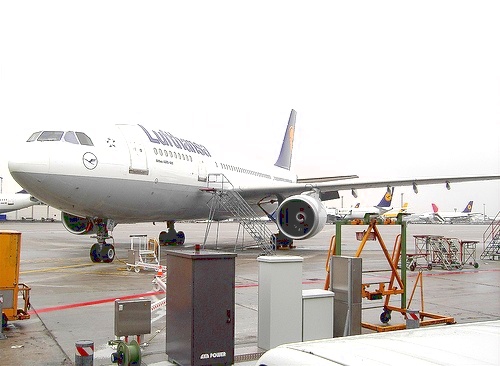}
		\caption{corrupted image}
	\end{subfigure}~
	\begin{subfigure}[t!]{0.20\linewidth}
		\centering
		\includegraphics[width=1\linewidth]{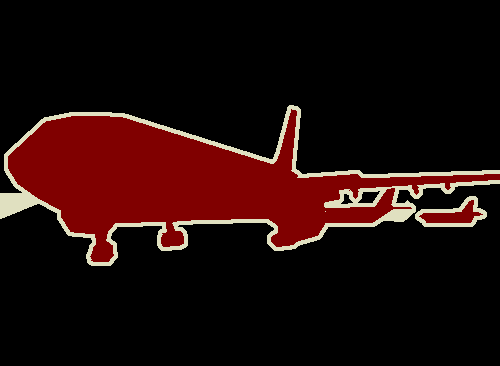}
		\caption{ground truth}
	\end{subfigure}~
	\begin{subfigure}[t!]{0.20\linewidth}
		\centering
		\includegraphics[width=1\linewidth]{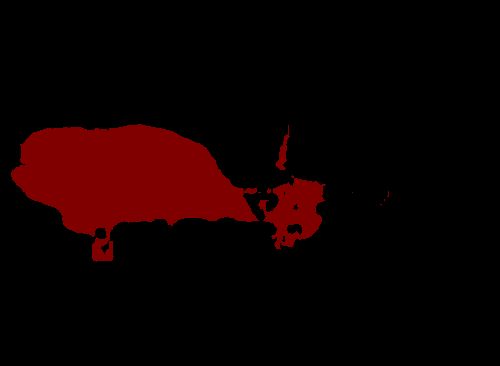}
		\caption{prediction of ref. model}
	\end{subfigure}~
	\begin{subfigure}[t!]{0.20\linewidth}
		\centering
		\includegraphics[width=1\linewidth]{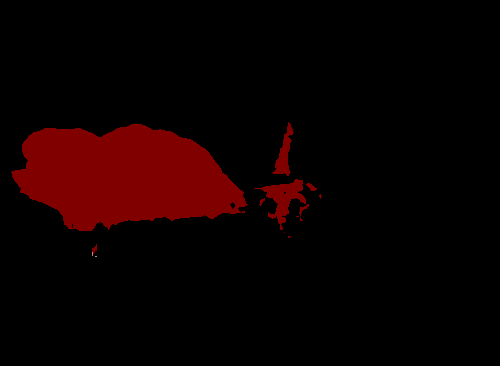}
		\caption{prediction w/o ASPP}
	\end{subfigure}
	
    \begin{subfigure}[t!]{0.20\linewidth}
		\includegraphics[width=1\linewidth]{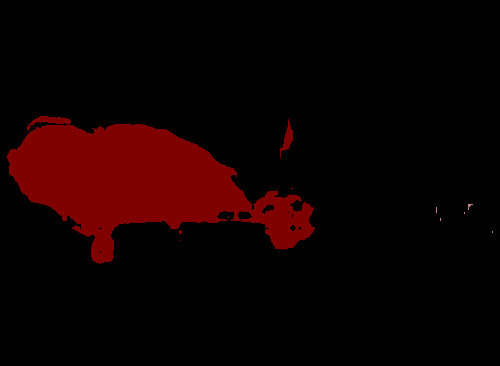}
		\caption{prediction w/o AC}
	\end{subfigure}~
	\begin{subfigure}[t!]{0.20\linewidth}
		\centering
		\includegraphics[width=1\linewidth]{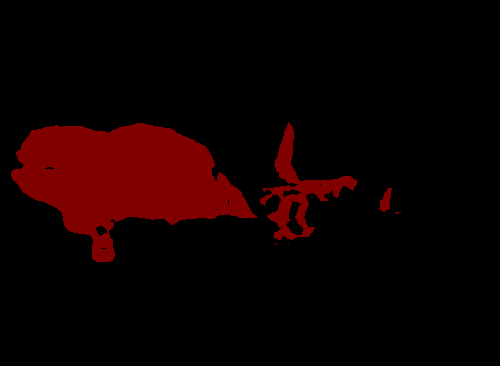}
		\caption{prediction w/ DPC}
	\end{subfigure}~
	\begin{subfigure}[t!]{0.20\linewidth}
		\centering
		\includegraphics[width=1\linewidth]{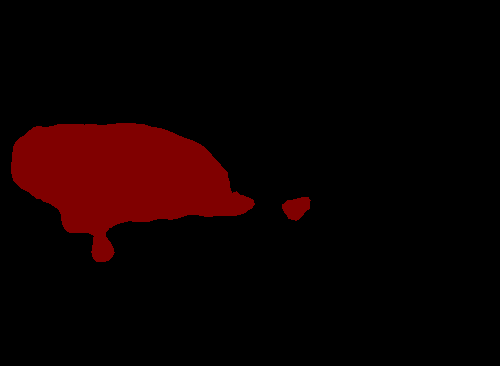}
		\caption{prediction w/o LRL}
	\end{subfigure}~
	\begin{subfigure}[t!]{0.20\linewidth}
		\centering
		\includegraphics[width=1\linewidth]{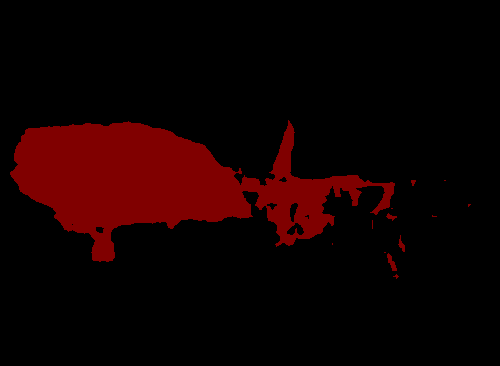}
		\caption{prediction w/ GAP}
	\end{subfigure}~
	\caption{Predictions of reference architecture and ablations on a validation image of PASCAL VOC 2012, corrupted by \textit{brightness}.}
	\label{fig:ex3_digital}	
\end{figure*}
\begin{figure*}[h!]
\centering
    \begin{subfigure}[t!]{0.20\linewidth}
		\includegraphics[width=1\linewidth]{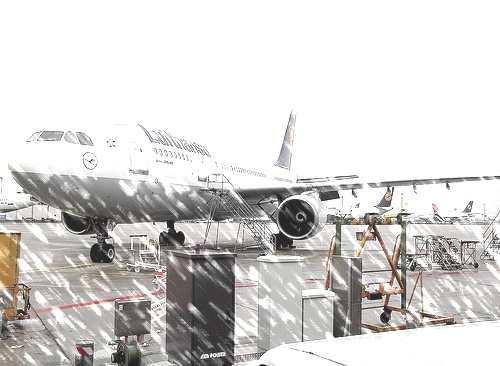}
		\caption{corrupted image}
	\end{subfigure}~
	\begin{subfigure}[t!]{0.20\linewidth}
		\centering
		\includegraphics[width=1\linewidth]{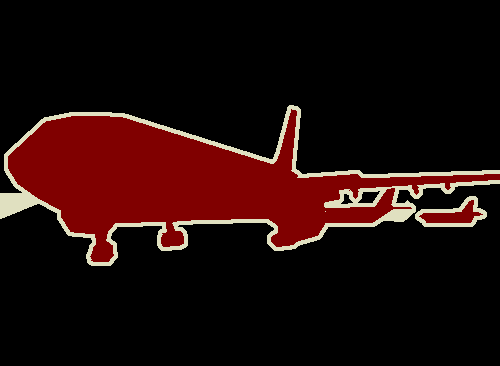}
		\caption{ground truth}
	\end{subfigure}~
	\begin{subfigure}[t!]{0.20\linewidth}
		\centering
		\includegraphics[width=1\linewidth]{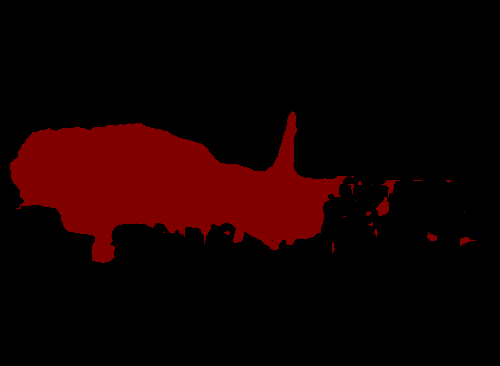}
		\caption{prediction of ref. model}
	\end{subfigure}~
	\begin{subfigure}[t!]{0.20\linewidth}
		\centering
		\includegraphics[width=1\linewidth]{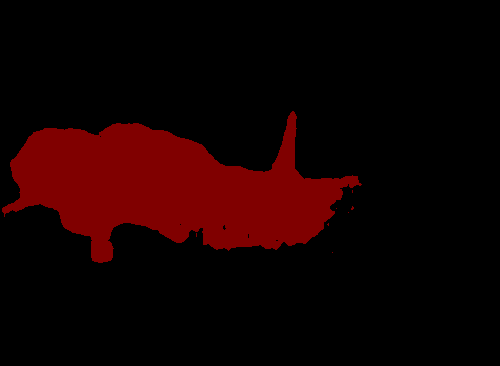}
		\caption{prediction w/o ASPP}
	\end{subfigure}
	
    \begin{subfigure}[t!]{0.20\linewidth}
		\includegraphics[width=1\linewidth]{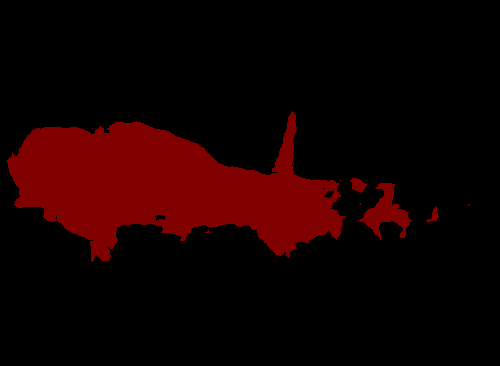}
		\caption{prediction w/o AC}
	\end{subfigure}~
	\begin{subfigure}[t!]{0.20\linewidth}
		\centering
		\includegraphics[width=1\linewidth]{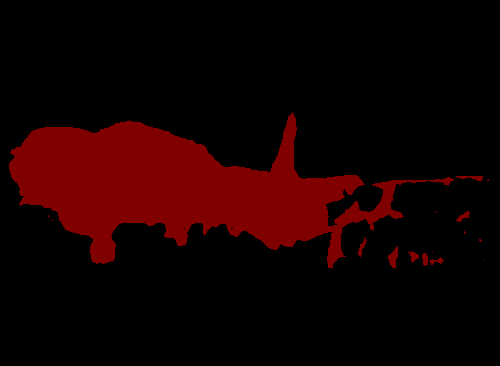}
		\caption{prediction w/ DPC}
	\end{subfigure}~
	\begin{subfigure}[t!]{0.20\linewidth}
		\centering
		\includegraphics[width=1\linewidth]{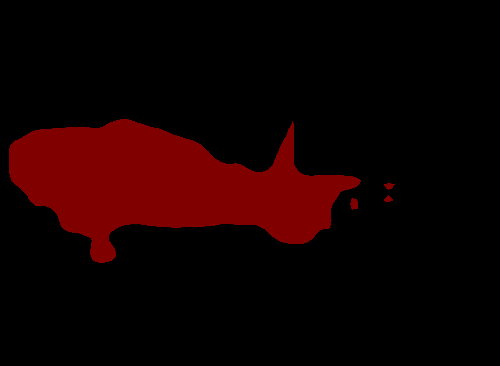}
		\caption{prediction w/o LRL}
	\end{subfigure}~
	\begin{subfigure}[t!]{0.20\linewidth}
		\centering
		\includegraphics[width=1\linewidth]{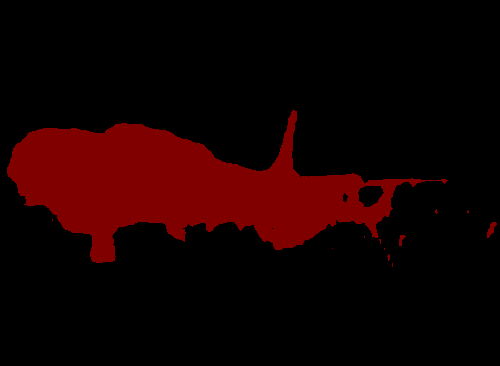}
		\caption{prediction w/ GAP}
	\end{subfigure}~

	\caption{Predictions of the reference architecture and ablations on a validation image of PASCAL VOC 2012, corrupted by \textit{snow}.\vspace{+3cm}}
	\label{fig:ex4_weather}	
\end{figure*}
\\
\indent We report in the main paper, that--of the utilized image corruptions in this work--image noise affects model performance the most, as pointed out using mIoU scores.
To give a visual example, we selected two noisy variants of a validation image of the Cityscapes dataset and show the predictions of the reference architecture using Xception-71 as network backbone in Figure~\ref{fig:noise_influence}. 
The mIoU for both predictions is less than \SI{15}{\%}.
\\ \indent Finally, we show qualitative results of every ablated architecture for one image corruption of category blur, noise, digital, and weather. 
Figure~\ref{fig:ex1_blur} shows a blurred validation image of the Cityscapes dataset and the corresponding predictions. 
Note that the ablated variants w/o AC and w/ DPC are especially vulnerable. 
Figure~\ref{fig:ex2_noise} shows a noisy validation image of the Cityscapes dataset. 
Note that the ablated variants w/o AC, w/o ASPP and w/ DPC are especially vulnerable. 
Figure~\ref{fig:ex3_digital} and Figure~\ref{fig:ex4_weather} show a validation image of PASCAL VOC 2012, corrupted by \textit{brightness} and \textit{snow}, respectively. 

\subsection{Experimental Results on Cityscapes}
\label{sec:experimentalresults_cityscapes}
\begin{figure*}[h]
	\centering
	\begin{subfigure}[t]{0.50\linewidth}
		\centering
		\includegraphics[width=1\linewidth]{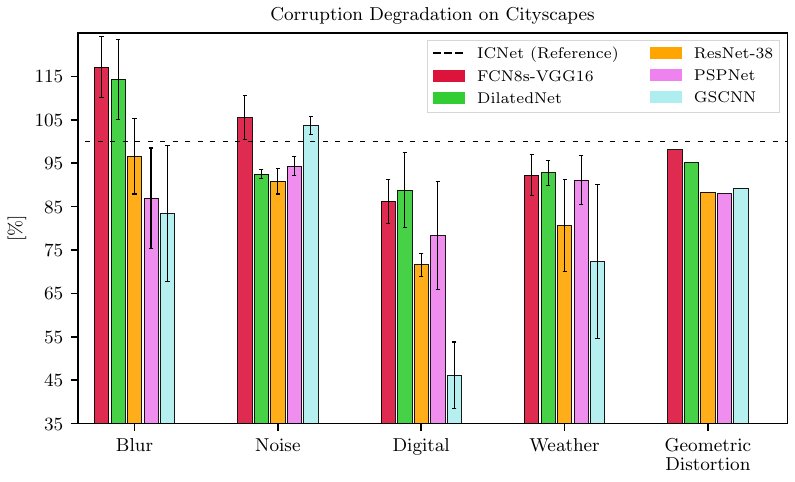}
	\end{subfigure}~
	\begin{subfigure}[t]{0.50\linewidth}
		\centering
		\includegraphics[width=1\linewidth]{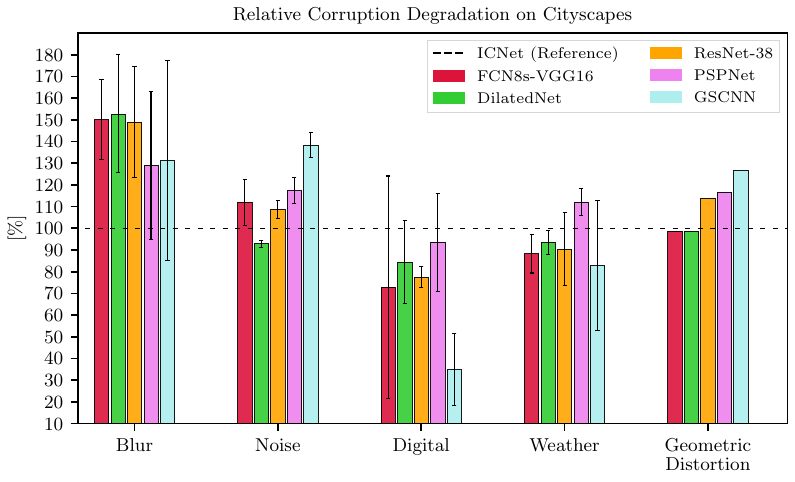}
	\end{subfigure}~
	\caption{
CD (left) and rCD (right) evaluated on Cityscapes for ICNet (set as reference architecture), FCN8s-VGG16, DilatedNet, ResNet-38, PSPNet, GSCNN w.r.t. image corruptions of category blur, noise, digital, weather, and geometric distortion. 
Each bar except for geometric distortion is averaged within a corruption category (error bars indicate the standard deviation).
The CD of image corruption ``jpeg compression'' of category digital is not included in this barplot, since, contrary to the remaining image corruptions of that category, the respective CDs range between \SI{107}{\%} and \SI{133}{\%}.
Bars above \SI{100}{\%} represent a decrease in performance compared to the reference architecture.
FCN8s-VGG16 and DilatedNet are vulnerable to corruptions of category blur.
DilatedNet is more robust against corruptions of category noise, digital, and weather than the reference.
ResNet-38 is robust against corruptions of category weather.
The rCD of PSPNet is oftentimes higher than \SI{100}{\%} for each category.
GSCNN is vulnerable to image noise.
The rCD is considerably high, indicating a high decrease of mIoU in the presence of this corruption. 
Best viewed in color.
}
\label{fig:CD_rCD_nondeeplab}
\end{figure*}

In this section, we provide a more detailed analysis of both the non-Deeplabv3$+$ and Deeplabv3$+$ based segmentation models.
Figure \ref{fig:CD_rCD_nondeeplab} illustrates the CD and rCD averaged for the proposed image corruption categories.
Please note that the CD of image corruption ``jpeg compression'' of category digital is not included in this barplot. 
Contrary to the remaining image corruptions of that category, the respective CDs are considerably high (see Tab. ~\ref{tab:miou_cs_nondeeplab}).
FCN8s-VGG16 and DilatedNet are vulnerable to blur. 
The CD of \textit{defocus blur} \SI{124}{\%} and \SI{115}{\%}, respectively.
However, DilatedNet is more robust against corruptions of category noise, digital, and weather than ICNet.
For example, the CD of \textit{intensity noise} is \SI{92}{\%}.
ResNet-38 is robust against corruptions of category weather. 
Both the CD and rCD for \textit{fog} are roughly \SI{65}{\%}.
The CD of PSPNet is oftentimes less than \SI{100}{\%} (see Table ~\ref{tab:miou_cs_nondeeplab}).
GSCNN performs very well on digital corruptions (except \textit{JPEG compression}) and weather corruptions, especially \textit{fog} (CD is \SI{44}{\%}, rCD is \SI{34}{\%}.).
The model is, however, vulnerable to image noise, as CD and rCD are always higher than \SI{100}{\%}.
We list the individual CD and rCD scores in Table~\ref{tab:miou_cs_nondeeplab}.
Please find the absolute mIoU values in Table 1 (bottom) in the main paper.
\\ 
\indent
Table~\ref{tab:miou_cs_allbackbones_and_ablations} contains the mIoU for clean and corrupted variants of the validation set of the Cityscapes dataset for several network backbones of the DeepLabv3+ architecture and each respective architectural modification. 
In addition to the main paper, where we discuss the CD score (see Figure 5 in the main paper), Figure \ref{fig:rCD_cs} illustrates the rCD (see equation 2 in the main paper) for each ablated variant evaluated on the Cityscapes dataset.
In the following, we briefly discuss the ablated variants \wrt rCD.
\\ \indent \textbf{Effect of ASPP.}
The rCD score is especially pronounced for geometrically distorted image data (\SI{146}{\%} for Xception-41, \SI{46}{\%} for MobileNet-V2). 
The mIoU of Xception-41 on geometrically distorted data is low, resulting hence in a high CD and rCD. 
Regarding MobileNet-V2, the averaged mIoU is even similar to the other ablated variants (see the last column of Table ~\ref{tab:miou_cs_allbackbones_and_ablations}).
\\ \indent \textbf{Effect of AC.} 
As discussed in the main paper, AC show often an aiding effect against corruptions of type blur, noise, and geometric distortion (especially for ResNets and Xception-71).
The rCD mostly has a similar tendency as the CD.
\\ \indent \textbf{Effect of DPC.} 
The rCD scores for the ablated variant without ASPP and with Dense Prediction Cell, generally show the same tendency as the CD illustrated in the main paper in Figure 5. 
\\ \indent \textbf{Effect of LRL.} 
As mentioned in the main paper, this ablated variant is vulnerable to \textit{intensity noise} (\textit{defocus blur}), when applied in ResNet-101 (MobileNet-V2), since its rCD is \SI{124}{\%} (\SI{118}{\%}).
The rCD with respect to geometric distortion is especially high for MobileNet-V2 (\SI{155}{\%}).
\\ \indent \textbf{Effect of GAP.} 
The rCD of Xception-71 and ResNet-101, using GAP, show a similar tendency as the CD discussed in the main paper.
The rCD is low for Xception-71 \wrt geometric distortion.
\\ \indent Finally, we list for completeness the individual CD and rCD scores, evaluated on Cityscapes, in Table \ref{tab:CD_cs_allbackbones_and_ablations} and Table~\ref{tab:rCD_cs_allbackbones_and_ablations}.

\subsection{Experimental Results on PASCAL VOC 2012}
\label{sec:experimentalresults_pascal}
\indent Table~\ref{tab:miou_voc_allbackbones_and_ablations} contains the mIoU for clean and corrupted variants of the validation set of PASCAL VOC 2012 for several network backbones of the DeepLabv3+ architecture. 
In contrast to the model performance evaluated on Cityscapes, corruptions of category noise and weather have a less corrupting impact. 
Each backbone performs best on clean data when GAP is used.
Each backbone performs significantly worse without ASPP.
When GAP is used with ResNets and Xception-65, the resulting model is the best performing model on most types of image corruptions. 
Regarding Xception-41 and Xception-71, the ablated variant without LRL often has the highest mIoU \wrt image corruptions of category noise.
Figure \ref{fig:CD_pascal} and Figure \ref{fig:rCD_pascal} illustrates the CD and rCD for each ablated variant evaluated on PASCAL VOC 2012.
In the following, we will briefly discuss the ablated variants \wrt rCD.
\\ \indent \textbf{Effect of ASPP.} 
For geometric distortion, the rCD of this ablated variant often shows except for Xception-41 a similar tendency the rCD as on Cityscapes (see Figure ~\ref{fig:rCD_cs}). 
The rCD for ranges from \SI{24}{\%} (ResNet-101) to \SI{62}{\%} (Xception-65).
\\ \indent \textbf{Effect of AC.} 
As mentioned in the main paper, AC show no positive effect against blur. 
We explain this with the fundamentally different datasets. 
On Cityscapes, a model without AC often overlooks classes covering small image-regions, especially when far away. 
Such images are hardly present in PASCAL VOC 2012. 
An example of Cityscapes is illustrated in Figure~\ref{fig:blur_missclassify}. 
The tendencies of CD and rCD are often similar to geometric distortion. 
As on Cityscapes, AC aids the robustness for ResNet-101 against this image corruption, but not for ResNet-50.
\\ \indent \textbf{Effect of DPC.}
The harming effect of DPC with respect to image corruptions is especially present for Xception-71. 
As mentioned in the main paper, a possible explanation might be that the neural-architecture-search has been performed on Xception-71. 
\\ \indent \textbf{Effect of LRL.} 
The rCD for this ablated variant is especially high \wrt geometrically distorted image data when applied in Xception-41 and ResNet-50 (\SI{144}{\%} and \SI{127}{\%}, respectively). 
As the CD reported in the main paper, also the rCD of Xception-71 and Xception-41 for image noise is below \SI{100}{\%}.
\\ \indent \textbf{Effect of GAP.} 
Unlike the tendency of CD, the rCD of this ablated variant is rarely below \SI{100.0}{\%}.
The rCD \wrt is high for geometric distortion.
\\ 
\indent 
Finally, we list the individual CD and rCD scores, evaluated on PASCAL VOC 2012, in Table \ref{tab:cd_voc_allbackbones_and_ablations} and Table~\ref{tab:rcd_voc_allbackbones_and_ablations}.

\subsection{Experimental Results on ADE20K}
\label{sec:experimentalresults_ade}
\indent Table~\ref{tab:miou_ade_allbackbones_and_ablations} contains the mIoU for clean and corrupted variants of the validation set of ADE20K for several network backbones of the DeepLabv3+ architecture. 
When comparing the respective reference model of each backbone (\ie no ablated variants), Xception-71 performs best for every type of image corruption.
MobileNet-V2 (ResNets) oftentimes performs best when DPC (GAP) is used.
Xception-41 and Xception-71 perform best on clean data when DPC is used. 
Most backbones without ASPP perform significantly worse than respective reference.
Figure \ref{fig:CD_ade} and Figure \ref{fig:rCD_ade} illustrates the CD and rCD for each ablated variant evaluated on ADE20K.
As mentioned in the main paper, the CD is--except for models without ASPP--oftentimes around \SI{100}{\%}. 
In the following, we will briefly discuss the ablated variants \wrt CD and rCD.
\\ \indent \textbf{Effect of ASPP.} 
The rCD is in general above \SI{100}{\%} for both Xception-65 and Xception-71, and below \SI{100}{\%} for remaining backbones. 
The performance gap \wrt the reference model is for the aforementioned Xception-based backbones significantly less than for the remaining backbones.
\\ \indent \textbf{Effect of AC.} 
The removal of AC decreases the performance slightly for most backbones against corruptions of category digital and weather.
\\ \indent \textbf{Effect of DPC.}
As on PASCAL VOC 2012 and Cityscapes, applying DPC oftentimes decreases the robustness, especially for Xception-71 against most image corruptions. 
As on Cityscapes, using DPC along Xception-71, results in the best-performing model on clean data.
\\ \indent \textbf{Effect of LRL.} 
The removal of LRL impacts especially Xception-71 against image noise.
\\ \indent \textbf{Effect of GAP.} 
When GAP is applied, the models perform generally most robust.
\\ \indent Finally, we list the individual CD and rCD scores, evaluated on ADE20K, in Table \ref{tab:cd_ade_allbackbones_and_ablations} and Table~\ref{tab:rcd_ade_allbackbones_and_ablations}.

\subsection{Performance without ASPP}
As mentioned in the main paper, we provide a more detailed evaluation for the ablated architecture without ASPP for every dataset, in this subsection.
The Atrous Spatial Pyramid Pooling (ASPP) module reduces, in general, the model performance significantly. 
On PASCAL VOC 2012, the mIoU on clean data reduces between \SI{5.9}{\%} (Xception-65) and \SI{12.0}{\%} (ResNet-50).
On ADE20K, the mIoU decreases between \SI{1.2}{\%} (Xception-65) and \SI{7.7}{\%} (ResNet-50).
On Cityscapes, the mIoU decreases between \SI{2.4}{\%} (Xception-41) and \SI{7.1}{\%} (MobileNet-V2). 
Therefore, the corresponding CD scores are oftentimes considerably high. 
On PASCAL VOC 2012 and ADE20K, for example, the CD score for the ablated variant w/o ASPP is the highest \wrt every image corruption and for every network backbone (see bold values Table~\ref{tab:cd_voc_allbackbones_and_ablations} and Table~\ref{tab:cd_ade_allbackbones_and_ablations}).
Regarding the evaluation on Cityscapes (see Table~\ref{tab:CD_cs_allbackbones_and_ablations}), the CD score of the ablated variant w/o ASPP is often high against image corruptions of category blur for ResNets and MobileNet-V2.
Stronger backbones, on the other hand, as Xception-based ones, perform better without ASPP and have thus a lower CD.

\subsection{Performance with respect to Individual Severity Levels}
\label{sec:rebuttal_degrade_severity_levels}
\begin{figure*}[h]
\centering
\includegraphics[width=1\linewidth]{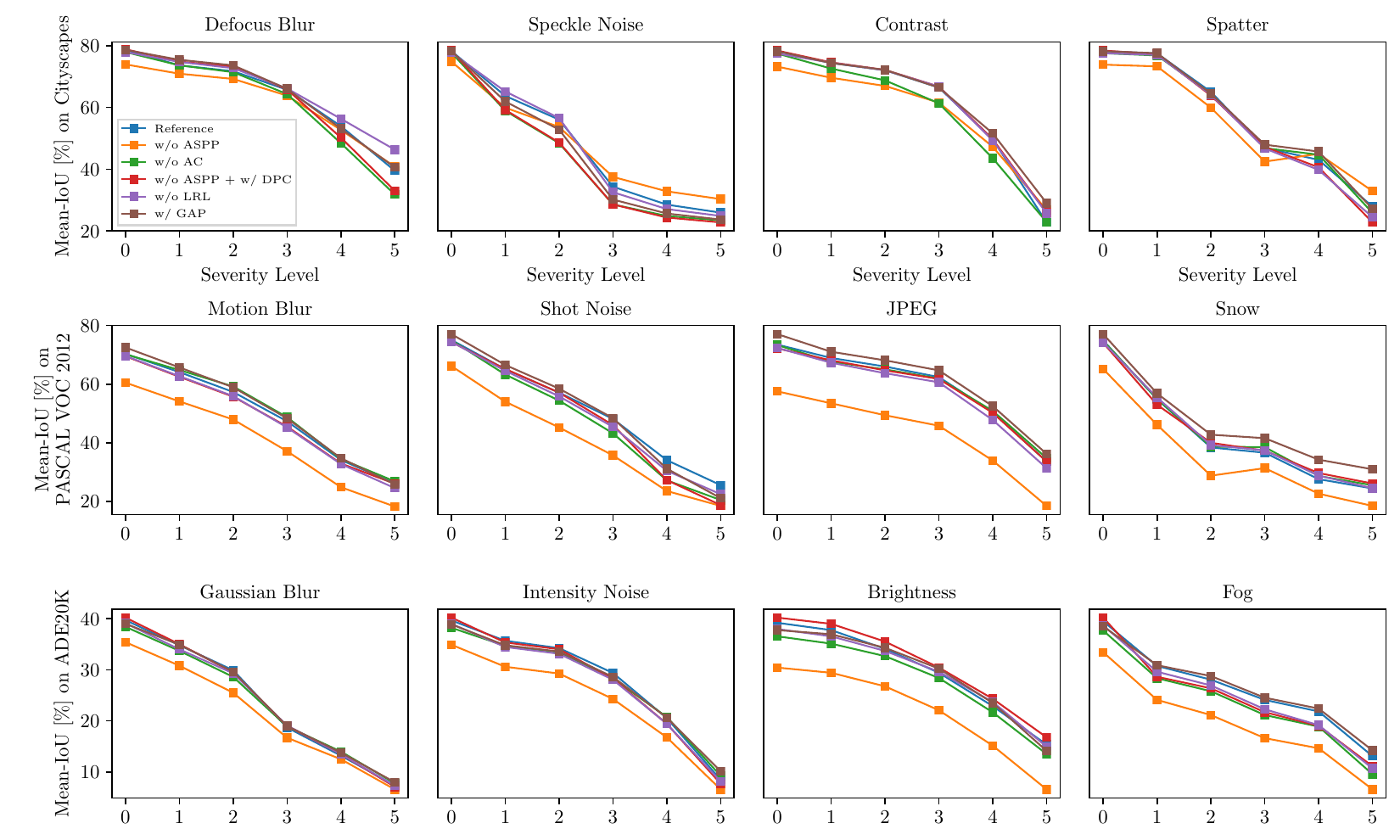}
\caption{
Model performance (mIoU) for many candidates with respect to the image corruption categories blur (first column), noise (second column), digital (third column), and weather (fourth column) for a reference model and all corresponding architectural ablated variants, evaluated for every severity levels on Cityscapes, PASCAL VOC 2012, and ADE20K.
Severity level 0 corresponds to clean data.
\textbf{First row:} Xception-71 evaluated on the Cityscapes dataset for defocus blur, speckle noise, contrast, and spatter.
\textbf{Second row:} ResNet-101 evaluated on PASCAL VOC 2012 for motion blur, shot noise, JPEG, and snow.
\textbf{Third row:} Xception-41 evaluated on ADE20K for Gaussian blur, intensity noise, brightness, and fog.
The ablated variant without ASPP oftentimes has the lowest mIoU.
However, it performs best on speckle noise for severity level 3 and above.
The mIoU of the ablated variant without AC is relatively low for defocus blur and contrast.
The mIoU of the ablated variant without ASPP and with DPC is relatively low for speckle noise, shot noise (for severity level 4 and 5), spatter.
The mIoU of the ablated variant without LRL is relatively high for speckle noise and shot noise.
The mIoU of the ablated variant with GAP is high for PASCAL VOC 2012 on clean data and low for speckle noise. 
}
\label{fig:rebuttal_severityimg}	
\end{figure*}

\begin{figure*}
	\centering
	\includegraphics[width=\textwidth]{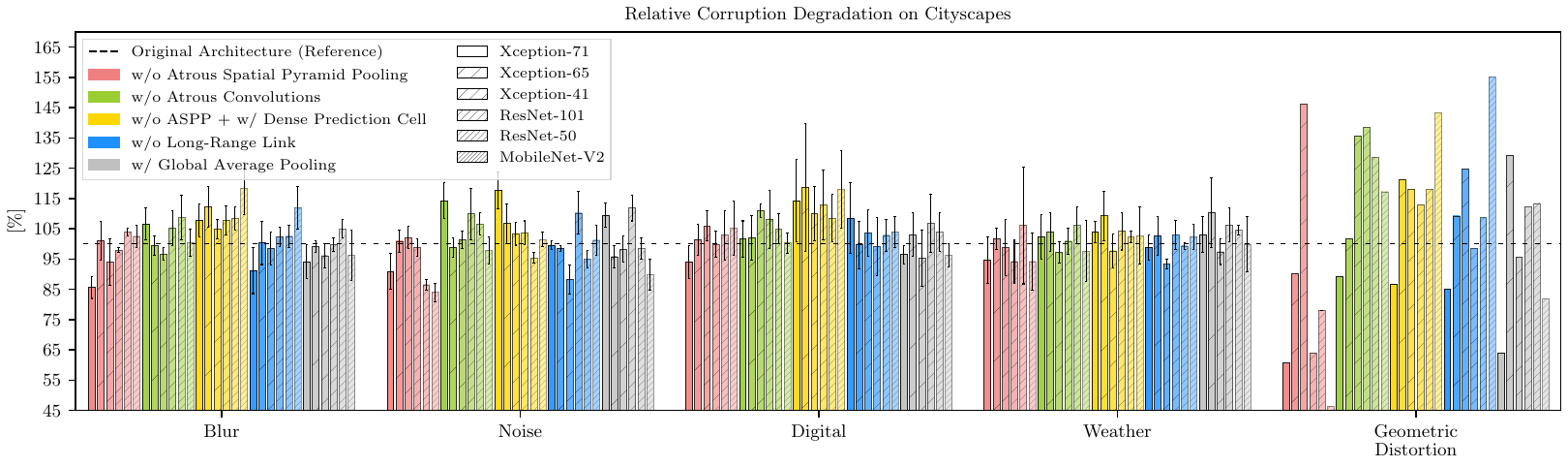}
	\caption{
		Relative CD evaluated on Cityscapes for the proposed ablated variants of the DeepLabv3$+$ architecture \wrt image corruptions, employing six different network backbones. 
		Each bar except for geometric distortion is averaged within a corruption category (error bars indicate the standard deviation).
		Bars above \SI{100}{\%} represent a relative decrease in performance compared to the respective reference architecture.
		Each ablated architecture is re-trained on the original training dataset. 
		Removing ASPP may decrease performance significantly.
		The low rCD for geometric distortion indicates that the relative decrease of performance for this ablated variant is low.
		AC affect model performance, particularly against geometric distortion.
		The relative CD is often high against most image corruptions when DPC is used.
		The effect of GAP depends strongly on the network backbone.
		Best viewed in color.}
	\label{fig:rCD_cs}
\end{figure*}

\begin{figure*}
	\centering
	\includegraphics[width=\textwidth]{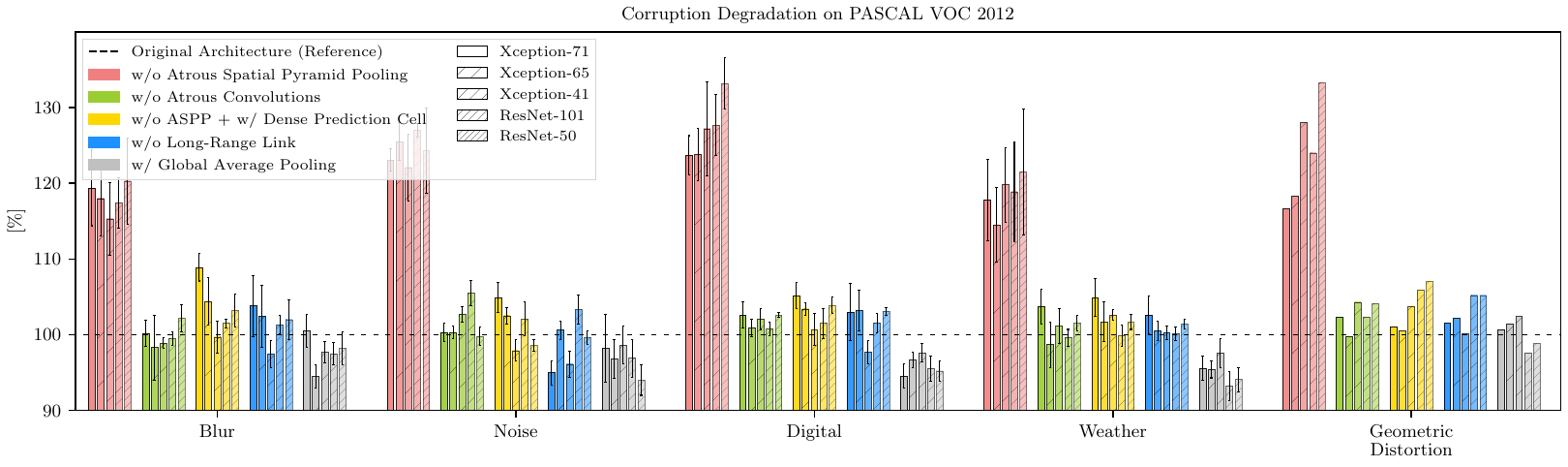}
	\caption{
		CD evaluated on PASCAL VOC 2012 for the proposed ablated variants of the DeepLabv3$+$ architecture \wrt image corruptions, employing five different network backbones. 
		Each bar except for geometric distortion is averaged within a corruption category (error bars indicate the standard deviation).
		Bars above \SI{100}{\%} represent a decrease in performance compared to the respective reference architecture.
		Each ablated architecture is re-trained on the original training dataset. 
		Removing ASPP reduces the model performance significantly. 
		AC and LRL decrease robustness against corruptions of category \textit{digital} slightly. 
		Xception-71 is vulnerable against many corruptions when DPC is used. 
		GAP increases performance against many corruptions. Each backbone performs further best on clean data when GAP is used.
		Best viewed in color.
	}
	\label{fig:CD_pascal}
\end{figure*}

\begin{figure*}
	\centering
	\includegraphics[width=\textwidth]{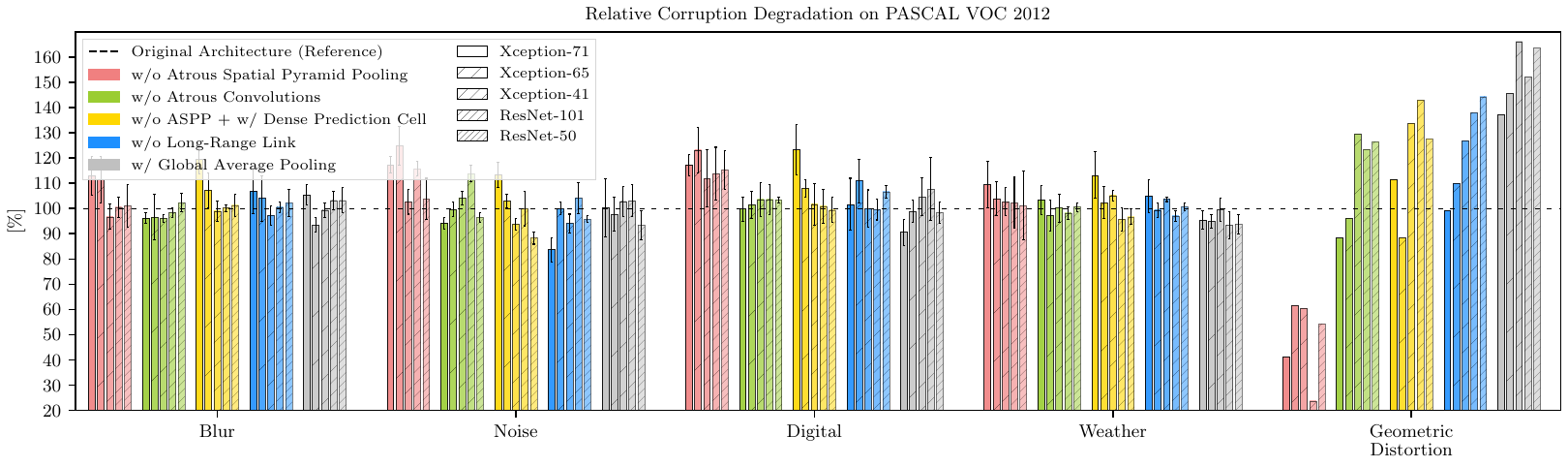}
	\caption{
		Relative CD evaluated on PASCAL VOC 2012 for the proposed ablated variants of the DeepLabv3$+$ architecture \wrt image corruptions, employing five different network backbones. 
		Each bar except for geometric distortion is averaged within a corruption category (error bars indicate the standard deviation).
		Bars above \SI{100}{\%} represent a relative decrease in performance compared to the respective reference architecture.
		Each ablated architecture is re-trained on the original training dataset. 
		Removing ASPP decreases performance oftentimes significantly.
		The low rCD for geometric distortion indicates that the relative decrease of performance for this ablated variant is low.
		AC aids the robustness against geometric distortion for several backbones.
		The harming effect of DPC with respect to image corruptions is especially pronounced for Xception-71.
		The rCD of LRL is large against geometric distortion for ResNet-50. 
		The rCD of GAP has, oftentimes, a contrary tendency as the CD. 
		Best viewed in color.
	}
	\label{fig:rCD_pascal}
\end{figure*}

\begin{figure*}
	\centering
	\includegraphics[width=\textwidth]{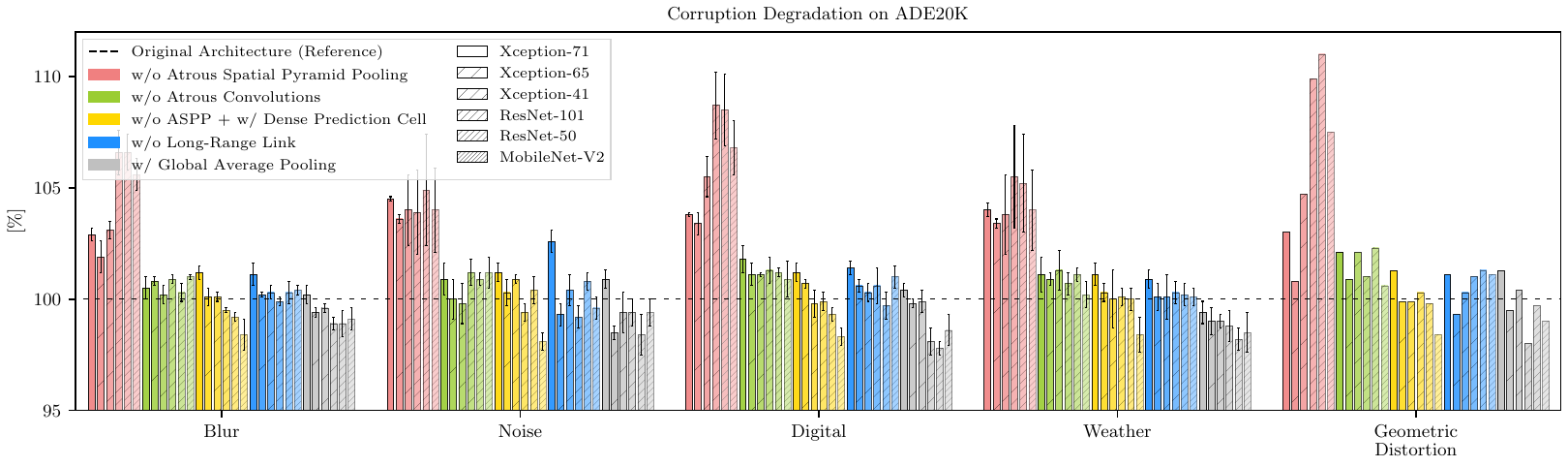}
	\caption{
		CD evaluated on ADE20K for the proposed ablated variants of the DeepLabv3$+$ architecture \wrt image corruptions, employing six different network backbones. 
		Each bar except for geometric distortion is averaged within a corruption category (error bars indicate the standard deviation).
		Bars above \SI{100}{\%} represent a relative decrease in performance compared to the respective reference architecture.
		Each ablated architecture is re-trained on the original training dataset. 
        Removing ASPP decreases performance oftentimes.
        AC increase performance slightly against most corruptions.
        DPC and LRL hamper the performance for Xception-71 \wrt several image corruptions.
 		GAP increases the robustness for most backbones against many image corruptions.
		Best viewed in color.
}
	\label{fig:CD_ade}
\end{figure*}

\begin{figure*}
	\centering
	\includegraphics[width=\textwidth]{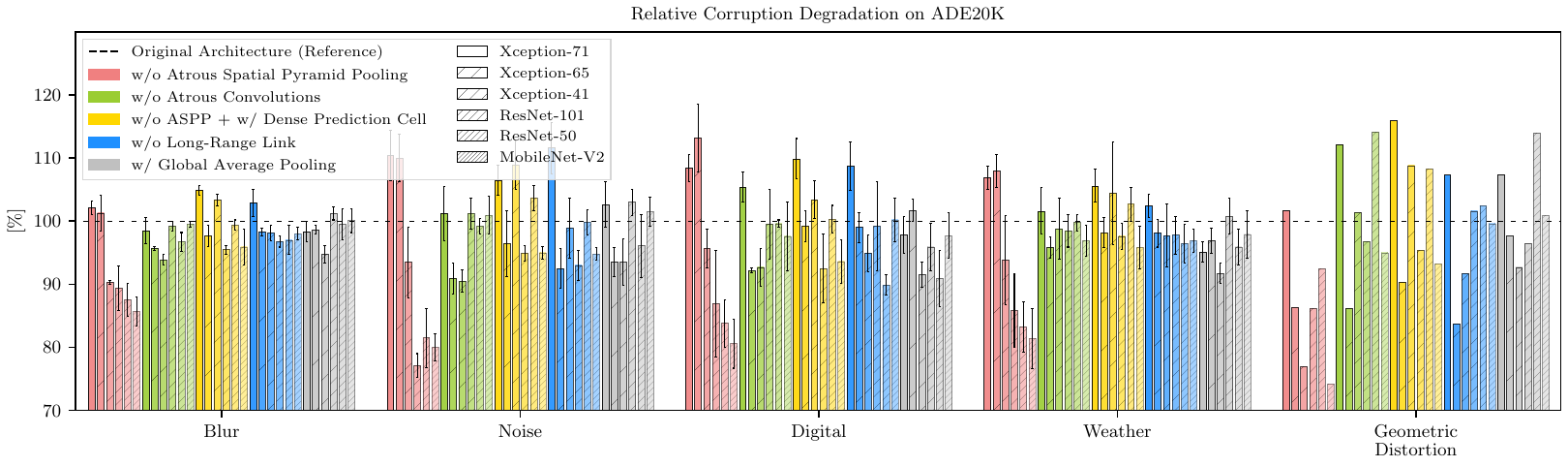}
	\caption{
		Relative CD evaluated on ADE20K for the proposed ablated variants of the DeepLabv3$+$ architecture \wrt image corruptions, employing six different network backbones. Bars above \SI{100}{\%} represent a relative decrease in performance compared to the respective reference architecture.
		Each bar except for geometric distortion is averaged within a corruption category (error bars indicate the standard deviation).
		Each ablated architecture is re-trained on the original training dataset. 
		Removing ASPP decreases performance oftentimes significantly.
		The low rCD for geometric distortion indicates that the relative decrease of performance for this ablated variant is low (except for Xception-71).
		The rCD of DPC and LRL are oftentimes highest for Xception-71.
		GAP increases the robustness for most backbones against many image corruptions.
		Best viewed in color.
}
	\label{fig:rCD_ade}
\end{figure*}

We illustrate in Fig.~\ref{fig:rebuttal_severityimg} the model performance evaluated on every dataset with respect to individual severity levels.
The Figure shows the degrading performance with increasing severity level for some candidates of category blur, noise, digital, and weather of a reference model and all corresponding architectural ablations.
Please see the caption for discussion.

\begin{table*}[h]
	\begin{adjustbox}{width=\textwidth}
	\begin{tabular}{@{}cccccccccccccccccccc@{}}
		\toprule
		& \multicolumn{5}{c}{\textbf{Blur}} & \multicolumn{5}{c}{\textbf{Noise}} & \multicolumn{4}{c}{\textbf{Digital}} & \multicolumn{4}{c}{\textbf{Weather}} & \textbf{} \\ \midrule
		\multicolumn{1}{c|}{\begin{tabular}[c]{@{}c@{}}Deeplab-v3+ \\ Backbone\end{tabular}} & Motion & Defocus & \begin{tabular}[c]{@{}c@{}}Frosted \\ Glass\end{tabular} & Gaussian & \multicolumn{1}{c|}{PSF} & Gaussian & Impulse & Shot & Speckle & \multicolumn{1}{c|}{Intensity} & Brightness & Contrast & Saturate & \multicolumn{1}{c|}{JPEG} & Snow & Spatter & Fog & \multicolumn{1}{c|}{Frost} & \begin{tabular}[c]{@{}c@{}}Geometric\\ Distortion\end{tabular} \\ \midrule
		\multicolumn{1}{c|}{\textbf{ICNet}} & 100.0 & 100.0 & 100.0 & 100.0 & \multicolumn{1}{c|}{100.0} & 100.0 & 100.0 & 100.0 & 100.0 & \multicolumn{1}{c|}{100.0} & \textbf{100.0} & 100.0 & \textbf{100.0} & \multicolumn{1}{c|}{100.0} & \textbf{100.0} & \textbf{100.0} & \textbf{100.0} & \multicolumn{1}{c|}{\textbf{100.0}} & \textbf{100.0} \\
		\multicolumn{1}{c|}{FCN8s-VGG16} & \textbf{105.6} & \textbf{124.3} & 119.6 & \textbf{119.1} & \multicolumn{1}{c|}{110.8} & 101.8 & 103.0 & 103.2 & 104.1 & \multicolumn{1}{c|}{\textbf{115.6}} & 79.2 & 91.2 & 88.2 & \multicolumn{1}{c|}{119.4} & 94.7 & 98.4 & 85.8 & \multicolumn{1}{c|}{90.2} & 98.1 \\
		\multicolumn{1}{c|}{DilatedNet} & 102.6 & 115.1 & \textbf{128.3} & 111.4 & \multicolumn{1}{c|}{\textbf{111.8}} & 92.2 & 93.9 & 91.3 & 93.3 & \multicolumn{1}{c|}{91.9} & 80.2 & \textbf{100.7} & 85.4 & \multicolumn{1}{c|}{107.3} & 93.4 & 97.3 & 89.8 & \multicolumn{1}{c|}{90.7} & 95.1 \\
		\multicolumn{1}{c|}{ResNet-38} & 83.7 & 99.2 & 107.8 & 95.5 & \multicolumn{1}{c|}{72.0} & 94.3 & 91.8 & 91.5 & 85.5 & \multicolumn{1}{c|}{91.2} & 67.8 & 73.8 & 73.3 & \multicolumn{1}{c|}{129.2} & 92.4 & 77.9 & 64.7 & \multicolumn{1}{c|}{87.4} & 88.4 \\
		\multicolumn{1}{c|}{PSPNet} & 74.1 & 84.6 & 105.7 & 83.3 & \multicolumn{1}{c|}{66.3} & 97.1 & 92.4 & 94.7 & 91.1 & \multicolumn{1}{c|}{96.2} & 67.1 & 72.1 & 95.7 & \multicolumn{1}{c|}{119.1} & 97.8 & 82.5 & 90.2 & \multicolumn{1}{c|}{94.1} & 88.0 \\
		\multicolumn{1}{c|}{GSCNN} & 75.9 & 75.1 & 110.4 & 72.2 & \multicolumn{1}{c|}{56.5} & \textbf{103.2} & \textbf{106.4} & \textbf{104.3} & \textbf{104.4} & \multicolumn{1}{c|}{100.0} & 40.9 & 57.0 & 40.4 & \multicolumn{1}{c|}{\textbf{133.2}} & 93.5 & 75.8 & 44.1 & \multicolumn{1}{c|}{75.7} & 89.2 \\ \midrule
		\multicolumn{1}{c|}{\textbf{ICNet}} & 100.0 & 100.0 & 100.0 & 100.0 & \multicolumn{1}{c|}{100.0} & 100.0 & 100.0 & 100.0 & 100.0 & \multicolumn{1}{c|}{100.0} & \textbf{100.0} & 100.0 & 100.0 & \multicolumn{1}{c|}{100.0} & 100.0 & 100.0 & 100.0 & \multicolumn{1}{c|}{100.0} & 100.0 \\
		\multicolumn{1}{c|}{FCN8s-VGG16} & 119.1 & \textbf{167.3} & 160.1 & \textbf{153.8} & \multicolumn{1}{c|}{779.8} & 104.3 & 106.1 & 106.6 & 109.9 & \multicolumn{1}{c|}{132.5} & 54.0 & 84.6 & 79.9 & \multicolumn{1}{c|}{142.7} & 93.1 & 99.1 & 75.3 & \multicolumn{1}{c|}{85.5} & 98.6 \\
		\multicolumn{1}{c|}{DilatedNet} & \textbf{120.5} & 152.2 & 195.4 & 142.8 & \multicolumn{1}{c|}{\textbf{1117.9}} & 92.3 & 95.0 & 90.8 & 94.4 & \multicolumn{1}{c|}{91.8} & 64.0 & \textbf{109.9} & 79.5 & \multicolumn{1}{c|}{123.8} & 94.3 & 102.5 & 87.8 & \multicolumn{1}{c|}{89.9} & 98.6 \\
		\multicolumn{1}{c|}{ResNet-38} & 114.2 & 152.9 & 185.3 & 143.5 & \multicolumn{1}{c|}{388.9} & 111.2 & 107.2 & 107.4 & 103.1 & \multicolumn{1}{c|}{115.2} & 70.6 & 82.1 & 80.0 & \multicolumn{1}{c|}{197.2} & 107.7 & 89.5 & 63.7 & \multicolumn{1}{c|}{100.8} & 114.0 \\
		\multicolumn{1}{c|}{PSPNet} & 93.7 & 120.0 & 185.3 & 116.8 & \multicolumn{1}{c|}{256.0} & 117.7 & 110.2 & 114.6 & 116.8 & \multicolumn{1}{c|}{127.9} & 73.5 & 82.1 & \textbf{125.2} & \multicolumn{1}{c|}{179.7} & \textbf{117.9} & \textbf{101.7} & \textbf{114.7} & \multicolumn{1}{c|}{\textbf{113.7}} & 116.8 \\
		\multicolumn{1}{c|}{GSCNN} & 109.7 & 105.9 & \textbf{211.0} & 98.3 & \multicolumn{1}{c|}{85.1} & \textbf{131.2} & \textbf{136.4} & \textbf{134.2} & \textbf{147.9} & \multicolumn{1}{c|}{\textbf{141.6}} & 20.2 & 58.1 & 26.5 & \multicolumn{1}{c|}{\textbf{216.0}} & 115.0 & 95.0 & 33.7 & \multicolumn{1}{c|}{87.9} & \textbf{126.8} \\ \bottomrule
	\end{tabular}
	\end{adjustbox}
	\caption{
		CD (top) and rCD (bottom) for corrupted variants of the validation set of the Cityscapes dataset for several non-Deeplabv3$+$ based architectures. 
		ICNet is used as reference model.
		Highest CD and rCD per corruption is bold.}
	\label{tab:miou_cs_nondeeplab}
\end{table*}

\begin{table*}[h]
	\begin{adjustbox}{width=\textwidth}
		\begin{tabular}{@{}ccccccccccccccccccccc@{}}
			\toprule
			\multicolumn{2}{c}{} & \multicolumn{5}{c}{\textbf{Blur}} & \multicolumn{5}{c}{\textbf{Noise}} & \multicolumn{4}{c}{\textbf{Digital}} & \multicolumn{4}{c}{\textbf{Weather}} & \textbf{} \\ \midrule
			\multicolumn{1}{c|}{\begin{tabular}[c]{@{}c@{}}Deeplab-v3+ \\ Backbone\end{tabular}} & \multicolumn{1}{c|}{Clean} & Motion & Defocus & \begin{tabular}[c]{@{}c@{}}Frosted \\ Glass\end{tabular} & Gaussian & \multicolumn{1}{c|}{PSF} & Gaussian & Impulse & Shot & Speckle & \multicolumn{1}{c|}{Intensity} & Brightness & Contrast & Saturate & \multicolumn{1}{c|}{JPEG} & Snow & Spatter & Fog & \multicolumn{1}{c|}{Frost} & \begin{tabular}[c]{@{}c@{}}Geometric\\ Distortion\end{tabular} \\ \midrule
			\multicolumn{1}{c|}{\textbf{MobileNet-V2}} & \multicolumn{1}{c|}{\textbf{72.0}} & \textbf{53.5} & 49.0 & \textbf{45.3} & 49.1 & \multicolumn{1}{c|}{\textbf{70.5}} & 6.4 & 7.0 & 6.6 & 16.6 & \multicolumn{1}{c|}{26.9} & 51.7 & 46.7 & 32.4 & \multicolumn{1}{c|}{27.2} & \textbf{13.7} & 38.9 & \textbf{47.4} & \multicolumn{1}{c|}{\textbf{17.3}} & 65.5 \\
			\multicolumn{1}{c|}{w/o ASPP} & \multicolumn{1}{c|}{64.9} & 45.5 & 40.4 & 39.0 & 41.5 & \multicolumn{1}{c|}{63.3} & 7.7 & 8.7 & 8.9 & 19.9 & \multicolumn{1}{c|}{28.4} & 41.7 & 36.1 & 27.6 & \multicolumn{1}{c|}{20.7} & 13.0 & 36.8 & 37.8 & \multicolumn{1}{c|}{14.0} & 61.9 \\
			\multicolumn{1}{c|}{w/o AC} & \multicolumn{1}{c|}{71.2} & 52.1 & 49.1 & 42.8 & 49.3 & \multicolumn{1}{c|}{69.8} & 3.6 & 7.2 & 4.5 & 19.6 & \multicolumn{1}{c|}{29.2} & 49.8 & 46.2 & 31.4 & \multicolumn{1}{c|}{\textbf{28.1}} & 10.0 & \textbf{44.6} & 45.2 & \multicolumn{1}{c|}{16.7} & 63.5 \\
			\multicolumn{1}{c|}{w/ DPC} & \multicolumn{1}{c|}{71.6} & 49.4 & 42.2 & 43.7 & 43.8 & \multicolumn{1}{c|}{69.2} & 3.5 & 4.9 & 3.9 & 16.1 & \multicolumn{1}{c|}{27.4} & 45.0 & 38.6 & 30.1 & \multicolumn{1}{c|}{24.1} & 9.8 & 42.8 & 43.9 & \multicolumn{1}{c|}{14.0} & 62.2 \\
			\multicolumn{1}{c|}{w/o LRL} & \multicolumn{1}{c|}{71.1} & 49.7 & 43.9 & 44.4 & 45.1 & \multicolumn{1}{c|}{68.9} & 1.9 & 2.6 & 2.5 & 19.6 & \multicolumn{1}{c|}{26.6} & 49.1 & 43.5 & 32.2 & \multicolumn{1}{c|}{26.3} & 10.4 & 39.5 & 44.9 & \multicolumn{1}{c|}{14.7} & 60.9 \\
			\multicolumn{1}{c|}{w/ GAP} & \multicolumn{1}{c|}{71.4} & 52.2 & \textbf{50.6} & 43.3 & \textbf{51.8} & \multicolumn{1}{c|}{69.8} & \textbf{8.5} & \textbf{10.9} & \textbf{10.8} & \textbf{26.0} & \multicolumn{1}{c|}{\textbf{32.5}} & \textbf{51.8} & \textbf{47.3} & \textbf{35.3} & \multicolumn{1}{c|}{25.7} & 12.7 & 43.4 & 45.1 & \multicolumn{1}{c|}{12.6} & \textbf{66.0} \\ \midrule
			\multicolumn{1}{c|}{\textbf{ResNet-50}} & \multicolumn{1}{c|}{76.6} & \textbf{58.5} & \textbf{56.6} & 47.2 & \textbf{57.7} & \multicolumn{1}{c|}{74.8} & 6.5 & 7.2 & 10.0 & 31.1 & \multicolumn{1}{c|}{30.9} & \textbf{58.2} & \textbf{54.7} & 41.3 & \multicolumn{1}{c|}{27.4} & \textbf{12.0} & \textbf{42.0} & 55.9 & \multicolumn{1}{c|}{\textbf{22.8}} & \textbf{69.5} \\
			\multicolumn{1}{c|}{w/o ASPP} & \multicolumn{1}{c|}{71.4} & 52.3 & 50.7 & 41.2 & 52.0 & \multicolumn{1}{c|}{69.7} & 10.1 & 11.3 & \textbf{13.8} & 31.2 & \multicolumn{1}{c|}{33.3} & 50.2 & 48.4 & 37.0 & \multicolumn{1}{c|}{25.3} & 12.0 & 38.6 & 42.7 & \multicolumn{1}{c|}{18.7} & 65.9 \\
			\multicolumn{1}{c|}{w/o AC} & \multicolumn{1}{c|}{76.0} & 56.7 & 53.1 & \textbf{47.3} & 54.1 & \multicolumn{1}{c|}{73.8} & 2.4 & 6.1 & 5.1 & 25.5 & \multicolumn{1}{c|}{25.7} & 56.8 & 51.4 & 38.9 & \multicolumn{1}{c|}{27.6} & 9.7 & 40.8 & 52.0 & \multicolumn{1}{c|}{20.1} & 66.9 \\
			\multicolumn{1}{c|}{w/ DPC} & \multicolumn{1}{c|}{\textbf{76.9}} & 57.0 & 54.7 & 46.9 & 56.2 & \multicolumn{1}{c|}{74.2} & \textbf{10.7} & \textbf{12.6} & 13.6 & \textbf{33.1} & \multicolumn{1}{c|}{32.0} & 54.5 & 53.6 & \textbf{41.5} & \multicolumn{1}{c|}{25.1} & 11.4 & 41.3 & \textbf{56.3} & \multicolumn{1}{c|}{20.4} & 68.6 \\
			\multicolumn{1}{c|}{w/o LRL} & \multicolumn{1}{c|}{75.6} & 57.9 & 54.4 & 46.4 & 55.6 & \multicolumn{1}{c|}{73.8} & 7.9 & 9.3 & 11.2 & 31.8 & \multicolumn{1}{c|}{\textbf{34.7}} & 56.2 & 51.6 & 40.2 & \multicolumn{1}{c|}{\textbf{28.5}} & 11.9 & 41.4 & 55.4 & \multicolumn{1}{c|}{21.1} & 67.9 \\
			\multicolumn{1}{c|}{w/ GAP} & \multicolumn{1}{c|}{76.5} & 56.7 & 55.7 & 45.8 & 57.4 & \multicolumn{1}{c|}{\textbf{75.2}} & 5.5 & 7.8 & 9.5 & 31.3 & \multicolumn{1}{c|}{34.5} & 57.7 & 51.4 & 41.1 & \multicolumn{1}{c|}{28.3} & 10.5 & 40.4 & 54.5 & \multicolumn{1}{c|}{20.1} & 68.5 \\ \midrule
			\multicolumn{1}{c|}{\textbf{ResNet-101}} & \multicolumn{1}{c|}{77.1} & \textbf{59.1} & 56.3 & 47.7 & 57.3 & \multicolumn{1}{c|}{75.2} & \textbf{13.2} & \textbf{13.9} & \textbf{16.3} & \textbf{36.9} & \multicolumn{1}{c|}{\textbf{39.9}} & \textbf{59.2} & 54.5 & \textbf{41.5} & \multicolumn{1}{c|}{\textbf{37.4}} & \textbf{11.9} & \textbf{47.8} & 55.1 & \multicolumn{1}{c|}{\textbf{22.7}} & \textbf{69.7} \\
			\multicolumn{1}{c|}{w/o ASPP} & \multicolumn{1}{c|}{71.1} & 53.8 & 50.6 & 42.2 & 51.7 & \multicolumn{1}{c|}{68.8} & 9.5 & 9.8 & 12.7 & 30.7 & \multicolumn{1}{c|}{32.5} & 52.1 & 48.3 & 36.7 & \multicolumn{1}{c|}{33.2} & 13.3 & 43.5 & 47.8 & \multicolumn{1}{c|}{23.2} & 66.4 \\
			\multicolumn{1}{c|}{w/o AC} & \multicolumn{1}{c|}{75.7} & 57.9 & 52.5 & 46.6 & 53.9 & \multicolumn{1}{c|}{73.3} & 8.4 & 11.0 & 11.6 & 31.5 & \multicolumn{1}{c|}{28.8} & 53.5 & 53.1 & 39.1 & \multicolumn{1}{c|}{34.2} & 9.9 & 44.7 & 55.0 & \multicolumn{1}{c|}{20.0} & 65.5 \\
			\multicolumn{1}{c|}{w/ DPC} & \multicolumn{1}{c|}{77.0} & 58.5 & 53.5 & 46.7 & 54.8 & \multicolumn{1}{c|}{75.3} & 11.7 & 12.1 & 15.6 & 36.4 & \multicolumn{1}{c|}{35.5} & 53.7 & 54.3 & 39.8 & \multicolumn{1}{c|}{30.9} & 10.1 & 44.0 & \textbf{56.0} & \multicolumn{1}{c|}{19.3} & 68.6 \\
			\multicolumn{1}{c|}{w/o LRL} & \multicolumn{1}{c|}{76.5} & 58.7 & 54.6 & 47.5 & 55.7 & \multicolumn{1}{c|}{74.3} & 9.1 & 8.3 & 12.1 & 33.5 & \multicolumn{1}{c|}{30.3} & 57.0 & \textbf{57.6} & 40.9 & \multicolumn{1}{c|}{35.7} & 9.3 & 44.3 & 55.4 & \multicolumn{1}{c|}{20.8} & 69.2 \\
			\multicolumn{1}{c|}{w/ GAP} & \multicolumn{1}{c|}{\textbf{77.3}} & 58.7 & \textbf{56.9} & \textbf{48.4} & \textbf{57.8} & \multicolumn{1}{c|}{\textbf{75.9}} & 8.2 & 7.4 & 11.6 & 32.0 & \multicolumn{1}{c|}{32.8} & 55.6 & 55.8 & 39.3 & \multicolumn{1}{c|}{36.4} & 11.5 & 44.8 & 52.5 & \multicolumn{1}{c|}{22.6} & 69.0 \\ \midrule
			\multicolumn{1}{c|}{\textbf{Xception-41}} & \multicolumn{1}{c|}{\textbf{77.8}} & 61.6 & 54.9 & 51.0 & \textbf{54.7} & \multicolumn{1}{c|}{\textbf{76.1}} & 17.0 & 17.3 & 21.6 & 43.7 & \multicolumn{1}{c|}{\textbf{48.6}} & 63.6 & 56.9 & \textbf{51.7} & \multicolumn{1}{c|}{\textbf{38.5}} & 18.2 & 46.6 & 57.6 & \multicolumn{1}{c|}{20.6} & \textbf{73.0} \\
			\multicolumn{1}{c|}{w/o ASPP} & \multicolumn{1}{c|}{75.4} & 59.7 & 55.5 & 47.4 & 55.4 & \multicolumn{1}{c|}{73.1} & 15.1 & 14.4 & 19.7 & 40.7 & \multicolumn{1}{c|}{43.6} & 60.4 & 52.5 & 46.8 & \multicolumn{1}{c|}{37.0} & 18.0 & 47.2 & 52.4 & \multicolumn{1}{c|}{22.1} & 68.4 \\
			\multicolumn{1}{c|}{w/o AC} & \multicolumn{1}{c|}{77.4} & 62.2 & \textbf{55.6} & 51.3 & 54.5 & \multicolumn{1}{c|}{75.4} & 17.7 & 15.7 & 22.1 & 42.8 & \multicolumn{1}{c|}{46.5} & 61.6 & 54.9 & 47.8 & \multicolumn{1}{c|}{34.3} & 17.8 & 46.6 & \textbf{59.1} & \multicolumn{1}{c|}{20.9} & 70.9 \\
			\multicolumn{1}{c|}{w/ DPC} & \multicolumn{1}{c|}{77.5} & 60.6 & 53.0 & 50.8 & 52.5 & \multicolumn{1}{c|}{75.8} & 15.1 & 10.7 & 20.3 & 42.7 & \multicolumn{1}{c|}{48.4} & 63.6 & 53.4 & 46.0 & \multicolumn{1}{c|}{36.0} & 17.6 & \textbf{50.0} & 56.7 & \multicolumn{1}{c|}{20.6} & 71.8 \\
			\multicolumn{1}{c|}{w/o LRL} & \multicolumn{1}{c|}{76.8} & \textbf{62.3} & 53.2 & 50.6 & 53.0 & \multicolumn{1}{c|}{75.1} & \textbf{21.3} & \textbf{19.2} & \textbf{27.6} & \textbf{49.3} & \multicolumn{1}{c|}{51.7} & 63.9 & 55.2 & 48.0 & \multicolumn{1}{c|}{33.8} & 20.5 & 48.3 & 57.6 & \multicolumn{1}{c|}{23.9} & 70.8 \\
			\multicolumn{1}{c|}{w/ GAP} & \multicolumn{1}{c|}{77.1} & 61.5 & 54.8 & \textbf{53.1} & 53.9 & \multicolumn{1}{c|}{75.6} & 20.0 & 16.4 & 24.8 & 43.4 & \multicolumn{1}{c|}{46.6} & \textbf{65.7} & \textbf{57.6} & 50.4 & \multicolumn{1}{c|}{36.2} & 16.5 & 48.6 & 56.8 & \multicolumn{1}{c|}{22.6} & 72.5 \\ \midrule
			\multicolumn{1}{c|}{\textbf{Xception-65}} & \multicolumn{1}{c|}{\textbf{78.4}} & \textbf{63.9} & 59.1 & 52.8 & \textbf{59.2} & \multicolumn{1}{c|}{\textbf{76.8}} & 15.0 & 10.6 & 19.8 & 42.4 & \multicolumn{1}{c|}{46.5} & \textbf{65.9} & \textbf{59.1} & 46.1 & \multicolumn{1}{c|}{31.4} & \textbf{19.3} & 50.7 & \textbf{63.6} & \multicolumn{1}{c|}{\textbf{23.8}} & \textbf{72.7} \\
			\multicolumn{1}{c|}{w/o ASPP} & \multicolumn{1}{c|}{75.8} & 61.6 & 56.1 & 51.8 & 54.6 & \multicolumn{1}{c|}{74.1} & 14.3 & 7.7 & 18.8 & 39.0 & \multicolumn{1}{c|}{41.6} & 62.0 & 57.2 & 43.1 & \multicolumn{1}{c|}{29.7} & 15.6 & 46.9 & 60.3 & \multicolumn{1}{c|}{23.4} & 70.6 \\
			\multicolumn{1}{c|}{w/o AC} & \multicolumn{1}{c|}{77.7} & 63.9 & 58.7 & 51.5 & 57.8 & \multicolumn{1}{c|}{75.7} & 14.1 & 14.8 & 19.5 & 41.9 & \multicolumn{1}{c|}{45.1} & 63.9 & 58.3 & 42.9 & \multicolumn{1}{c|}{35.0} & 15.7 & \textbf{51.4} & 60.9 & \multicolumn{1}{c|}{21.4} & 71.8 \\
			\multicolumn{1}{c|}{w/ DPC} & \multicolumn{1}{c|}{77.7} & 62.4 & 55.0 & 50.4 & 54.5 & \multicolumn{1}{c|}{74.7} & 8.9 & 4.8 & 13.2 & 37.1 & \multicolumn{1}{c|}{\textbf{47.7}} & 62.5 & 48.4 & 45.4 & \multicolumn{1}{c|}{30.3} & 17.3 & 47.1 & 59.6 & \multicolumn{1}{c|}{21.9} & 70.7 \\
			\multicolumn{1}{c|}{w/o LRL} & \multicolumn{1}{c|}{77.7} & 64.5 & 58.6 & 49.5 & 57.9 & \multicolumn{1}{c|}{75.9} & 15.1 & 12.0 & 19.9 & 42.1 & \multicolumn{1}{c|}{45.9} & 63.8 & 57.9 & 46.1 & \multicolumn{1}{c|}{\textbf{35.9}} & 18.4 & 46.3 & 63.5 & \multicolumn{1}{c|}{22.0} & 71.4 \\
			\multicolumn{1}{c|}{w/ GAP} & \multicolumn{1}{c|}{78.4} & 63.9 & \textbf{59.4} & \textbf{53.5} & 58.8 & \multicolumn{1}{c|}{76.2} & \textbf{18.8} & \textbf{15.4} & \textbf{23.7} & \textbf{43.7} & \multicolumn{1}{c|}{45.7} & 65.2 & 56.5 & \textbf{48.0} & \multicolumn{1}{c|}{31.5} & 18.8 & 49.4 & 59.1 & \multicolumn{1}{c|}{20.7} & 71.0 \\ \bottomrule
		\end{tabular}
	\end{adjustbox}
	\caption{
		Mean IoU for clean and corrupted variants of the validation set of the Cityscapes dataset for several network backbones of the DeepLabv3+ architecture and respective architectural ablations. 
		Every mIoU is averaged over all available severity levels, except for corruptions of category noise where only the first three severity levels are considered. 
		The standard deviation for image corruptions of category noise is $0.2$ or less.
		Highest mIoU per corruption is bold.}
	\label{tab:miou_cs_allbackbones_and_ablations}
\end{table*}

\begin{table*}[h]
	\begin{adjustbox}{width=\textwidth}
		\begin{tabular}{@{}cccccccccccccccccccc@{}}
			\toprule
			& \multicolumn{5}{c}{\textbf{Blur}} & \multicolumn{5}{c}{\textbf{Noise}} & \multicolumn{4}{c}{\textbf{Digital}} & \multicolumn{4}{c}{\textbf{Weather}} & \textbf{} \\ \midrule
			\begin{tabular}[c]{@{}c@{}}Deeplab-v3+ \\ Backbone\end{tabular} & Motion & Defocus & \begin{tabular}[c]{@{}c@{}}Frosted \\ Glass\end{tabular} & Gaussian & \multicolumn{1}{c|}{PSF} & Gaussian & Impulse & Shot & Speckle & \multicolumn{1}{c|}{Intensity} & Brightness & Contrast & Saturate & \multicolumn{1}{c|}{JPEG} & Snow & Spatter & Fog & \multicolumn{1}{c|}{Frost} & \begin{tabular}[c]{@{}c@{}}Geometric\\ Distortion\end{tabular} \\ \midrule
			\textbf{MobileNet-V2} & 100.0 & 100.0 & 100.0 & 100.0 & \multicolumn{1}{c|}{100.0} & 100.0 & 100.0 & 100.0 & 100.0 & \multicolumn{1}{c|}{100.0} & 100.0 & 100.0 & 100.0 & \multicolumn{1}{c|}{100.0} & 100.0 & 100.0 & 100.0 & \multicolumn{1}{c|}{100.0} & 100.0 \\
			w/o ASPP & \textbf{117.2} & \textbf{116.8} & \textbf{111.5} & \textbf{115.0} & \multicolumn{1}{c|}{\textbf{124.2}} & 98.6 & 98.1 & 97.6 & 96.1 & \multicolumn{1}{c|}{97.9} & \textbf{120.6} & \textbf{120.0} & \textbf{107.1} & \multicolumn{1}{c|}{\textbf{109.0}} & 100.7 & \textbf{103.5} & \textbf{118.2} & \multicolumn{1}{c|}{104.0} & \textbf{110.4} \\
			w/o AC & 103.1 & 99.9 & 104.6 & 99.8 & \multicolumn{1}{c|}{102.2} & 103.0 & 99.7 & 102.2 & 96.4 & \multicolumn{1}{c|}{96.9} & 104.0 & 101.1 & 101.5 & \multicolumn{1}{c|}{98.8} & 104.2 & 90.8 & 104.1 & \multicolumn{1}{c|}{100.7} & 105.7 \\
			w/ DPC & 108.9 & 113.4 & 102.9 & 110.4 & \multicolumn{1}{c|}{104.2} & 103.1 & 102.3 & 102.9 & \textbf{100.6} & \multicolumn{1}{c|}{99.3} & 113.9 & 115.3 & 103.4 & \multicolumn{1}{c|}{104.3} & \textbf{104.5} & 93.7 & 106.7 & \multicolumn{1}{c|}{103.9} & 109.4 \\
			w/o LRL & 108.2 & 110.0 & 101.7 & 108.0 & \multicolumn{1}{c|}{105.3} & \textbf{104.8} & \textbf{104.8} & \textbf{104.4} & 96.3 & \multicolumn{1}{c|}{\textbf{100.0}} & 105.5 & 106.1 & 100.3 & \multicolumn{1}{c|}{101.2} & 103.8 & 99.1 & 104.8 & \multicolumn{1}{c|}{103.1} & 113.2 \\
			w/ GAP & 102.8 & 96.9 & 103.6 & 94.8 & \multicolumn{1}{c|}{102.2} & 97.7 & 95.8 & 95.5 & 88.7 & \multicolumn{1}{c|}{92.3} & 99.9 & 98.9 & 95.7 & \multicolumn{1}{c|}{102.1} & 101.1 & 92.8 & 104.3 & \multicolumn{1}{c|}{\textbf{105.6}} & 98.4 \\ \midrule
			\textbf{ResNet-50} & 100.0 & 100.0 & 100.0 & 100.0 & \multicolumn{1}{c|}{100.0} & 100.0 & 100.0 & 100.0 & 100.0 & \multicolumn{1}{c|}{100.0} & 100.0 & 100.0 & 100.0 & \multicolumn{1}{c|}{100.0} & 100.0 & 100.0 & 100.0 & \multicolumn{1}{c|}{100.0} & 100.0 \\
			w/o ASPP & \textbf{115.1} & \textbf{113.6} & \textbf{111.3} & \textbf{113.5} & \multicolumn{1}{c|}{\textbf{120.3}} & 96.2 & 95.5 & 95.8 & 99.9 & \multicolumn{1}{c|}{96.4} & \textbf{119.0} & \textbf{113.9} & \textbf{107.3} & \multicolumn{1}{c|}{102.9} & 100.1 & \textbf{105.8} & \textbf{130.1} & \multicolumn{1}{c|}{\textbf{105.3}} & \textbf{111.8} \\
			w/o AC & 104.4 & 108.1 & 99.7 & 108.5 & \multicolumn{1}{c|}{104.2} & \textbf{104.3} & \textbf{101.2} & \textbf{105.4} & \textbf{108.1} & \multicolumn{1}{c|}{\textbf{107.5}} & 103.2 & 107.4 & 104.1 & \multicolumn{1}{c|}{99.8} & \textbf{102.6} & 102.0 & 109.0 & \multicolumn{1}{c|}{103.5} & 108.6 \\
			w/ DPC & 103.6 & 104.3 & 100.5 & 103.7 & \multicolumn{1}{c|}{102.3} & 95.5 & 94.2 & 96.0 & 97.1 & \multicolumn{1}{c|}{98.3} & 108.6 & 102.4 & 99.7 & \multicolumn{1}{c|}{\textbf{103.3}} & 100.7 & 101.0 & 99.1 & \multicolumn{1}{c|}{103.1} & 103.0 \\
			w/o LRL & 101.5 & 104.9 & 101.5 & 105.1 & \multicolumn{1}{c|}{104.2} & 98.5 & 97.7 & 98.6 & 99.1 & \multicolumn{1}{c|}{94.4} & 104.6 & 106.8 & 101.8 & \multicolumn{1}{c|}{98.5} & 100.1 & 101.0 & 101.3 & \multicolumn{1}{c|}{102.2} & 105.1 \\
			w/ GAP & 104.3 & 102.1 & 102.7 & 100.7 & \multicolumn{1}{c|}{98.3} & 101.1 & 99.3 & 100.6 & 99.7 & \multicolumn{1}{c|}{94.7} & 101.2 & 107.3 & 100.3 & \multicolumn{1}{c|}{98.8} & 101.8 & 102.8 & 103.3 & \multicolumn{1}{c|}{103.4} & 103.2 \\ \midrule
			\textbf{ResNet-101} & 100.0 & 100.0 & 100.0 & 100.0 & \multicolumn{1}{c|}{100.0} & 100.0 & 100.0 & 100.0 & 100.0 & \multicolumn{1}{c|}{100.0} & 100.0 & 100.0 & 100.0 & \multicolumn{1}{c|}{100.0} & 100.0 & 100.0 & 100.0 & \multicolumn{1}{c|}{100.0} & 100.0 \\
			w/o ASPP & \textbf{113.2} & \textbf{113.1} & \textbf{110.5} & \textbf{113.3} & \multicolumn{1}{c|}{\textbf{125.6}} & 104.3 & 104.8 & 104.4 & \textbf{109.9} & \multicolumn{1}{c|}{112.2} & \textbf{117.6} & \textbf{113.4} & \textbf{108.3} & \multicolumn{1}{c|}{106.7} & 98.4 & \textbf{108.1} & \textbf{116.2} & \multicolumn{1}{c|}{99.4} & 111.0 \\
			w/o AC & 103.1 & 108.6 & 102.2 & 108.0 & \multicolumn{1}{c|}{107.5} & 105.6 & 103.3 & 105.7 & 108.6 & \multicolumn{1}{c|}{118.5} & 114.2 & 103.1 & 104.2 & \multicolumn{1}{c|}{105.2} & 102.4 & 105.9 & 100.3 & \multicolumn{1}{c|}{103.5} & \textbf{113.9} \\
			w/ DPC & 101.5 & 106.5 & 101.9 & 105.8 & \multicolumn{1}{c|}{99.8} & 101.8 & 102.1 & 100.8 & 100.8 & \multicolumn{1}{c|}{107.2} & 113.5 & 100.4 & 103.0 & \multicolumn{1}{c|}{\textbf{110.5}} & 102.1 & 107.2 & 98.0 & \multicolumn{1}{c|}{\textbf{104.4}} & 103.5 \\
			w/o LRL & 101.2 & 103.9 & 100.5 & 103.8 & \multicolumn{1}{c|}{103.8} & 104.8 & 106.4 & 105.1 & 105.4 & \multicolumn{1}{c|}{\textbf{115.9}} & 105.6 & 93.1 & 101.1 & \multicolumn{1}{c|}{102.8} & \textbf{103.1} & 106.7 & 99.5 & \multicolumn{1}{c|}{102.5} & 101.5 \\
			w/ GAP & 101.1 & 98.6 & 98.7 & 98.9 & \multicolumn{1}{c|}{97.1} & \textbf{105.8} & \textbf{107.5} & \textbf{105.7} & 107.8 & \multicolumn{1}{c|}{111.8} & 108.9 & 97.1 & 103.7 & \multicolumn{1}{c|}{101.7} & 100.5 & 105.6 & 105.8 & \multicolumn{1}{c|}{100.2} & 102.2 \\ \midrule
			\textbf{Xception-41} & 100.0 & 100.0 & 100.0 & 100.0 & \multicolumn{1}{c|}{100.0} & 100.0 & 100.0 & 100.0 & 100.0 & \multicolumn{1}{c|}{100.0} & 100.0 & 100.0 & 100.0 & \multicolumn{1}{c|}{100.0} & 100.0 & 100.0 & 100.0 & \multicolumn{1}{c|}{\textbf{100.0}} & 100.0 \\
			w/o ASPP & \textbf{105.1} & 98.5 & \textbf{107.5} & 98.5 & \multicolumn{1}{c|}{\textbf{112.2}} & \textbf{102.4} & 103.5 & \textbf{102.4} & \textbf{105.3} & \multicolumn{1}{c|}{\textbf{109.8}} & \textbf{108.9} & \textbf{110.2} & 110.2 & \multicolumn{1}{c|}{102.3} & 100.2 & 98.8 & \textbf{112.4} & \multicolumn{1}{c|}{98.2} & \textbf{116.9} \\
			w/o AC & 98.5 & 98.3 & 99.5 & 100.4 & \multicolumn{1}{c|}{102.6} & 99.1 & 102.0 & 99.3 & 101.6 & \multicolumn{1}{c|}{104.0} & 105.6 & 104.7 & 108.2 & \multicolumn{1}{c|}{106.8} & 100.5 & \textbf{100.0} & 96.6 & \multicolumn{1}{c|}{99.6} & 107.5 \\
			w/ DPC & 102.8 & \textbf{104.2} & 100.5 & \textbf{104.9} & \multicolumn{1}{c|}{101.0} & 102.4 & \textbf{108.0} & 101.6 & 101.9 & \multicolumn{1}{c|}{100.4} & 100.0 & 108.1 & \textbf{111.9} & \multicolumn{1}{c|}{104.1} & 100.8 & 93.6 & 102.1 & \multicolumn{1}{c|}{100.0} & 104.4 \\
			w/o LRL & 98.2 & 103.6 & 100.9 & 103.7 & \multicolumn{1}{c|}{103.9} & 94.9 & 97.7 & 92.3 & 90.1 & \multicolumn{1}{c|}{94.0} & 99.2 & 103.9 & 107.7 & \multicolumn{1}{c|}{\textbf{107.6}} & 97.2 & 96.9 & 100.0 & \multicolumn{1}{c|}{95.9} & 108.0 \\
			w/ GAP & 100.3 & 100.2 & 95.7 & 101.7 & \multicolumn{1}{c|}{101.8} & 96.5 & 101.1 & 95.9 & 100.5 & \multicolumn{1}{c|}{103.9} & 94.3 & 98.5 & 102.6 & \multicolumn{1}{c|}{103.8} & \textbf{102.1} & 96.2 & 101.9 & \multicolumn{1}{c|}{97.5} & 101.6 \\ \midrule
			\textbf{Xception-65} & 100.0 & 100.0 & 100.0 & 100.0 & \multicolumn{1}{c|}{100.0} & 100.0 & 100.0 & 100.0 & 100.0 & \multicolumn{1}{c|}{100.0} & 100.0 & 100.0 & 100.0 & \multicolumn{1}{c|}{100.0} & 100.0 & 100.0 & 100.0 & \multicolumn{1}{c|}{100.0} & 100.0 \\
			w/o ASPP & \textbf{106.4} & 107.2 & 102.0 & 111.4 & \multicolumn{1}{c|}{\textbf{111.7}} & 100.8 & 103.3 & 101.4 & 105.9 & \multicolumn{1}{c|}{\textbf{109.3}} & \textbf{111.2} & 104.7 & 105.5 & \multicolumn{1}{c|}{\textbf{102.5}} & \textbf{104.5} & 107.7 & 109.1 & \multicolumn{1}{c|}{100.6} & \textbf{107.5} \\
			w/o AC & 100.1 & 101.0 & 102.6 & 103.3 & \multicolumn{1}{c|}{104.8} & 101.1 & 95.3 & 100.5 & 100.9 & \multicolumn{1}{c|}{102.7} & 105.6 & 102.1 & \textbf{106.0} & \multicolumn{1}{c|}{94.8} & 104.4 & 98.7 & 107.4 & \multicolumn{1}{c|}{103.1} & 103.1 \\
			w/ DPC & 104.0 & \textbf{109.8} & 105.1 & \textbf{111.6} & \multicolumn{1}{c|}{108.9} & \textbf{107.2} & \textbf{106.5} & \textbf{108.3} & \textbf{109.3} & \multicolumn{1}{c|}{97.9} & 109.9 & \textbf{126.3} & 101.4 & \multicolumn{1}{c|}{101.7} & 102.4 & 107.4 & 111.0 & \multicolumn{1}{c|}{102.5} & 107.1 \\
			w/o LRL & 98.3 & 101.1 & \textbf{107.0} & 103.2 & \multicolumn{1}{c|}{104.0} & 99.8 & 98.4 & 99.9 & 100.6 & \multicolumn{1}{c|}{101.3} & 106.1 & 103.1 & 99.9 & \multicolumn{1}{c|}{93.4} & 101.1 & \textbf{108.9} & 100.3 & \multicolumn{1}{c|}{102.4} & 104.8 \\
			w/ GAP & 100.0 & 99.1 & 98.5 & 101.0 & \multicolumn{1}{c|}{102.5} & 95.5 & 94.6 & 95.2 & 97.8 & \multicolumn{1}{c|}{101.5} & 101.9 & 106.6 & 96.5 & \multicolumn{1}{c|}{99.9} & 100.6 & 102.6 & \textbf{112.4} & \multicolumn{1}{c|}{\textbf{104.2}} & 106.3 \\ \midrule
			\textbf{Xception-71} & 100.0 & 100.0 & 100.0 & 100.0 & \multicolumn{1}{c|}{100.0} & 100.0 & 100.0 & 100.0 & 100.0 & \multicolumn{1}{c|}{100.0} & 100.0 & 100.0 & 100.0 & \multicolumn{1}{c|}{100.0} & 100.0 & 100.0 & 100.0 & \multicolumn{1}{c|}{\textbf{100.0}} & 100.0 \\
			w/o ASPP & \textbf{109.6} & 103.6 & 101.1 & 105.2 & \multicolumn{1}{c|}{\textbf{115.1}} & 95.7 & 95.6 & 96.5 & 102.4 & \multicolumn{1}{c|}{110.9} & 114.6 & 106.0 & \textbf{108.3} & \multicolumn{1}{c|}{100.5} & 101.6 & 102.9 & \textbf{116.1} & \multicolumn{1}{c|}{97.9} & \textbf{105.9} \\
			w/o AC & 105.3 & \textbf{107.6} & 100.4 & \textbf{105.7} & \multicolumn{1}{c|}{100.9} & \textbf{108.4} & 105.8 & 110.2 & \textbf{114.3} & \multicolumn{1}{c|}{114.0} & 101.4 & \textbf{107.4} & 102.0 & \multicolumn{1}{c|}{99.1} & 100.9 & 100.8 & 108.2 & \multicolumn{1}{c|}{98.2} & 99.7 \\
			w/ DPC & 103.7 & 103.7 & 98.8 & 105.6 & \multicolumn{1}{c|}{97.9} & 108.9 & \textbf{109.0} & \textbf{110.8} & 113.9 & \multicolumn{1}{c|}{\textbf{115.5}} & 110.2 & 98.2 & 103.4 & \multicolumn{1}{c|}{\textbf{113.7}} & \textbf{105.5} & \textbf{103.7} & 100.2 & \multicolumn{1}{c|}{99.2} & 96.0 \\
			w/o LRL & 99.8 & 94.0 & \textbf{102.8} & 95.4 & \multicolumn{1}{c|}{98.7} & 101.1 & 101.7 & 101.5 & 99.9 & \multicolumn{1}{c|}{100.5} & \textbf{111.0} & 98.7 & 105.4 & \multicolumn{1}{c|}{106.9} & 102.4 & 103.6 & 99.3 & \multicolumn{1}{c|}{98.5} & 98.7 \\
			w/ GAP & 99.8 & 97.8 & 91.8 & 99.3 & \multicolumn{1}{c|}{94.0} & 106.1 & 102.8 & 106.9 & 107.3 & \multicolumn{1}{c|}{109.1} & 100.1 & 96.2 & 97.7 & \multicolumn{1}{c|}{99.3} & 102.6 & 98.9 & 105.4 & \multicolumn{1}{c|}{99.0} & 90.8 \\ \bottomrule
		\end{tabular}

	\end{adjustbox}
	\caption{
		CD for corrupted variants of the validation set of the Cityscapes dataset for several network backbones of the DeepLabv3+ architecture and respective architectural ablations. 
		Highest CD per corruption is bold.}
	\label{tab:CD_cs_allbackbones_and_ablations}
\end{table*}

\begin{table*}[h]
	\begin{adjustbox}{width=\textwidth}
	\begin{tabular}{@{}cccccccccccccccccccc@{}}
		\toprule
		& \multicolumn{5}{c}{\textbf{Blur}} & \multicolumn{5}{c}{\textbf{Noise}} & \multicolumn{4}{c}{\textbf{Digital}} & \multicolumn{4}{c}{\textbf{Weather}} & \textbf{} \\ \midrule
		\multicolumn{1}{c|}{\begin{tabular}[c]{@{}c@{}}Deeplab-v3+ \\ Backbone\end{tabular}} & Motion & Defocus & \begin{tabular}[c]{@{}c@{}}Frosted \\ Glass\end{tabular} & Gaussian & \multicolumn{1}{c|}{PSF} & Gaussian & Impulse & Shot & Speckle & \multicolumn{1}{c|}{Intensity} & Brightness & Contrast & Saturate & \multicolumn{1}{c|}{JPEG} & Snow & Spatter & Fog & \multicolumn{1}{c|}{Frost} & \begin{tabular}[c]{@{}c@{}}Geometric\\ Distortion\end{tabular} \\ \midrule
		\multicolumn{1}{c|}{\textbf{MobileNet-V2}} & 100.0 & 100.0 & 100.0 & 100.0 & \multicolumn{1}{c|}{100.0} & 100.0 & 100.0 & 100.0 & 100.0 & \multicolumn{1}{c|}{100.0} & 100.0 & 100.0 & 100.0 & \multicolumn{1}{c|}{100.0} & 100.0 & \textbf{100.0} & 100.0 & \multicolumn{1}{c|}{100.0} & 100.0 \\
		\multicolumn{1}{c|}{w/o ASPP} & 104.8 & 106.3 & 96.8 & 102.1 & \multicolumn{1}{c|}{101.3} & 87.2 & 86.3 & 85.7 & 81.2 & \multicolumn{1}{c|}{80.8} & 114.0 & 114.0 & 94.1 & \multicolumn{1}{c|}{98.7} & 88.9 & 84.9 & 109.9 & \multicolumn{1}{c|}{93.0} & 46.3 \\
		\multicolumn{1}{c|}{w/o AC} & 103.2 & 96.1 & \textbf{106.4} & 95.8 & \multicolumn{1}{c|}{89.0} & 103.0 & 98.3 & 101.9 & 93.1 & \multicolumn{1}{c|}{93.1} & 105.5 & 99.0 & 100.5 & \multicolumn{1}{c|}{96.3} & 104.9 & 80.5 & 105.4 & \multicolumn{1}{c|}{99.6} & 117.2 \\
		\multicolumn{1}{c|}{w/ DPC} & \textbf{119.9} & \textbf{127.8} & 104.3 & \textbf{121.3} & \multicolumn{1}{c|}{\textbf{152.5}} & 103.8 & 102.6 & 103.4 & \textbf{100.2} & \multicolumn{1}{c|}{97.9} & \textbf{130.9} & \textbf{130.5} & \textbf{104.7} & \multicolumn{1}{c|}{\textbf{106.0}} & \textbf{105.9} & 87.1 & \textbf{112.5} & \multicolumn{1}{c|}{105.1} & 143.2 \\
		\multicolumn{1}{c|}{w/o LRL} & 115.5 & 118.1 & 100.0 & 113.8 & \multicolumn{1}{c|}{141.6} & \textbf{105.4} & \textbf{105.4} & \textbf{104.8} & 92.8 & \multicolumn{1}{c|}{97.9} & 108.5 & 109.2 & 98.1 & \multicolumn{1}{c|}{100.0} & 104.1 & 95.5 & 106.5 & \multicolumn{1}{c|}{103.0} & \textbf{155.2} \\
		\multicolumn{1}{c|}{w/ GAP} & 103.5 & 90.4 & 105.0 & 85.8 & \multicolumn{1}{c|}{100.6} & 95.8 & 93.0 & 92.6 & 81.9 & \multicolumn{1}{c|}{86.1} & 96.7 & 95.2 & 91.1 & \multicolumn{1}{c|}{102.0} & 100.5 & 84.7 & 106.6 & \multicolumn{1}{c|}{\textbf{107.3}} & 81.9 \\ \midrule
		\multicolumn{1}{c|}{\textbf{ResNet-50}} & 100.0 & 100.0 & 100.0 & 100.0 & \multicolumn{1}{c|}{100.0} & 100.0 & 100.0 & 100.0 & 100.0 & \multicolumn{1}{c|}{100.0} & 100.0 & 100.0 & 100.0 & \multicolumn{1}{c|}{100.0} & 100.0 & 100.0 & 100.0 & \multicolumn{1}{c|}{100.0} & 100.0 \\
		\multicolumn{1}{c|}{w/o ASPP} & 106.1 & 103.7 & 102.8 & 102.9 & \multicolumn{1}{c|}{98.2} & 87.5 & 86.6 & 86.5 & 88.4 & \multicolumn{1}{c|}{83.3} & \textbf{115.1} & 105.2 & 97.5 & \multicolumn{1}{c|}{93.8} & 92.1 & 94.9 & \textbf{139.3} & \multicolumn{1}{c|}{98.0} & 78.0 \\
		\multicolumn{1}{c|}{w/o AC} & 106.7 & \textbf{114.5} & 97.5 & \textbf{115.9} & \multicolumn{1}{c|}{126.3} & \textbf{104.9} & \textbf{100.7} & \textbf{106.5} & \textbf{111.0} & \multicolumn{1}{c|}{\textbf{110.0}} & 103.9 & 112.5 & \textbf{105.0} & \multicolumn{1}{c|}{98.4} & 102.6 & 101.7 & 116.3 & \multicolumn{1}{c|}{103.9} & \textbf{128.5} \\
		\multicolumn{1}{c|}{w/ DPC} & \textbf{110.3} & 111.1 & 102.1 & 110.2 & \multicolumn{1}{c|}{\textbf{154.5}} & 94.5 & 92.8 & 95.2 & 96.3 & \multicolumn{1}{c|}{98.3} & 121.6 & 106.6 & 100.5 & \multicolumn{1}{c|}{\textbf{105.5}} & 101.5 & 102.8 & 99.8 & \multicolumn{1}{c|}{\textbf{105.1}} & 118.1 \\
		\multicolumn{1}{c|}{w/o LRL} & 98.2 & 106.0 & 99.5 & 106.5 & \multicolumn{1}{c|}{106.5} & 96.7 & 95.5 & 96.8 & 96.5 & \multicolumn{1}{c|}{89.5} & 105.4 & 109.8 & 100.3 & \multicolumn{1}{c|}{95.8} & 98.7 & 99.0 & 98.3 & \multicolumn{1}{c|}{101.4} & 108.8 \\
		\multicolumn{1}{c|}{w/ GAP} & 109.7 & 104.3 & \textbf{104.7} & 101.3 & \multicolumn{1}{c|}{73.1} & 101.4 & 99.0 & 100.7 & 99.4 & \multicolumn{1}{c|}{91.9} & 102.5 & \textbf{114.9} & 100.3 & \multicolumn{1}{c|}{98.1} & \textbf{102.3} & \textbf{104.5} & 106.9 & \multicolumn{1}{c|}{104.9} & 113.4 \\ \midrule
		\multicolumn{1}{c|}{\textbf{ResNet-101}} & 100.0 & 100.0 & 100.0 & 100.0 & \multicolumn{1}{c|}{100.0} & 100.0 & 100.0 & 100.0 & 100.0 & \multicolumn{1}{c|}{100.0} & 100.0 & 100.0 & 100.0 & \multicolumn{1}{c|}{100.0} & 100.0 & 100.0 & 100.0 & \multicolumn{1}{c|}{100.0} & 100.0 \\
		\multicolumn{1}{c|}{w/o ASPP} & 96.5 & 98.6 & 98.3 & 98.3 & \multicolumn{1}{c|}{119.0} & 96.4 & 97.0 & 96.1 & 100.6 & \multicolumn{1}{c|}{103.6} & 106.5 & \textbf{100.5} & 96.7 & \multicolumn{1}{c|}{95.5} & 88.7 & 94.0 & 105.9 & \multicolumn{1}{c|}{88.1} & 63.9 \\
		\multicolumn{1}{c|}{w/o AC} & 99.3 & 111.4 & 99.2 & 110.4 & \multicolumn{1}{c|}{\textbf{125.1}} & 105.5 & 102.3 & 105.5 & 110.1 & \multicolumn{1}{c|}{\textbf{126.1}} & 124.6 & 100.1 & 103.0 & \multicolumn{1}{c|}{104.7} & 101.1 & 105.8 & 94.3 & \multicolumn{1}{c|}{102.4} & \textbf{138.4} \\
		\multicolumn{1}{c|}{w/ DPC} & 103.0 & \textbf{113.2} & \textbf{103.1} & \textbf{112.2} & \multicolumn{1}{c|}{92.1} & 102.3 & 102.7 & 101.0 & 101.0 & \multicolumn{1}{c|}{111.4} & \textbf{130.4} & 100.4 & 104.7 & \multicolumn{1}{c|}{\textbf{116.3}} & 102.7 & \textbf{112.6} & 95.5 & \multicolumn{1}{c|}{\textbf{106.1}} & 112.9 \\
		\multicolumn{1}{c|}{w/o LRL} & 99.5 & 105.5 & 98.9 & 105.4 & \multicolumn{1}{c|}{120.0} & 105.7 & 107.9 & 106.1 & 107.1 & \multicolumn{1}{c|}{124.1} & 109.5 & 83.7 & 100.3 & \multicolumn{1}{c|}{102.9} & \textbf{103.3} & 110.0 & 96.3 & \multicolumn{1}{c|}{102.4} & 98.4 \\
		\multicolumn{1}{c|}{w/ GAP} & \textbf{103.8} & 98.2 & 98.5 & 98.9 & \multicolumn{1}{c|}{74.0} & \textbf{108.3} & \textbf{110.6} & \textbf{108.3} & \textbf{112.8} & \multicolumn{1}{c|}{119.8} & 121.7 & 95.1 & \textbf{106.8} & \multicolumn{1}{c|}{103.4} & 101.0 & 110.8 & \textbf{112.9} & \multicolumn{1}{c|}{100.7} & 112.2 \\ \midrule
		\multicolumn{1}{c|}{\textbf{Xception-41}} & 100.0 & 100.0 & 100.0 & 100.0 & \multicolumn{1}{c|}{100.0} & 100.0 & 100.0 & 100.0 & 100.0 & \multicolumn{1}{c|}{100.0} & 100.0 & 100.0 & 100.0 & \multicolumn{1}{c|}{100.0} & 100.0 & \textbf{100.0} & 100.0 & \multicolumn{1}{c|}{\textbf{100.0}} & 100.0 \\
		\multicolumn{1}{c|}{w/o ASPP} & 97.7 & 86.9 & \textbf{104.9} & 86.8 & \multicolumn{1}{c|}{\textbf{133.6}} & 99.4 & 100.9 & 99.1 & 101.9 & \multicolumn{1}{c|}{\textbf{109.2}} & 106.3 & 109.9 & 109.9 & \multicolumn{1}{c|}{97.7} & 96.4 & 90.5 & \textbf{114.4} & \multicolumn{1}{c|}{93.4} & \textbf{146.1} \\
		\multicolumn{1}{c|}{w/o AC} & 94.3 & 95.3 & 97.8 & 99.4 & \multicolumn{1}{c|}{116.9} & 98.3 & 102.2 & 98.5 & 101.6 & \multicolumn{1}{c|}{106.0} & \textbf{112.1} & 108.2 & 114.0 & \multicolumn{1}{c|}{\textbf{109.8}} & 100.1 & 98.9 & 91.2 & \multicolumn{1}{c|}{98.9} & 135.6 \\
		\multicolumn{1}{c|}{w/ DPC} & \textbf{104.7} & \textbf{106.9} & 99.8 & \textbf{108.2} & \multicolumn{1}{c|}{96.2} & \textbf{102.7} & \textbf{110.4} & \textbf{101.7} & \textbf{102.2} & \multicolumn{1}{c|}{99.7} & 97.8 & \textbf{115.3} & \textbf{120.9} & \multicolumn{1}{c|}{105.6} & 100.5 & 88.1 & 102.8 & \multicolumn{1}{c|}{99.5} & 118.1 \\
		\multicolumn{1}{c|}{w/o LRL} & 89.6 & 102.9 & 98.0 & 103.0 & \multicolumn{1}{c|}{97.2} & 91.4 & 95.3 & 87.5 & 80.8 & \multicolumn{1}{c|}{86.0} & 91.0 & 103.3 & 110.5 & \multicolumn{1}{c|}{109.4} & 94.5 & 91.5 & 95.2 & \multicolumn{1}{c|}{92.6} & 124.8 \\
		\multicolumn{1}{c|}{w/ GAP} & 96.7 & 97.6 & 89.7 & 100.5 & \multicolumn{1}{c|}{88.3} & 94.1 & 100.4 & 93.1 & 99.0 & \multicolumn{1}{c|}{104.6} & 80.9 & 93.8 & 102.4 & \multicolumn{1}{c|}{104.3} & \textbf{101.8} & 91.4 & 100.9 & \multicolumn{1}{c|}{95.4} & 95.5 \\ \midrule
		\multicolumn{1}{c|}{\textbf{Xception-65}} & 100.0 & 100.0 & 100.0 & 100.0 & \multicolumn{1}{c|}{100.0} & 100.0 & 100.0 & 100.0 & 100.0 & \multicolumn{1}{c|}{100.0} & 100.0 & 100.0 & 100.0 & \multicolumn{1}{c|}{100.0} & 100.0 & 100.0 & 100.0 & \multicolumn{1}{c|}{100.0} & 100.0 \\
		\multicolumn{1}{c|}{w/o ASPP} & 98.0 & 101.7 & 93.6 & \textbf{110.6} & \multicolumn{1}{c|}{106.5} & 97.0 & 100.4 & 97.4 & 102.1 & \multicolumn{1}{c|}{\textbf{107.4}} & 109.7 & 96.4 & 101.1 & \multicolumn{1}{c|}{98.1} & 101.7 & 104.3 & 104.8 & \multicolumn{1}{c|}{96.1} & 90.2 \\
		\multicolumn{1}{c|}{w/o AC} & 95.0 & 98.1 & 101.8 & 103.1 & \multicolumn{1}{c|}{121.2} & 100.2 & 92.7 & 99.3 & 99.3 & \multicolumn{1}{c|}{102.2} & 109.3 & 100.5 & \textbf{107.7} & \multicolumn{1}{c|}{90.7} & \textbf{104.7} & 94.9 & 113.0 & \multicolumn{1}{c|}{103.0} & 101.7 \\
		\multicolumn{1}{c|}{w/ DPC} & \textbf{105.0} & \textbf{117.0} & 106.5 & 120.8 & \multicolumn{1}{c|}{\textbf{181.5}} & \textbf{108.5} & \textbf{107.4} & \textbf{110.2} & \textbf{112.8} & \multicolumn{1}{c|}{94.2} & \textbf{121.2} & \textbf{152.0} & 100.0 & \multicolumn{1}{c|}{\textbf{100.9}} & 102.1 & 110.6 & 122.2 & \multicolumn{1}{c|}{102.2} & 121.3 \\
		\multicolumn{1}{c|}{w/o LRL} & 90.4 & 98.3 & \textbf{109.8} & 102.7 & \multicolumn{1}{c|}{108.8} & 98.6 & 96.8 & 98.5 & 98.8 & \multicolumn{1}{c|}{99.7} & 110.2 & 102.5 & 97.5 & \multicolumn{1}{c|}{88.8} & 100.1 & \textbf{113.1} & 95.5 & \multicolumn{1}{c|}{101.9} & 109.2 \\
		\multicolumn{1}{c|}{w/ GAP} & 99.7 & 97.8 & 97.0 & 102.0 & \multicolumn{1}{c|}{132.6} & 93.9 & 92.8 & 93.4 & 96.4 & \multicolumn{1}{c|}{102.4} & 104.8 & 113.8 & 94.1 & \multicolumn{1}{c|}{99.7} & 100.7 & 104.4 & \textbf{130.2} & \multicolumn{1}{c|}{\textbf{105.7}} & \textbf{129.1} \\ \midrule
		\multicolumn{1}{c|}{\textbf{Xception-71}} & 100.0 & 100.0 & 100.0 & 100.0 & \multicolumn{1}{c|}{100.0} & 100.0 & 100.0 & 100.0 & 100.0 & \multicolumn{1}{c|}{100.0} & 100.0 & 100.0 & 100.0 & \multicolumn{1}{c|}{100.0} & 100.0 & 100.0 & 100.0 & \multicolumn{1}{c|}{100.0} & \textbf{100.0} \\
		\multicolumn{1}{c|}{w/o ASPP} & 91.2 & 81.3 & 84.3 & 85.5 & \multicolumn{1}{c|}{49.1} & 86.9 & 87.3 & 87.3 & 91.1 & \multicolumn{1}{c|}{102.4} & 99.6 & 88.8 & 99.0 & \multicolumn{1}{c|}{88.4} & 94.2 & 88.3 & 107.3 & \multicolumn{1}{c|}{89.1} & 60.7 \\
		\multicolumn{1}{c|}{w/o AC} & 107.9 & \textbf{112.4} & 97.9 & 108.1 & \multicolumn{1}{c|}{76.3} & 110.0 & 106.5 & 112.6 & 120.5 & \multicolumn{1}{c|}{121.8} & 97.1 & \textbf{111.5} & 101.0 & \multicolumn{1}{c|}{96.7} & 99.9 & 98.7 & \textbf{114.9} & \multicolumn{1}{c|}{96.3} & 89.1 \\
		\multicolumn{1}{c|}{w/ DPC} & \textbf{110.3} & 109.0 & 98.5 & \textbf{113.1} & \multicolumn{1}{c|}{\textbf{84.6}} & \textbf{112.2} & \textbf{112.1} & \textbf{115.0} & \textbf{122.3} & \multicolumn{1}{c|}{\textbf{127.6}} & \textbf{132.1} & 97.1 & 106.2 & \multicolumn{1}{c|}{\textbf{121.8}} & \textbf{107.7} & \textbf{107.1} & 101.6 & \multicolumn{1}{c|}{\textbf{99.2}} & 86.8 \\
		\multicolumn{1}{c|}{w/o LRL} & 94.2 & 82.5 & \textbf{102.1} & 85.8 & \multicolumn{1}{c|}{53.0} & 100.3 & 101.1 & 100.7 & 97.7 & \multicolumn{1}{c|}{98.1} & 125.9 & 93.4 & \textbf{106.6} & \multicolumn{1}{c|}{108.7} & 102.0 & 103.6 & 92.9 & \multicolumn{1}{c|}{96.7} & 85.1 \\
		\multicolumn{1}{c|}{w/ GAP} & 98.9 & 94.6 & 85.0 & 98.1 & \multicolumn{1}{c|}{33.6} & 108.0 & 103.5 & 109.2 & 111.2 & \multicolumn{1}{c|}{115.6} & 99.5 & 91.8 & 95.8 & \multicolumn{1}{c|}{98.8} & 103.3 & 97.8 & 112.7 & \multicolumn{1}{c|}{98.5} & 64.1 \\ \bottomrule
	\end{tabular}

	\end{adjustbox}
	\caption{
		Relative CD for corrupted variants of the validation set of the Cityscapes dataset for several network backbones of the DeepLabv3+ architecture and respective architectural ablations. 
		Highest rCD per corruption is bold.}
	\label{tab:rCD_cs_allbackbones_and_ablations}
\end{table*}

\begin{table*}[h]
	\begin{adjustbox}{width=\textwidth}
		\begin{tabular}{@{}cccccccccccccccccccc@{}}
			\toprule
			\multicolumn{2}{c}{} & \multicolumn{4}{c}{\textbf{Blur}} & \multicolumn{5}{c}{\textbf{Noise}} & \multicolumn{4}{c}{\textbf{Digital}} & \multicolumn{4}{c}{\textbf{Weather}} & \textbf{} \\ \midrule
			\multicolumn{1}{c|}{\begin{tabular}[c]{@{}c@{}}Deeplab-v3+ \\ Backbone\end{tabular}} & \multicolumn{1}{c|}{Clean} & Motion & Defocus & \begin{tabular}[c]{@{}c@{}}Frosted \\ Glass\end{tabular} & \multicolumn{1}{c|}{Gaussian} & Gaussian & Impulse & Shot & Speckle & \multicolumn{1}{c|}{Intensity} & Brightness & Contrast & Saturate & \multicolumn{1}{c|}{JPEG} & Snow & Spatter & Fog & \multicolumn{1}{c|}{Frost} & \begin{tabular}[c]{@{}c@{}}Geometric\\ Distortion\end{tabular} \\ \midrule
			\multicolumn{1}{c|}{\textbf{ResNet-50}} & \multicolumn{1}{c|}{69.6} & 38.7 & \textbf{43.5} & 31.1 & \multicolumn{1}{c|}{45.5} & 43.2 & 40.7 & 44.2 & 50.9 & \multicolumn{1}{c|}{59.8} & 63.5 & 50.3 & 63.8 & \multicolumn{1}{c|}{58.2} & 31.3 & 47.0 & 56.9 & \multicolumn{1}{c|}{39.8} & 67.2 \\
			\multicolumn{1}{c|}{w/o ASPP} & \multicolumn{1}{c|}{57.6} & 28.8 & 28.8 & 21.9 & \multicolumn{1}{c|}{31.5} & 31.7 & 29.2 & 32.3 & 38.4 & \multicolumn{1}{c|}{45.9} & 49.8 & 36.0 & 51.1 & \multicolumn{1}{c|}{45.0} & 23.1 & 37.6 & 42.0 & \multicolumn{1}{c|}{26.7} & 56.3 \\
			\multicolumn{1}{c|}{w/o AC} & \multicolumn{1}{c|}{68.9} & 39.3 & 41.6 & 29.0 & \multicolumn{1}{c|}{43.6} & 43.6 & 42.0 & 44.1 & 50.8 & \multicolumn{1}{c|}{59.3} & 62.7 & 48.9 & 62.9 & \multicolumn{1}{c|}{56.9} & 31.2 & 46.4 & 55.9 & \multicolumn{1}{c|}{38.3} & 65.9 \\
			\multicolumn{1}{c|}{w/ DPC} & \multicolumn{1}{c|}{68.0} & 38.6 & 40.6 & 29.7 & \multicolumn{1}{c|}{\textbf{42.5}} & 44.0 & 42.2 & 45.2 & 51.6 & \multicolumn{1}{c|}{59.8} & 61.5 & 48.9 & 62.6 & \multicolumn{1}{c|}{56.5} & 30.4 & 46.8 & 56.0 & \multicolumn{1}{c|}{37.9} & 64.9 \\
			\multicolumn{1}{c|}{w/o LRL} & \multicolumn{1}{c|}{69.0} & 40.0 & 41.2 & 30.1 & \multicolumn{1}{c|}{43.0} & 43.5 & 41.9 & 44.3 & 50.8 & \multicolumn{1}{c|}{59.7} & 62.4 & 48.5 & 62.5 & \multicolumn{1}{c|}{57.2} & 30.4 & 46.6 & 55.8 & \multicolumn{1}{c|}{39.1} & 65.5 \\
			\multicolumn{1}{c|}{w/ GAP} & \multicolumn{1}{c|}{\textbf{71.5}} & \textbf{41.6} & 42.9 & \textbf{33.0} & \multicolumn{1}{c|}{\textbf{45.9}} & \textbf{45.6} & \textbf{45.3} & \textbf{46.4} & \textbf{53.7} & \multicolumn{1}{c|}{\textbf{63.4}} & \textbf{65.7} & \textbf{52.1} & \textbf{66.1} & \multicolumn{1}{c|}{\textbf{59.5}} & \textbf{33.6} & \textbf{50.0} & \textbf{60.3} & \multicolumn{1}{c|}{\textbf{43.8}} & \textbf{67.6} \\ \midrule
			\multicolumn{1}{c|}{\textbf{ResNet-101}} & \multicolumn{1}{c|}{70.3} & 45.8 & 45.6 & 33.2 & \multicolumn{1}{c|}{46.6} & 49.4 & 48.3 & 50.1 & 55.4 & \multicolumn{1}{c|}{61.3} & 64.5 & 50.6 & 65.3 & \multicolumn{1}{c|}{59.7} & 31.4 & 50.4 & 57.6 & \multicolumn{1}{c|}{41.2} & 67.6 \\
			\multicolumn{1}{c|}{w/o ASPP} & \multicolumn{1}{c|}{60.5} & 36.4 & 34.2 & 25.1 & \multicolumn{1}{c|}{36.3} & 36.4 & 34.1 & 37.1 & 43.0 & \multicolumn{1}{c|}{50.5} & 53.6 & 39.2 & 54.1 & \multicolumn{1}{c|}{49.8} & 24.5 & 41.9 & 45.5 & \multicolumn{1}{c|}{29.6} & 59.8 \\
			\multicolumn{1}{c|}{w/o AC} & \multicolumn{1}{c|}{70.2} & \textbf{46.8} & 45.8 & 33.5 & \multicolumn{1}{c|}{46.3} & 46.0 & 45.2 & 46.6 & 52.9 & \multicolumn{1}{c|}{60.5} & 64.4 & 50.5 & 64.5 & \multicolumn{1}{c|}{59.6} & 32.3 & 51.0 & 57.9 & \multicolumn{1}{c|}{40.4} & 66.8 \\
			\multicolumn{1}{c|}{w/ DPC} & \multicolumn{1}{c|}{69.5} & 44.5 & 44.8 & 32.3 & \multicolumn{1}{c|}{46.2} & 48.4 & 45.0 & 49.4 & 54.8 & \multicolumn{1}{c|}{61.5} & 63.5 & 51.3 & 64.0 & \multicolumn{1}{c|}{59.4} & 32.3 & 49.9 & 58.3 & \multicolumn{1}{c|}{40.5} & 65.7 \\
			\multicolumn{1}{c|}{w/o LRL} & \multicolumn{1}{c|}{69.6} & 44.2 & 44.8 & 33.5 & \multicolumn{1}{c|}{45.8} & 47.4 & 45.2 & 48.5 & 53.8 & \multicolumn{1}{c|}{61.3} & 64.1 & 50.7 & 64.6 & \multicolumn{1}{c|}{58.4} & 32.2 & 50.6 & 57.5 & \multicolumn{1}{c|}{40.4} & 65.9 \\
			\multicolumn{1}{c|}{w/ GAP} & \multicolumn{1}{c|}{\textbf{72.5}} & 46.7 & \textbf{46.3} & \textbf{36.5} & \multicolumn{1}{c|}{\textbf{47.6}} & \textbf{50.5} & \textbf{48.5} & \textbf{51.3} & \textbf{56.6} & \multicolumn{1}{c|}{\textbf{64.3}} & \textbf{66.7} & \textbf{53.6} & \textbf{66.0} & \multicolumn{1}{c|}{\textbf{61.1}} & \textbf{36.4} & \textbf{52.6} & \textbf{61.7} & \multicolumn{1}{c|}{\textbf{44.7}} & \textbf{68.4} \\ \midrule
			\multicolumn{1}{c|}{\textbf{Xception-41}} & \multicolumn{1}{c|}{75.5} & 52.9 & 54.7 & 35.5 & \multicolumn{1}{c|}{53.9} & 55.8 & 53.3 & 56.7 & 62.8 & \multicolumn{1}{c|}{67.6} & 70.8 & 51.9 & 70.9 & \multicolumn{1}{c|}{64.6} & 42.5 & 59.0 & 63.1 & \multicolumn{1}{c|}{48.4} & \textbf{73.0} \\
			\multicolumn{1}{c|}{w/o ASPP} & \multicolumn{1}{c|}{66.9} & 45.9 & 45.3 & 30.4 & \multicolumn{1}{c|}{45.6} & 47.2 & 45.5 & 48.0 & 52.9 & \multicolumn{1}{c|}{58.5} & 61.0 & 43.1 & 61.5 & \multicolumn{1}{c|}{56.0} & 34.6 & 50.6 & 53.1 & \multicolumn{1}{c|}{39.3} & 65.4 \\
			\multicolumn{1}{c|}{w/o AC} & \multicolumn{1}{c|}{75.0} & 53.2 & 54.9 & 36.5 & \multicolumn{1}{c|}{54.9} & 54.1 & 52.6 & 55.5 & 61.4 & \multicolumn{1}{c|}{67.1} & 69.7 & 50.5 & 70.5 & \multicolumn{1}{c|}{64.5} & 40.9 & 60.1 & 62.3 & \multicolumn{1}{c|}{47.0} & 71.8 \\
			\multicolumn{1}{c|}{w/ DPC} & \multicolumn{1}{c|}{75.3} & 51.6 & 54.8 & \textbf{37.5} & \multicolumn{1}{c|}{54.3} & \textbf{56.8} & 55.1 & 58.1 & 63.0 & \multicolumn{1}{c|}{67.8} & 70.0 & 50.8 & 70.7 & \multicolumn{1}{c|}{65.5} & 40.6 & 58.1 & 61.9 & \multicolumn{1}{c|}{47.7} & 72.0 \\
			\multicolumn{1}{c|}{w/o LRL} & \multicolumn{1}{c|}{76.1} & 52.9 & \textbf{56.7} & 36.7 & \multicolumn{1}{c|}{\textbf{55.8}} & 56.7 & \textbf{56.6} & \textbf{58.3} & \textbf{63.9} & \multicolumn{1}{c|}{68.8} & 70.9 & \textbf{53.8} & \textbf{71.9} & \multicolumn{1}{c|}{65.1} & 41.4 & 59.1 & 63.3 & \multicolumn{1}{c|}{48.4} & 72.9 \\
			\multicolumn{1}{c|}{w/ GAP} & \multicolumn{1}{c|}{\textbf{76.5}} & \textbf{55.0} & 55.2 & 36.3 & \multicolumn{1}{c|}{55.0} & 55.3 & 55.7 & 56.6 & 62.7 & \multicolumn{1}{c|}{\textbf{68.8}} & \textbf{71.1} & 52.8 & 71.5 & \multicolumn{1}{c|}{\textbf{66.1}} & \textbf{43.3} & \textbf{61.4} & \textbf{63.7} & \multicolumn{1}{c|}{\textbf{48.9}} & 72.3 \\ \midrule
			\multicolumn{1}{c|}{\textbf{Xception-65}} & \multicolumn{1}{c|}{76.5} & 53.5 & 58.3 & 37.7 & \multicolumn{1}{c|}{57.2} & 56.6 & 54.7 & 57.4 & 62.5 & \multicolumn{1}{c|}{69.3} & 71.8 & 55.9 & 72.1 & \multicolumn{1}{c|}{66.7} & 40.2 & 58.5 & 64.0 & \multicolumn{1}{c|}{47.5} & \textbf{73.6} \\
			\multicolumn{1}{c|}{w/o ASPP} & \multicolumn{1}{c|}{70.6} & 47.5 & 47.8 & 29.1 & \multicolumn{1}{c|}{48.6} & 45.4 & 44.2 & 45.6 & 51.8 & \multicolumn{1}{c|}{62.4} & 64.7 & 48.1 & 64.6 & \multicolumn{1}{c|}{58.3} & 35.7 & 52.6 & 56.4 & \multicolumn{1}{c|}{39.4} & 68.7 \\
			\multicolumn{1}{c|}{w/o AC} & \multicolumn{1}{c|}{76.4} & \textbf{57.6} & 57.3 & 38.4 & \multicolumn{1}{c|}{56.9} & 56.5 & 54.5 & 57.0 & 62.2 & \multicolumn{1}{c|}{69.6} & 71.4 & 55.0 & 72.3 & \multicolumn{1}{c|}{66.3} & 42.5 & 60.4 & 63.6 & \multicolumn{1}{c|}{46.4} & 73.6 \\
			\multicolumn{1}{c|}{w/ DPC} & \multicolumn{1}{c|}{76.1} & 53.7 & 55.0 & 34.6 & \multicolumn{1}{c|}{55.0} & 54.8 & 54.0 & 56.0 & 61.6 & \multicolumn{1}{c|}{68.9} & 70.9 & 54.0 & 71.1 & \multicolumn{1}{c|}{66.0} & 40.9 & 58.3 & 61.9 & \multicolumn{1}{c|}{46.5} & 73.4 \\
			\multicolumn{1}{c|}{w/o LRL} & \multicolumn{1}{c|}{76.2} & 55.2 & 55.1 & 36.6 & \multicolumn{1}{c|}{55.6} & 56.5 & 55.4 & 56.8 & 61.8 & \multicolumn{1}{c|}{68.9} & 71.1 & 56.1 & 71.1 & \multicolumn{1}{c|}{64.3} & 40.3 & 58.4 & 64.2 & \multicolumn{1}{c|}{46.1} & 73.0 \\
			\multicolumn{1}{c|}{w/ GAP} & \multicolumn{1}{c|}{\textbf{77.5}} & 56.8 & \textbf{59.8} & \textbf{41.8} & \multicolumn{1}{c|}{\textbf{59.1}} & \textbf{57.9} & \textbf{57.6} & \textbf{57.6} & \textbf{62.6} & \multicolumn{1}{c|}{\textbf{71.0}} & \textbf{73.1} & \textbf{57.4} & \textbf{73.0} & \multicolumn{1}{c|}{\textbf{67.3}} & \textbf{42.8} & \textbf{61.1} & \textbf{65.3} & \multicolumn{1}{c|}{\textbf{49.7}} & 73.2 \\ \midrule
			\multicolumn{1}{c|}{\textbf{Xception-71}} & \multicolumn{1}{c|}{76.7} & 56.5 & 59.1 & 40.2 & \multicolumn{1}{c|}{59.5} & 56.6 & 57.8 & 57.6 & 63.2 & \multicolumn{1}{c|}{69.9} & 72.1 & 57.1 & 72.6 & \multicolumn{1}{c|}{68.1} & 43.9 & 60.9 & 66.1 & \multicolumn{1}{c|}{50.9} & \textbf{73.6} \\
			\multicolumn{1}{c|}{w/o ASPP} & \multicolumn{1}{c|}{70.5} & 48.3 & 49.2 & 33.3 & \multicolumn{1}{c|}{50.1} & 47.5 & 47.1 & 48.2 & 54.6 & \multicolumn{1}{c|}{62.7} & 65.1 & 48.8 & 65.6 & \multicolumn{1}{c|}{60.2} & 37.0 & 53.4 & 57.3 & \multicolumn{1}{c|}{44.1} & 69.3 \\
			\multicolumn{1}{c|}{w/o AC} & \multicolumn{1}{c|}{75.7} & 55.9 & 58.8 & \textbf{41.8} & \multicolumn{1}{c|}{59.0} & 57.1 & 58.2 & 57.3 & 62.6 & \multicolumn{1}{c|}{69.5} & 71.0 & 56.9 & 71.4 & \multicolumn{1}{c|}{67.6} & 41.9 & 60.9 & 64.1 & \multicolumn{1}{c|}{48.2} & 73.0 \\
			\multicolumn{1}{c|}{w/ DPC} & \multicolumn{1}{c|}{76.8} & 53.5 & 54.6 & 35.8 & \multicolumn{1}{c|}{55.4} & 55.5 & 55.6 & 54.7 & 60.5 & \multicolumn{1}{c|}{69.0} & 71.4 & 54.0 & 71.0 & \multicolumn{1}{c|}{66.3} & 42.5 & 58.3 & 63.3 & \multicolumn{1}{c|}{49.7} & 73.3 \\
			\multicolumn{1}{c|}{w/o LRL} & \multicolumn{1}{c|}{76.3} & 56.4 & 56.4 & 40.5 & \multicolumn{1}{c|}{55.9} & \textbf{59.9} & \textbf{59.3} & \textbf{60.2} & \textbf{64.6} & \multicolumn{1}{c|}{71.1} & 72.1 & 53.0 & 72.3 & \multicolumn{1}{c|}{67.7} & 43.8 & 59.3 & 64.1 & \multicolumn{1}{c|}{50.8} & 73.2 \\
			\multicolumn{1}{c|}{w/ GAP} & \multicolumn{1}{c|}{\textbf{77.7}} & \textbf{57.8} & \textbf{58.7} & 38.3 & \multicolumn{1}{c|}{\textbf{59.1}} & 58.8 & 55.2 & 58.4 & 63.7 & \multicolumn{1}{c|}{\textbf{71.9}} & \textbf{73.8} & \textbf{60.4} & \textbf{73.9} & \multicolumn{1}{c|}{\textbf{69.2}} & \textbf{46.9} & \textbf{61.6} & \textbf{67.9} & \multicolumn{1}{c|}{\textbf{53.5}} & 73.5 \\ \bottomrule
		\end{tabular}
	\end{adjustbox}
	\caption{
		Mean IoU for clean and corrupted variants of the validation set of PASCAL VOC 2012 for several network backbones of the DeepLabv3+ architecture and respective architectural ablations. 
		Every mIoU is averaged over all available severity levels, except for corruptions of category noise where only the first three severity levels are considered. 
		The standard deviation for image corruptions of category noise is $0.3$ or less.
		Highest mIoU per corruption is bold.}
	\label{tab:miou_voc_allbackbones_and_ablations}
\end{table*}

\begin{table*}[h]
	\begin{adjustbox}{width=\textwidth}
		\begin{tabular}{@{}ccccccccccccccccccc@{}}
			\toprule
			& \multicolumn{4}{c}{\textbf{Blur}} & \multicolumn{5}{c}{\textbf{Noise}} & \multicolumn{4}{c}{\textbf{Digital}} & \multicolumn{4}{c}{\textbf{Weather}} & \textbf{} \\ \midrule
			\multicolumn{1}{c|}{\begin{tabular}[c]{@{}c@{}}Deeplab-v3+ \\ Backbone\end{tabular}} & Motion & Defocus & \begin{tabular}[c]{@{}c@{}}Frosted \\ Glass\end{tabular} & \multicolumn{1}{c|}{Gaussian} & Gaussian & Impulse & Shot & Speckle & \multicolumn{1}{c|}{Intensity} & Brightness & Contrast & Saturate & \multicolumn{1}{c|}{JPEG} & Snow & Spatter & Fog & \multicolumn{1}{c|}{Frost} & \begin{tabular}[c]{@{}c@{}}Geometric\\ Distortion\end{tabular} \\ \midrule
			\multicolumn{1}{c|}{\textbf{ResNet-50}} & 100.0 & 100.0 & 100.0 & \multicolumn{1}{c|}{100.0} & 100.0 & 100.0 & 100.0 & 100.0 & \multicolumn{1}{c|}{100.0} & 100.0 & 100.0 & 100.0 & \multicolumn{1}{c|}{100.0} & 100.0 & 100.0 & 100.0 & \multicolumn{1}{c|}{100.0} & 100.0 \\
			\multicolumn{1}{c|}{w/o ASPP} & \textbf{116.1} & \textbf{126.1} & \textbf{113.3} & \multicolumn{1}{c|}{\textbf{125.7}} & \textbf{120.3} & \textbf{119.4} & \textbf{121.4} & \textbf{125.6} & \multicolumn{1}{c|}{\textbf{134.7}} & \textbf{137.7} & \textbf{128.7} & \textbf{135.0} & \multicolumn{1}{c|}{\textbf{131.4}} & \textbf{111.9} & \textbf{117.8} & \textbf{134.5} & \multicolumn{1}{c|}{\textbf{121.7}} & \textbf{133.3} \\
			\multicolumn{1}{c|}{w/o AC} & 99.1 & 103.4 & 103.0 & \multicolumn{1}{c|}{103.5} & 99.3 & 97.9 & 100.2 & 100.4 & \multicolumn{1}{c|}{101.3} & 102.3 & 102.6 & 102.4 & \multicolumn{1}{c|}{103.1} & 100.0 & 101.1 & 102.2 & \multicolumn{1}{c|}{102.6} & 104.1 \\
			\multicolumn{1}{c|}{w/ DPC} & 100.2 & 105.2 & 101.9 & \multicolumn{1}{c|}{105.5} & 98.6 & 97.6 & 98.2 & 98.6 & \multicolumn{1}{c|}{100.1} & 105.5 & 102.6 & 103.2 & \multicolumn{1}{c|}{104.0} & 101.3 & 100.5 & 101.9 & \multicolumn{1}{c|}{103.3} & 107.0 \\
			\multicolumn{1}{c|}{w/o LRL} & 98.0 & 104.1 & 101.5 & \multicolumn{1}{c|}{104.6} & 99.4 & 98.0 & 99.9 & 100.3 & \multicolumn{1}{c|}{100.3} & 103.0 & 103.6 & 103.4 & \multicolumn{1}{c|}{102.4} & 101.2 & 100.9 & 102.4 & \multicolumn{1}{c|}{101.2} & 105.2 \\
			\multicolumn{1}{c|}{w/ GAP} & 95.4 & 101.1 & 97.2 & \multicolumn{1}{c|}{99.4} & 95.9 & 92.3 & 96.2 & 94.3 & \multicolumn{1}{c|}{91.2} & 94.0 & 96.3 & 93.7 & \multicolumn{1}{c|}{96.7} & 96.6 & 94.4 & 92.1 & \multicolumn{1}{c|}{93.5} & 98.9 \\ \midrule
			\multicolumn{1}{c|}{\textbf{ResNet-101}} & 100.0 & 100.0 & 100.0 & \multicolumn{1}{c|}{100.0} & 100.0 & 100.0 & 100.0 & 100.0 & \multicolumn{1}{c|}{100.0} & 100.0 & 100.0 & 100.0 & \multicolumn{1}{c|}{100.0} & 100.0 & 100.0 & 100.0 & \multicolumn{1}{c|}{100.0} & 100.0 \\
			\multicolumn{1}{c|}{w/o ASPP} & \textbf{117.2} & \textbf{120.9} & \textbf{112.2} & \multicolumn{1}{c|}{\textbf{119.4}} & \textbf{125.7} & \textbf{127.4} & \textbf{126.1} & \textbf{127.8} & \multicolumn{1}{c|}{\textbf{127.9}} & \textbf{130.9} & \textbf{123.0} & \textbf{132.3} & \multicolumn{1}{c|}{\textbf{124.5}} & \textbf{110.1} & \textbf{117.1} & \textbf{128.5} & \multicolumn{1}{c|}{\textbf{119.8}} & \textbf{124.0} \\
			\multicolumn{1}{c|}{w/o AC} & 98.1 & 99.6 & 99.5 & \multicolumn{1}{c|}{100.7} & 106.7 & 106.0 & 107.0 & 105.7 & \multicolumn{1}{c|}{102.1} & 100.5 & 100.1 & 102.3 & \multicolumn{1}{c|}{100.4} & 98.7 & 98.8 & 99.3 & \multicolumn{1}{c|}{101.4} & 102.3 \\
			\multicolumn{1}{c|}{w/ DPC} & 102.3 & 101.4 & 101.4 & \multicolumn{1}{c|}{100.8} & 102.0 & 106.2 & 101.5 & 101.4 & \multicolumn{1}{c|}{99.5} & 103.0 & 98.6 & 103.6 & \multicolumn{1}{c|}{100.9} & 98.7 & 101.2 & 98.3 & \multicolumn{1}{c|}{101.3} & 105.9 \\
			\multicolumn{1}{c|}{w/o LRL} & 102.9 & 101.4 & 99.6 & \multicolumn{1}{c|}{101.5} & 103.9 & 105.9 & 103.3 & 103.5 & \multicolumn{1}{c|}{100.0} & 101.2 & 99.7 & 102.0 & \multicolumn{1}{c|}{103.3} & 99.0 & 99.6 & 100.2 & \multicolumn{1}{c|}{101.5} & 105.2 \\
			\multicolumn{1}{c|}{w/ GAP} & 98.3 & 98.6 & 95.0 & \multicolumn{1}{c|}{98.2} & 97.8 & 99.5 & 97.7 & 97.2 & \multicolumn{1}{c|}{92.2} & 93.8 & 93.9 & 97.8 & \multicolumn{1}{c|}{96.6} & 92.8 & 95.6 & 90.4 & \multicolumn{1}{c|}{94.0} & 97.6 \\ \midrule
			\multicolumn{1}{c|}{\textbf{Xception-41}} & 100.0 & 100.0 & 100.0 & \multicolumn{1}{c|}{100.0} & 100.0 & 100.0 & 100.0 & 100.0 & \multicolumn{1}{c|}{100.0} & 100.0 & 100.0 & 100.0 & \multicolumn{1}{c|}{100.0} & 100.0 & 100.0 & 100.0 & \multicolumn{1}{c|}{100.0} & 100.0 \\
			\multicolumn{1}{c|}{w/o ASPP} & \textbf{114.8} & \textbf{120.8} & \textbf{107.9} & \multicolumn{1}{c|}{\textbf{117.9}} & \textbf{119.3} & \textbf{116.6} & \textbf{120.1} & \textbf{126.5} & \multicolumn{1}{c|}{\textbf{127.9}} & \textbf{133.7} & \textbf{118.3} & \textbf{132.3} & \multicolumn{1}{c|}{\textbf{124.4}} & \textbf{113.6} & \textbf{120.6} & \textbf{127.2} & \multicolumn{1}{c|}{\textbf{117.6}} & \textbf{128.0} \\
			\multicolumn{1}{c|}{w/o AC} & 99.4 & 99.7 & 98.5 & \multicolumn{1}{c|}{97.8} & 103.9 & 101.5 & 103.0 & 103.6 & \multicolumn{1}{c|}{101.6} & 103.7 & 103.0 & 101.4 & \multicolumn{1}{c|}{100.2} & 102.7 & 97.3 & 102.2 & \multicolumn{1}{c|}{102.7} & 104.3 \\
			\multicolumn{1}{c|}{w/ DPC} & 102.9 & 99.8 & 97.0 & \multicolumn{1}{c|}{99.1} & 97.8 & 96.1 & 96.8 & 99.4 & \multicolumn{1}{c|}{99.4} & 102.6 & 102.3 & 100.5 & \multicolumn{1}{c|}{97.4} & 103.2 & 102.3 & 103.2 & \multicolumn{1}{c|}{101.5} & 103.7 \\
			\multicolumn{1}{c|}{w/o LRL} & 100.0 & 95.6 & 98.2 & \multicolumn{1}{c|}{95.8} & 97.9 & 92.9 & 96.5 & 97.0 & \multicolumn{1}{c|}{96.1} & 99.6 & 96.0 & 96.6 & \multicolumn{1}{c|}{98.6} & 101.8 & 99.8 & 99.5 & \multicolumn{1}{c|}{100.0} & 100.1 \\
			\multicolumn{1}{c|}{w/ GAP} & 95.5 & 99.0 & 98.8 & \multicolumn{1}{c|}{97.5} & 101.2 & 94.9 & 100.3 & 100.3 & \multicolumn{1}{c|}{96.4} & 99.0 & 98.1 & 97.8 & \multicolumn{1}{c|}{95.7} & 98.6 & 94.3 & 98.4 & \multicolumn{1}{c|}{99.0} & 102.4 \\ \midrule
			\multicolumn{1}{c|}{\textbf{Xception-65}} & 100.0 & 100.0 & 100.0 & \multicolumn{1}{c|}{100.0} & 100.0 & 100.0 & 100.0 & 100.0 & \multicolumn{1}{c|}{100.0} & 100.0 & 100.0 & 100.0 & \multicolumn{1}{c|}{100.0} & 100.0 & 100.0 & 100.0 & \multicolumn{1}{c|}{100.0} & 100.0 \\
			\multicolumn{1}{c|}{w/o ASPP} & \textbf{112.8} & \textbf{125.2} & \textbf{113.9} & \multicolumn{1}{c|}{\textbf{120.2}} & \textbf{125.9} & \textbf{123.1} & \textbf{127.5} & \textbf{128.7} & \multicolumn{1}{c|}{\textbf{122.3}} & \textbf{125.3} & \textbf{117.7} & \textbf{126.7} & \multicolumn{1}{c|}{\textbf{125.3}} & \textbf{107.4} & \textbf{114.0} & \textbf{121.1} & \multicolumn{1}{c|}{\textbf{115.4}} & \textbf{118.3} \\
			\multicolumn{1}{c|}{w/o AC} & 91.1 & 102.5 & 98.9 & \multicolumn{1}{c|}{100.7} & 100.3 & 100.5 & 100.9 & 101.0 & \multicolumn{1}{c|}{98.8} & 101.3 & 102.2 & 99.1 & \multicolumn{1}{c|}{101.2} & 96.2 & 95.4 & 101.2 & \multicolumn{1}{c|}{102.1} & 99.8 \\
			\multicolumn{1}{c|}{w/ DPC} & 99.5 & 108.0 & 104.9 & \multicolumn{1}{c|}{105.3} & 104.1 & 101.5 & 103.2 & 102.6 & \multicolumn{1}{c|}{101.2} & 103.4 & 104.4 & 103.5 & \multicolumn{1}{c|}{102.3} & 98.8 & 100.3 & 105.8 & \multicolumn{1}{c|}{101.9} & 100.5 \\
			\multicolumn{1}{c|}{w/o LRL} & 96.3 & 107.7 & 101.8 & \multicolumn{1}{c|}{103.8} & 100.3 & 98.4 & 101.2 & 101.9 & \multicolumn{1}{c|}{101.2} & 102.6 & 99.7 & 103.4 & \multicolumn{1}{c|}{107.1} & 99.7 & 100.3 & 99.4 & \multicolumn{1}{c|}{102.7} & 102.2 \\
			\multicolumn{1}{c|}{w/ GAP} & 92.8 & 96.4 & 93.4 & \multicolumn{1}{c|}{95.6} & 97.0 & 93.6 & 99.4 & 99.8 & \multicolumn{1}{c|}{94.3} & 95.3 & 96.6 & 96.7 & \multicolumn{1}{c|}{98.2} & 95.6 & 93.6 & 96.5 & \multicolumn{1}{c|}{95.7} & 101.4 \\ \midrule
			\multicolumn{1}{c|}{\textbf{Xception-71}} & 100.0 & 100.0 & 100.0 & \multicolumn{1}{c|}{100.0} & 100.0 & 100.0 & 100.0 & 100.0 & \multicolumn{1}{c|}{100.0} & 100.0 & 100.0 & 100.0 & \multicolumn{1}{c|}{100.0} & 100.0 & 100.0 & 100.0 & \multicolumn{1}{c|}{100.0} & 100.0 \\
			\multicolumn{1}{c|}{w/o ASPP} & \textbf{118.9} & \textbf{124.3} & \textbf{111.5} & \multicolumn{1}{c|}{\textbf{123.0}} & \textbf{121.0} & \textbf{125.2} & \textbf{122.0} & \textbf{123.4} & \multicolumn{1}{c|}{\textbf{123.8}} & \textbf{125.0} & \textbf{119.2} & \textbf{125.6} & \multicolumn{1}{c|}{\textbf{124.9}} & \textbf{112.3} & \textbf{119.1} & \textbf{126.0} & \multicolumn{1}{c|}{\textbf{113.9}} & \textbf{116.6} \\
			\multicolumn{1}{c|}{w/o AC} & 101.3 & 100.9 & 97.3 & \multicolumn{1}{c|}{101.2} & 98.8 & 99.0 & 100.7 & 101.8 & \multicolumn{1}{c|}{101.4} & 103.9 & 100.4 & 104.4 & \multicolumn{1}{c|}{101.5} & 103.5 & 100.1 & 105.9 & \multicolumn{1}{c|}{105.4} & 102.3 \\
			\multicolumn{1}{c|}{w/ DPC} & 107.0 & 111.1 & 107.4 & \multicolumn{1}{c|}{110.1} & 102.5 & 105.1 & 106.7 & 107.5 & \multicolumn{1}{c|}{102.9} & 102.4 & 107.1 & 105.8 & \multicolumn{1}{c|}{105.5} & 102.5 & 106.7 & 108.1 & \multicolumn{1}{c|}{102.5} & 101.1 \\
			\multicolumn{1}{c|}{w/o LRL} & 100.2 & 106.7 & 99.5 & \multicolumn{1}{c|}{108.7} & 92.4 & 96.5 & 93.9 & 96.3 & \multicolumn{1}{c|}{95.9} & 99.9 & 109.4 & 101.2 & \multicolumn{1}{c|}{101.2} & 100.2 & 104.0 & 105.8 & \multicolumn{1}{c|}{100.2} & 101.5 \\
			\multicolumn{1}{c|}{w/ GAP} & 97.1 & 101.1 & 103.2 & \multicolumn{1}{c|}{100.8} & 94.8 & 106.1 & 98.2 & 98.7 & \multicolumn{1}{c|}{93.2} & 93.9 & 92.3 & 95.2 & \multicolumn{1}{c|}{96.5} & 94.7 & 98.3 & 94.7 & \multicolumn{1}{c|}{94.7} & 100.6 \\ \bottomrule
		\end{tabular}
	\end{adjustbox}
	\caption{
		Mean CD for corrupted variants of the validation set of PASCAL VOC 2012 for several network backbones of the DeepLabv3+ architecture and respective architectural ablations. 
		Highest CD per corruption is bold.
		\vspace{+1.6cm}}
	\label{tab:cd_voc_allbackbones_and_ablations}
\end{table*}

\begin{table*}[h]
	\begin{adjustbox}{width=\textwidth}
		\begin{tabular}{@{}ccccccccccccccccccc@{}}
			\toprule
			& \multicolumn{4}{c}{\textbf{Blur}} & \multicolumn{5}{c}{\textbf{Noise}} & \multicolumn{4}{c}{\textbf{Digital}} & \multicolumn{4}{c}{\textbf{Weather}} & \textbf{} \\ \midrule
			\multicolumn{1}{c|}{\begin{tabular}[c]{@{}c@{}}Deeplab-v3+ \\ Backbone\end{tabular}} & Motion & Defocus & \begin{tabular}[c]{@{}c@{}}Frosted \\ Glass\end{tabular} & \multicolumn{1}{c|}{Gaussian} & Gaussian & Impulse & Shot & Speckle & \multicolumn{1}{c|}{Intensity} & Brightness & Contrast & Saturate & \multicolumn{1}{c|}{JPEG} & Snow & Spatter & Fog & \multicolumn{1}{c|}{Frost} & \begin{tabular}[c]{@{}c@{}}Geometric\\ Distortion\end{tabular} \\ \midrule
			\multicolumn{1}{c|}{\textbf{ResNet-50}} & \textbf{100.0} & 100.0 & 100.0 & \multicolumn{1}{c|}{100.0} & \textbf{100.0} & \textbf{100.0} & \textbf{100.0} & \textbf{100.0} & \multicolumn{1}{c|}{100.0} & 100.0 & 100.0 & 100.0 & \multicolumn{1}{c|}{100.0} & 100.0 & \textbf{100.0} & 100.0 & \multicolumn{1}{c|}{100.0} & 100.0 \\
			\multicolumn{1}{c|}{w/o ASPP} & 93.1 & 110.4 & 92.6 & \multicolumn{1}{c|}{108.2} & 98.3 & 98.3 & 99.8 & 103.0 & \multicolumn{1}{c|}{\textbf{119.8}} & \textbf{128.6} & \textbf{111.6} & \textbf{111.1} & \multicolumn{1}{c|}{\textbf{109.9}} & 90.0 & 88.7 & \textbf{122.6} & \multicolumn{1}{c|}{\textbf{103.6}} & 54.4 \\
			\multicolumn{1}{c|}{w/o AC} & 95.9 & 104.7 & \textbf{103.6} & \multicolumn{1}{c|}{105.0} & 95.8 & 93.2 & 97.7 & 97.2 & \multicolumn{1}{c|}{98.2} & 102.1 & 103.2 & 102.8 & \multicolumn{1}{c|}{105.2} & 98.2 & 99.5 & 102.0 & \multicolumn{1}{c|}{102.8} & 126.4 \\
			\multicolumn{1}{c|}{w/ DPC} & 95.1 & 105.0 & 99.2 & \multicolumn{1}{c|}{105.7} & 90.7 & 89.4 & 89.7 & 87.7 & \multicolumn{1}{c|}{83.8} & 106.3 & 98.4 & 92.2 & \multicolumn{1}{c|}{100.5} & 98.1 & 94.0 & 93.6 & \multicolumn{1}{c|}{101.2} & 127.5 \\
			\multicolumn{1}{c|}{w/o LRL} & 93.9 & 106.4 & 100.9 & \multicolumn{1}{c|}{107.7} & 96.3 & 93.7 & 97.1 & 97.4 & \multicolumn{1}{c|}{94.7} & 107.4 & 106.0 & 110.1 & \multicolumn{1}{c|}{103.1} & \textbf{100.5} & 99.1 & 103.0 & \multicolumn{1}{c|}{100.2} & 144.3 \\
			\multicolumn{1}{c|}{w/ GAP} & 96.9 & \textbf{109.5} & 99.8 & \multicolumn{1}{c|}{\textbf{106.4}} & 98.2 & 90.8 & 99.1 & 95.2 & \multicolumn{1}{c|}{83.2} & 95.3 & 100.1 & 93.5 & \multicolumn{1}{c|}{104.5} & 98.7 & 95.1 & 88.1 & \multicolumn{1}{c|}{93.2} & \textbf{163.7} \\ \midrule
			\multicolumn{1}{c|}{\textbf{ResNet-101}} & 100.0 & 100.0 & 100.0 & \multicolumn{1}{c|}{100.0} & 100.0 & 100.0 & 100.0 & 100.0 & \multicolumn{1}{c|}{100.0} & 100.0 & 100.0 & 100.0 & \multicolumn{1}{c|}{100.0} & \textbf{100.0} & \textbf{100.0} & 100.0 & \multicolumn{1}{c|}{100.0} & 100.0 \\
			\multicolumn{1}{c|}{w/o ASPP} & 98.0 & \textbf{106.3} & 95.4 & \multicolumn{1}{c|}{102.3} & 115.3 & \textbf{119.7} & 115.8 & \textbf{117.3} & \multicolumn{1}{c|}{111.0} & \textbf{119.4} & \textbf{107.8} & 127.7 & \multicolumn{1}{c|}{100.5} & 92.6 & 93.3 & \textbf{117.7} & \multicolumn{1}{c|}{\textbf{106.2}} & 23.7 \\
			\multicolumn{1}{c|}{w/o AC} & 95.4 & 98.5 & 98.7 & \multicolumn{1}{c|}{101.1} & \textbf{115.7} & 113.4 & \textbf{116.6} & 116.1 & \multicolumn{1}{c|}{\textbf{107.8}} & 100.9 & 99.5 & 113.7 & \multicolumn{1}{c|}{100.1} & 97.4 & 96.4 & 96.8 & \multicolumn{1}{c|}{102.4} & 123.4 \\
			\multicolumn{1}{c|}{w/ DPC} & 102.0 & 99.9 & \textbf{100.4} & \multicolumn{1}{c|}{98.5} & 101.0 & 111.1 & 99.9 & 99.1 & \multicolumn{1}{c|}{89.4} & 105.3 & 92.7 & 109.4 & \multicolumn{1}{c|}{96.1} & 95.8 & 99.1 & 88.1 & \multicolumn{1}{c|}{99.9} & 142.9 \\
			\multicolumn{1}{c|}{w/o LRL} & 103.7 & 100.3 & 97.4 & \multicolumn{1}{c|}{100.6} & 106.3 & 110.9 & 105.0 & 106.1 & \multicolumn{1}{c|}{92.7} & 96.1 & 95.9 & 100.7 & \multicolumn{1}{c|}{106.3} & 96.5 & 95.6 & 95.3 & \multicolumn{1}{c|}{100.8} & 137.9 \\
			\multicolumn{1}{c|}{w/ GAP} & \textbf{105.1} & 105.7 & 96.8 & \multicolumn{1}{c|}{\textbf{105.1}} & 105.0 & 108.7 & 105.2 & 106.3 & \multicolumn{1}{c|}{90.6} & 99.6 & 95.7 & \textbf{127.9} & \multicolumn{1}{c|}{\textbf{107.4}} & 92.9 & 99.9 & 85.0 & \multicolumn{1}{c|}{95.4} & \textbf{152.1} \\ \midrule
			\multicolumn{1}{c|}{\textbf{Xception-41}} & 100.0 & 100.0 & 100.0 & \multicolumn{1}{c|}{\textbf{100.0}} & 100.0 & 100.0 & 100.0 & 100.0 & \multicolumn{1}{c|}{100.0} & 100.0 & 100.0 & 100.0 & \multicolumn{1}{c|}{100.0} & 100.0 & 100.0 & 100.0 & \multicolumn{1}{c|}{100.0} & 100.0 \\
			\multicolumn{1}{c|}{w/o ASPP} & 93.0 & \textbf{104.0} & 91.3 & \multicolumn{1}{c|}{98.6} & 99.9 & 96.4 & 100.7 & \textbf{110.4} & \multicolumn{1}{c|}{\textbf{106.0}} & \textbf{127.2} & 101.1 & \textbf{118.5} & \multicolumn{1}{c|}{100.7} & 97.8 & 99.3 & \textbf{111.8} & \multicolumn{1}{c|}{102.0} & 60.3 \\
			\multicolumn{1}{c|}{w/o AC} & 96.9 & 97.3 & 96.6 & \multicolumn{1}{c|}{93.4} & 106.5 & \textbf{101.1} & 104.5 & 107.1 & \multicolumn{1}{c|}{101.1} & 113.9 & \textbf{104.2} & 99.4 & \multicolumn{1}{c|}{96.7} & 103.4 & 90.6 & 103.1 & \multicolumn{1}{c|}{\textbf{103.6}} & 129.5 \\
			\multicolumn{1}{c|}{w/ DPC} & \textbf{105.3} & 98.9 & 94.8 & \multicolumn{1}{c|}{97.2} & 94.2 & 91.0 & 91.7 & 97.0 & \multicolumn{1}{c|}{95.5} & 112.9 & 103.9 & 99.9 & \multicolumn{1}{c|}{89.9} & \textbf{105.2} & \textbf{104.7} & 108.4 & \multicolumn{1}{c|}{102.2} & 133.7 \\
			\multicolumn{1}{c|}{w/o LRL} & 102.9 & 93.5 & 98.8 & \multicolumn{1}{c|}{94.0} & 98.5 & 87.9 & 95.4 & 96.2 & \multicolumn{1}{c|}{92.4} & 111.5 & 94.6 & 92.8 & \multicolumn{1}{c|}{\textbf{101.3}} & 105.1 & 103.5 & 103.7 & \multicolumn{1}{c|}{102.5} & 126.9 \\
			\multicolumn{1}{c|}{w/ GAP} & 95.0 & 102.8 & \textbf{100.6} & \multicolumn{1}{c|}{99.3} & \textbf{107.8} & 93.7 & \textbf{106.0} & 108.7 & \multicolumn{1}{c|}{97.9} & 115.4 & 100.4 & 107.8 & \multicolumn{1}{c|}{95.3} & 100.6 & 92.0 & 103.5 & \multicolumn{1}{c|}{101.9} & \textbf{166.2} \\ \midrule
			\multicolumn{1}{c|}{\textbf{Xception-65}} & 100.0 & 100.0 & 100.0 & \multicolumn{1}{c|}{100.0} & 100.0 & 100.0 & 100.0 & 100.0 & \multicolumn{1}{c|}{100.0} & 100.0 & 100.0 & 100.0 & \multicolumn{1}{c|}{100.0} & \textbf{100.0} & \textbf{100.0} & 100.0 & \multicolumn{1}{c|}{100.0} & 100.0 \\
			\multicolumn{1}{c|}{w/o ASPP} & \textbf{100.1} & \textbf{124.9} & 106.9 & \multicolumn{1}{c|}{\textbf{113.8}} & \textbf{126.4} & \textbf{120.7} & \textbf{130.2} & \textbf{134.3} & \multicolumn{1}{c|}{\textbf{112.4}} & \textbf{124.8} & \textbf{109.0} & \textbf{133.9} & \multicolumn{1}{c|}{\textbf{125.0}} & 95.8 & 99.3 & \textbf{113.0} & \multicolumn{1}{c|}{\textbf{107.3}} & 61.7 \\
			\multicolumn{1}{c|}{w/o AC} & 81.7 & 105.2 & 98.1 & \multicolumn{1}{c|}{101.3} & 100.4 & 100.6 & 101.7 & 102.0 & \multicolumn{1}{c|}{93.8} & 106.5 & 104.3 & 92.5 & \multicolumn{1}{c|}{103.3} & 93.5 & 89.0 & 103.0 & \multicolumn{1}{c|}{103.5} & 96.0 \\
			\multicolumn{1}{c|}{w/ DPC} & 96.9 & 115.7 & \textbf{106.7} & \multicolumn{1}{c|}{109.3} & 106.5 & 101.1 & 104.7 & 103.5 & \multicolumn{1}{c|}{98.4} & 110.5 & 107.2 & 111.3 & \multicolumn{1}{c|}{102.9} & 96.7 & 98.1 & 112.9 & \multicolumn{1}{c|}{101.8} & 88.3 \\
			\multicolumn{1}{c|}{w/o LRL} & 91.2 & 116.1 & 102.2 & \multicolumn{1}{c|}{106.9} & 99.3 & 95.5 & 101.3 & 103.1 & \multicolumn{1}{c|}{101.1} & 109.5 & 98.0 & 115.1 & \multicolumn{1}{c|}{121.4} & 98.7 & 99.1 & 96.0 & \multicolumn{1}{c|}{103.8} & 109.9 \\
			\multicolumn{1}{c|}{w/ GAP} & 89.7 & 97.1 & 91.9 & \multicolumn{1}{c|}{95.4} & 98.5 & 91.1 & 103.8 & 106.6 & \multicolumn{1}{c|}{89.3} & 92.7 & 97.5 & 101.6 & \multicolumn{1}{c|}{103.9} & 95.5 & 90.6 & 97.8 & \multicolumn{1}{c|}{95.6} & \textbf{145.7} \\ \midrule
			\multicolumn{1}{c|}{\textbf{Xception-71}} & 100.0 & 100.0 & 100.0 & \multicolumn{1}{c|}{100.0} & 100.0 & 100.0 & 100.0 & 100.0 & \multicolumn{1}{c|}{100.0} & 100.0 & 100.0 & 100.0 & \multicolumn{1}{c|}{100.0} & 100.0 & 100.0 & 100.0 & \multicolumn{1}{c|}{100.0} & 100.0 \\
			\multicolumn{1}{c|}{w/o ASPP} & 110.1 & 121.4 & 101.9 & \multicolumn{1}{c|}{118.3} & \textbf{114.5} & \textbf{123.5} & \textbf{116.5} & 118.0 & \multicolumn{1}{c|}{\textbf{114.6}} & \textbf{117.1} & 110.6 & 120.5 & \multicolumn{1}{c|}{120.6} & 102.2 & 108.2 & 124.9 & \multicolumn{1}{c|}{102.6} & 41.2 \\
			\multicolumn{1}{c|}{w/o AC} & 98.1 & 96.6 & 92.9 & \multicolumn{1}{c|}{97.3} & 92.7 & 92.7 & 96.6 & 97.7 & \multicolumn{1}{c|}{91.9} & 102.8 & 96.0 & 106.3 & \multicolumn{1}{c|}{94.3} & 103.1 & 94.3 & 109.8 & \multicolumn{1}{c|}{\textbf{106.6}} & 88.3 \\
			\multicolumn{1}{c|}{w/ DPC} & \textbf{115.3} & \textbf{126.1} & \textbf{112.2} & \multicolumn{1}{c|}{\textbf{124.2}} & 105.6 & 111.6 & 115.1 & \textbf{120.9} & \multicolumn{1}{c|}{113.5} & 115.8 & 115.9 & \textbf{140.2} & \multicolumn{1}{c|}{\textbf{121.2}} & \textbf{104.4} & \textbf{116.9} & \textbf{126.6} & \multicolumn{1}{c|}{104.9} & \textbf{111.6} \\
			\multicolumn{1}{c|}{w/o LRL} & 98.4 & 113.3 & 98.1 & \multicolumn{1}{c|}{118.1} & 81.6 & 90.0 & 84.3 & 86.8 & \multicolumn{1}{c|}{75.8} & 90.7 & \textbf{118.6} & 97.8 & \multicolumn{1}{c|}{99.8} & 99.0 & 107.4 & 114.8 & \multicolumn{1}{c|}{98.7} & 99.4 \\
			\multicolumn{1}{c|}{w/ GAP} & 98.5 & 108.2 & 107.9 & \multicolumn{1}{c|}{107.5} & 93.7 & 118.8 & 101.0 & 103.7 & \multicolumn{1}{c|}{84.2} & 84.3 & 88.1 & 91.6 & \multicolumn{1}{c|}{98.3} & 93.9 & 102.0 & 92.3 & \multicolumn{1}{c|}{93.7} & 137.2 \\ \bottomrule
		\end{tabular}
	\end{adjustbox}
	\caption{
		Relative CD for corrupted variants of the validation set of PASCAL VOC 2012 for several network backbones of the DeepLabv3+ architecture and respective architectural ablations. 
		Highest rCD per corruption is bold.}
	\label{tab:rcd_voc_allbackbones_and_ablations}
\end{table*}

\begin{table*}[h]
	\begin{adjustbox}{width=\textwidth}
		
		\begin{tabular}{@{}cccccccccccccccccccc@{}}
			\toprule
			\multicolumn{2}{c}{} & \multicolumn{4}{c}{\textbf{Blur}} & \multicolumn{5}{c}{\textbf{Noise}} & \multicolumn{4}{c}{\textbf{Digital}} & \multicolumn{4}{c}{\textbf{Weather}} & \textbf{} \\ \midrule
			\multicolumn{1}{c|}{\begin{tabular}[c]{@{}c@{}}Deeplab-v3+ \\ Backbone\end{tabular}} & \multicolumn{1}{c|}{Clean} & Motion & Defocus & \begin{tabular}[c]{@{}c@{}}Frosted \\ Glass\end{tabular} & \multicolumn{1}{c|}{Gaussian} & Gaussian & Impulse & Shot & Speckle & \multicolumn{1}{c|}{Intensity} & Brightness & Contrast & Saturate & \multicolumn{1}{c|}{JPEG} & Snow & Spatter & Fog & \multicolumn{1}{c|}{Frost} & \begin{tabular}[c]{@{}c@{}}Geometric\\ Distortion\end{tabular} \\ \midrule
			\multicolumn{1}{c|}{\textbf{MobileNet-V2}} & \multicolumn{1}{c|}{33.1} & 16.1 & 16.6 & 14.9 & \multicolumn{1}{c|}{16.5} & 12.1 & 11.5 & 12.4 & 17.0 & \multicolumn{1}{c|}{24.7} & 27.2 & 14.8 & 26.5 & \multicolumn{1}{c|}{25.1} & 7.8 & 18.5 & 20.1 & \multicolumn{1}{c|}{10.7} & 28.3 \\
			\multicolumn{1}{c|}{w/o ASPP} & \multicolumn{1}{c|}{27.3} & 12.2 & 11.3 & 10.5 & \multicolumn{1}{c|}{11.6} & 9.6 & 9.6 & 9.9 & 13.3 & \multicolumn{1}{c|}{19.0} & 22.1 & 10.8 & 20.8 & \multicolumn{1}{c|}{19.5} & 5.7 & 15.6 & 14.5 & \multicolumn{1}{c|}{7.8} & 22.9 \\
			\multicolumn{1}{c|}{w/o AC} & \multicolumn{1}{c|}{32.1} & 15.2 & 15.9 & 14.2 & \multicolumn{1}{c|}{15.7} & 11.2 & 11.5 & 11.4 & 15.6 & \multicolumn{1}{c|}{23.0} & 27.1 & 13.7 & 25.1 & \multicolumn{1}{c|}{25.1} & 7.6 & 18.8 & 19.2 & \multicolumn{1}{c|}{10.7} & 27.9 \\
			\multicolumn{1}{c|}{w/ DPC} & \multicolumn{1}{c|}{\textbf{34.7}} & \textbf{17.3} & \textbf{18.9} & \textbf{15.6} & \multicolumn{1}{c|}{\textbf{18.0}} & \textbf{13.9} & \textbf{13.7} & \textbf{13.8} & \textbf{18.4} & \multicolumn{1}{c|}{\textbf{25.8}} & \textbf{28.9} & \textbf{15.8} & \textbf{27.8} & \multicolumn{1}{c|}{26.2} & 8.6 & 20.8 & \textbf{21.5} & \multicolumn{1}{c|}{11.6} & \textbf{29.4} \\
			\multicolumn{1}{c|}{w/o LRL} & \multicolumn{1}{c|}{32.2} & 15.7 & 16.5 & 14.3 & \multicolumn{1}{c|}{16.3} & 12.9 & 11.6 & 13.2 & 17.3 & \multicolumn{1}{c|}{24.4} & 26.7 & 14.2 & 25.1 & \multicolumn{1}{c|}{24.6} & 7.5 & 18.9 & 19.7 & \multicolumn{1}{c|}{10.7} & 27.6 \\
			\multicolumn{1}{c|}{w/ GAP} & \multicolumn{1}{c|}{33.9} & 17.2 & 17.9 & 15.1 & \multicolumn{1}{c|}{17.2} & 12.5 & 12.8 & 12.8 & 16.8 & \multicolumn{1}{c|}{25.3} & 28.7 & 15.0 & 27.4 & \multicolumn{1}{c|}{\textbf{26.6}} & \textbf{8.7} & \textbf{20.9} & 20.8 & \multicolumn{1}{c|}{\textbf{11.7}} & 29.1 \\ \midrule
			\multicolumn{1}{c|}{\textbf{ResNet-50}} & \multicolumn{1}{c|}{37.4} & 18.0 & 19.7 & 16.9 & \multicolumn{1}{c|}{19.2} & 14.1 & 12.8 & 14.4 & 19.4 & \multicolumn{1}{c|}{28.5} & 31.1 & 18.0 & 30.1 & \multicolumn{1}{c|}{29.5} & 8.8 & 21.5 & 23.9 & \multicolumn{1}{c|}{13.6} & 32.9 \\
			\multicolumn{1}{c|}{w/o ASPP} & \multicolumn{1}{c|}{29.7} & 13.5 & 13.8 & 11.6 & \multicolumn{1}{c|}{13.5} & 11.1 & 10.1 & 11.6 & 15.5 & \multicolumn{1}{c|}{21.6} & 24.8 & 13.3 & 23.4 & \multicolumn{1}{c|}{22.7} & 6.7 & 17.0 & 17.6 & \multicolumn{1}{c|}{9.9} & 25.5 \\
			\multicolumn{1}{c|}{w/o AC} & \multicolumn{1}{c|}{36.5} & 18.2 & 19.3 & 16.6 & \multicolumn{1}{c|}{18.7} & 13.7 & 11.8 & 13.8 & 18.8 & \multicolumn{1}{c|}{27.5} & 30.2 & 17.3 & 29.1 & \multicolumn{1}{c|}{28.7} & 7.9 & 20.4 & 23.3 & \multicolumn{1}{c|}{12.8} & 31.3 \\
			\multicolumn{1}{c|}{w/ DPC} & \multicolumn{1}{c|}{37.9} & 18.9 & 20.3 & \textbf{17.5} & \multicolumn{1}{c|}{19.8} & 13.4 & 12.2 & 13.7 & 19.2 & \multicolumn{1}{c|}{29.0} & 31.6 & 18.9 & 30.3 & \multicolumn{1}{c|}{30.1} & 8.6 & 20.9 & 24.5 & \multicolumn{1}{c|}{13.6} & 33.0 \\
			\multicolumn{1}{c|}{w/o LRL} & \multicolumn{1}{c|}{36.6} & 18.3 & 19.8 & 16.1 & \multicolumn{1}{c|}{18.8} & 13.6 & 12.2 & 13.8 & 18.9 & \multicolumn{1}{c|}{27.4} & 31.0 & 19.1 & 30.1 & \multicolumn{1}{c|}{29.3} & 8.1 & 21.2 & 24.3 & \multicolumn{1}{c|}{13.4} & 32.0 \\
			\multicolumn{1}{c|}{w/ GAP} & \multicolumn{1}{c|}{\textbf{38.2}} & \textbf{19.3} & \textbf{21.1} & 17.2 & \multicolumn{1}{c|}{\textbf{19.9}} & \textbf{15.5} & \textbf{12.8} & \textbf{15.8} & \textbf{21.3} & \multicolumn{1}{c|}{\textbf{30.4}} & \textbf{32.9} & \textbf{19.5} & \textbf{31.7} & \multicolumn{1}{c|}{\textbf{30.8}} & \textbf{9.9} & \textbf{23.2} & \textbf{25.8} & \multicolumn{1}{c|}{\textbf{14.9}} & \textbf{33.0} \\ \midrule
			\multicolumn{1}{c|}{\textbf{ResNet-101}} & \multicolumn{1}{c|}{38.1} & 19.1 & 20.6 & 17.3 & \multicolumn{1}{c|}{19.8} & 15.4 & 14.6 & 15.7 & 20.7 & \multicolumn{1}{c|}{28.8} & 31.6 & 19.7 & 31.2 & \multicolumn{1}{c|}{31.4} & 10.2 & 22.9 & 25.6 & \multicolumn{1}{c|}{14.0} & 32.8 \\
			\multicolumn{1}{c|}{w/o ASPP} & \multicolumn{1}{c|}{30.7} & 14.3 & 14.1 & 12.8 & \multicolumn{1}{c|}{14.2} & 13.3 & 11.8 & 13.7 & 17.7 & \multicolumn{1}{c|}{23.4} & 25.9 & 14.4 & 24.7 & \multicolumn{1}{c|}{24.1} & 7.3 & 18.5 & 18.8 & \multicolumn{1}{c|}{10.7} & 26.2 \\
			\multicolumn{1}{c|}{w/o AC} & \multicolumn{1}{c|}{37.3} & 18.3 & 19.9 & 16.9 & \multicolumn{1}{c|}{19.0} & 14.4 & 14.4 & 14.7 & 19.4 & \multicolumn{1}{c|}{27.5} & 31.4 & 18.1 & 30.1 & \multicolumn{1}{c|}{30.5} & 9.4 & 22.9 & 24.6 & \multicolumn{1}{c|}{13.6} & 32.2 \\
			\multicolumn{1}{c|}{w/ DPC} & \multicolumn{1}{c|}{37.6} & 19.6 & 21.0 & 17.7 & \multicolumn{1}{c|}{20.0} & 15.9 & \textbf{15.1} & 16.4 & 21.6 & \multicolumn{1}{c|}{28.7} & 32.1 & 19.5 & 31.5 & \multicolumn{1}{c|}{31.2} & 9.7 & 23.3 & 25.4 & \multicolumn{1}{c|}{14.0} & 32.6 \\
			\multicolumn{1}{c|}{w/o LRL} & \multicolumn{1}{c|}{37.5} & 18.9 & 20.5 & 17.7 & \multicolumn{1}{c|}{19.9} & 16.5 & 14.6 & \textbf{16.8} & \textbf{21.7} & \multicolumn{1}{c|}{29.0} & 31.6 & 19.8 & 30.7 & \multicolumn{1}{c|}{30.1} & 9.8 & 22.2 & 25.9 & \multicolumn{1}{c|}{14.0} & 32.2 \\
			\multicolumn{1}{c|}{w/ GAP} & \multicolumn{1}{c|}{\textbf{39.3}} & \textbf{20.2} & \textbf{21.7} & \textbf{17.9} & \multicolumn{1}{c|}{\textbf{20.6}} & \textbf{15.9} & 14.2 & 16.1 & 21.4 & \multicolumn{1}{c|}{\textbf{29.9}} & \textbf{33.2} & \textbf{20.4} & \textbf{32.8} & \multicolumn{1}{c|}{\textbf{32.8}} & \textbf{10.8} & \textbf{23.3} & \textbf{27.0} & \multicolumn{1}{c|}{\textbf{15.6}} & \textbf{34.2} \\ \midrule
			\multicolumn{1}{c|}{\textbf{Xception-41}} & \multicolumn{1}{c|}{39.7} & 22.1 & 22.7 & 17.4 & \multicolumn{1}{c|}{\textbf{20.8}} & 20.8 & 18.1 & 20.5 & 24.8 & \multicolumn{1}{c|}{\textbf{33.7}} & 34.2 & 20.9 & 32.5 & \multicolumn{1}{c|}{32.6} & 13.0 & 25.0 & 28.4 & \multicolumn{1}{c|}{17.0} & \textbf{34.4} \\
			\multicolumn{1}{c|}{w/o ASPP} & \multicolumn{1}{c|}{35.4} & 19.4 & 20.0 & 15.3 & \multicolumn{1}{c|}{18.4} & 18.2 & 16.3 & 17.9 & 21.7 & \multicolumn{1}{c|}{29.2} & 30.3 & 17.7 & 28.5 & \multicolumn{1}{c|}{28.3} & 11.2 & 22.7 & 23.5 & \multicolumn{1}{c|}{14.4} & 31.3 \\
			\multicolumn{1}{c|}{w/o AC} & \multicolumn{1}{c|}{38.4} & 21.8 & 22.2 & \textbf{17.7} & \multicolumn{1}{c|}{20.6} & 21.8 & 18.3 & 21.1 & 25.0 & \multicolumn{1}{c|}{32.9} & 33.4 & 20.0 & 31.7 & \multicolumn{1}{c|}{32.0} & 12.2 & 24.8 & 26.4 & \multicolumn{1}{c|}{15.9} & 33.0 \\
			\multicolumn{1}{c|}{w/ DPC} & \multicolumn{1}{c|}{\textbf{40.2}} & 21.9 & 22.5 & 17.4 & \multicolumn{1}{c|}{20.8} & 20.2 & 17.5 & 19.6 & 23.9 & \multicolumn{1}{c|}{33.3} & \textbf{34.8} & 20.3 & \textbf{32.6} & \multicolumn{1}{c|}{\textbf{32.9}} & 13.8 & 25.6 & 26.9 & \multicolumn{1}{c|}{17.4} & 34.5 \\
			\multicolumn{1}{c|}{w/o LRL} & \multicolumn{1}{c|}{39.1} & 21.4 & 22.6 & 17.2 & \multicolumn{1}{c|}{20.6} & 20.8 & 17.6 & 20.5 & 25.0 & \multicolumn{1}{c|}{32.6} & 34.1 & 21.1 & 32.1 & \multicolumn{1}{c|}{32.2} & 13.8 & 25.5 & 27.1 & \multicolumn{1}{c|}{16.9} & 34.2 \\
			\multicolumn{1}{c|}{w/ GAP} & \multicolumn{1}{c|}{39.0} & \textbf{22.7} & \textbf{22.9} & 17.5 & \multicolumn{1}{c|}{21.0} & \textbf{21.9} & \textbf{18.5} & \textbf{21.6} & \textbf{25.4} & \multicolumn{1}{c|}{33.1} & 34.1 & \textbf{21.6} & 32.3 & \multicolumn{1}{c|}{32.6} & \textbf{14.3} & \textbf{25.8} & \textbf{28.9} & \multicolumn{1}{c|}{\textbf{17.7}} & 34.1 \\ \midrule
			\multicolumn{1}{c|}{\textbf{Xception-65}} & \multicolumn{1}{c|}{41.4} & 23.4 & 25.2 & 18.9 & \multicolumn{1}{c|}{22.7} & 23.2 & 19.8 & 22.9 & 27.1 & \multicolumn{1}{c|}{35.4} & 36.1 & 23.5 & 34.8 & \multicolumn{1}{c|}{\textbf{34.2}} & 14.8 & 27.7 & 30.0 & \multicolumn{1}{c|}{18.4} & 35.6 \\
			\multicolumn{1}{c|}{w/o ASPP} & \multicolumn{1}{c|}{40.2} & 21.4 & 23.3 & 18.1 & \multicolumn{1}{c|}{21.6} & 20.4 & 16.7 & 20.1 & 24.7 & \multicolumn{1}{c|}{33.0} & 34.1 & 21.4 & 32.6 & \multicolumn{1}{c|}{31.4} & 12.1 & 25.3 & 27.4 & \multicolumn{1}{c|}{15.6} & 35.1 \\
			\multicolumn{1}{c|}{w/o AC} & \multicolumn{1}{c|}{40.0} & 22.7 & 24.4 & 18.6 & \multicolumn{1}{c|}{22.1} & 23.6 & 20.9 & 22.8 & 26.4 & \multicolumn{1}{c|}{34.6} & 35.1 & 23.4 & 33.9 & \multicolumn{1}{c|}{33.3} & 13.9 & 27.2 & 29.1 & \multicolumn{1}{c|}{17.8} & 35.0 \\
			\multicolumn{1}{c|}{w/ DPC} & \multicolumn{1}{c|}{40.9} & 23.7 & 24.9 & 18.4 & \multicolumn{1}{c|}{22.8} & 23.0 & 18.8 & 22.9 & 27.0 & \multicolumn{1}{c|}{35.6} & 35.7 & 23.0 & 34.1 & \multicolumn{1}{c|}{33.9} & 14.6 & 28.0 & 29.5 & \multicolumn{1}{c|}{17.9} & 35.7 \\
			\multicolumn{1}{c|}{w/o LRL} & \multicolumn{1}{c|}{41.0} & 23.2 & 25.0 & 18.6 & \multicolumn{1}{c|}{22.7} & 24.2 & 19.8 & 23.9 & 27.6 & \multicolumn{1}{c|}{35.6} & 35.7 & 23.5 & 34.1 & \multicolumn{1}{c|}{33.8} & 14.9 & 27.5 & 29.3 & \multicolumn{1}{c|}{19.0} & \textbf{36.1} \\
			\multicolumn{1}{c|}{w/ GAP} & \multicolumn{1}{c|}{\textbf{41.7}} & \textbf{23.9} & \textbf{25.6} & \textbf{19.2} & \multicolumn{1}{c|}{\textbf{23.4}} & \textbf{24.7} & \textbf{20.8} & \textbf{24.2} & \textbf{28.1} & \multicolumn{1}{c|}{\textbf{36.2}} & \textbf{36.1} & \textbf{23.8} & \textbf{35.0} & \multicolumn{1}{c|}{34.1} & \textbf{15.4} & \textbf{28.2} & \textbf{30.5} & \multicolumn{1}{c|}{\textbf{20.1}} & 36.0 \\ \midrule
			\multicolumn{1}{c|}{\textbf{Xception-71}} & \multicolumn{1}{c|}{42.4} & \textbf{24.4} & 26.4 & 19.5 & \multicolumn{1}{c|}{23.9} & \textbf{24.0} & \textbf{20.3} & \textbf{23.3} & \textbf{27.5} & \multicolumn{1}{c|}{\textbf{36.8}} & \textbf{37.2} & \textbf{25.3} & \textbf{35.7} & \multicolumn{1}{c|}{\textbf{34.7}} & 16.1 & \textbf{29.4} & 31.3 & \multicolumn{1}{c|}{19.8} & \textbf{37.1} \\
			\multicolumn{1}{c|}{w/o ASPP} & \multicolumn{1}{c|}{40.6} & 21.9 & 24.2 & 17.5 & \multicolumn{1}{c|}{21.9} & 20.8 & 16.6 & 20.0 & 24.2 & \multicolumn{1}{c|}{34.0} & 34.8 & 22.5 & 33.1 & \multicolumn{1}{c|}{32.4} & 12.9 & 26.3 & 28.9 & \multicolumn{1}{c|}{16.5} & 35.2 \\
			\multicolumn{1}{c|}{w/o AC} & \multicolumn{1}{c|}{41.8} & 24.3 & 25.4 & \textbf{19.6} & \multicolumn{1}{c|}{23.6} & 24.0 & 19.9 & 22.9 & 26.2 & \multicolumn{1}{c|}{35.7} & 36.1 & 23.2 & 34.8 & \multicolumn{1}{c|}{33.7} & 15.7 & 28.3 & 29.9 & \multicolumn{1}{c|}{19.7} & 35.8 \\
			\multicolumn{1}{c|}{w/ DPC} & \multicolumn{1}{c|}{\textbf{42.5}} & 23.3 & 25.9 & 18.4 & \multicolumn{1}{c|}{23.1} & 23.4 & 18.9 & 22.4 & 26.5 & \multicolumn{1}{c|}{36.3} & 36.5 & 24.1 & 34.9 & \multicolumn{1}{c|}{34.2} & 15.8 & 28.3 & 30.6 & \multicolumn{1}{c|}{18.6} & 36.3 \\
			\multicolumn{1}{c|}{w/o LRL} & \multicolumn{1}{c|}{42.2} & 22.9 & 25.9 & 18.7 & \multicolumn{1}{c|}{23.5} & 21.7 & 18.7 & 21.1 & 25.4 & \multicolumn{1}{c|}{35.5} & 36.3 & 24.3 & 34.5 & \multicolumn{1}{c|}{34.0} & 15.1 & 28.6 & 30.6 & \multicolumn{1}{c|}{19.7} & 36.4 \\
			\multicolumn{1}{c|}{w/ GAP} & \multicolumn{1}{c|}{42.0} & 24.0 & \textbf{26.6} & 19.1 & \multicolumn{1}{c|}{\textbf{24.0}} & 23.6 & 19.8 & 22.8 & 26.7 & \multicolumn{1}{c|}{35.9} & 37.0 & 25.0 & 35.2 & \multicolumn{1}{c|}{34.6} & \textbf{16.7} & 29.3 & \textbf{31.6} & \multicolumn{1}{c|}{\textbf{20.9}} & 36.3 \\ \bottomrule
		\end{tabular}
		
	\end{adjustbox}
	\caption{
		Mean IoU for clean and corrupted variants of the validation set of ADE20K for several network backbones of the DeepLabv3+ architecture and respective architectural ablations. 
		Every mIoU is averaged over all available severity levels, except for corruptions of category noise where only the first three severity levels are considered. 
		The standard deviation for image corruptions of category noise is $0.2$ or less.
		Highest mIoU per corruption is bold.}
	\label{tab:miou_ade_allbackbones_and_ablations}
\end{table*}

\begin{table*}[h]
	\begin{adjustbox}{width=\textwidth}
		\begin{tabular}{@{}ccccccccccccccccccc@{}}
			\toprule
			& \multicolumn{4}{c}{\textbf{Blur}} & \multicolumn{5}{c}{\textbf{Noise}} & \multicolumn{4}{c}{\textbf{Digital}} & \multicolumn{4}{c}{\textbf{Weather}} & \textbf{} \\ \midrule
			\multicolumn{1}{c|}{\begin{tabular}[c]{@{}c@{}}Deeplab-v3+ \\ Backbone\end{tabular}} & Motion & Defocus & \begin{tabular}[c]{@{}c@{}}Frosted \\ Glass\end{tabular} & \multicolumn{1}{c|}{Gaussian} & Gaussian & Impulse & Shot & Speckle & \multicolumn{1}{c|}{Intensity} & Brightness & Contrast & Saturate & \multicolumn{1}{c|}{JPEG} & Snow & Spatter & Fog & \multicolumn{1}{c|}{Frost} & \begin{tabular}[c]{@{}c@{}}Geometric\\ Distortion\end{tabular} \\ \midrule
			\multicolumn{1}{c|}{\textbf{MobileNet-V2}} & 100.0 & 100.0 & 100.0 & \multicolumn{1}{c|}{100.0} & 100.0 & 100.0 & 100.0 & 100.0 & \multicolumn{1}{c|}{100.0} & 100.0 & 100.0 & 100.0 & \multicolumn{1}{c|}{100.0} & 100.0 & 100.0 & 100.0 & \multicolumn{1}{c|}{100.0} & 100.0 \\
			\multicolumn{1}{c|}{w/o ASPP} & \textbf{104.7} & \textbf{106.4} & \textbf{105.2} & \multicolumn{1}{c|}{\textbf{105.9}} & \textbf{102.9} & \textbf{102.1} & \textbf{102.8} & \textbf{104.5} & \multicolumn{1}{c|}{\textbf{107.5}} & \textbf{107.0} & \textbf{104.8} & \textbf{107.8} & \multicolumn{1}{c|}{\textbf{107.5}} & \textbf{102.3} & \textbf{103.6} & \textbf{107.1} & \multicolumn{1}{c|}{\textbf{103.3}} & \textbf{107.5} \\
			\multicolumn{1}{c|}{w/o AC} & 101.1 & 100.9 & 100.8 & \multicolumn{1}{c|}{101.0} & 101.1 & 100.0 & 101.1 & 101.7 & \multicolumn{1}{c|}{102.1} & 100.1 & 101.3 & 102.0 & \multicolumn{1}{c|}{100.0} & 100.2 & 99.6 & 101.1 & \multicolumn{1}{c|}{100.0} & 100.6 \\
			\multicolumn{1}{c|}{w/ DPC} & 98.7 & 97.3 & 99.2 & \multicolumn{1}{c|}{98.3} & 98.0 & 97.5 & 98.4 & 98.3 & \multicolumn{1}{c|}{98.5} & 97.7 & 98.8 & 98.3 & \multicolumn{1}{c|}{98.5} & 99.2 & 97.2 & 98.2 & \multicolumn{1}{c|}{99.0} & 98.4 \\
			\multicolumn{1}{c|}{w/o LRL} & 100.6 & 100.1 & 100.8 & \multicolumn{1}{c|}{100.3} & 99.1 & 99.8 & 99.1 & 99.6 & \multicolumn{1}{c|}{100.3} & 100.8 & 100.8 & 102.0 & \multicolumn{1}{c|}{100.7} & 100.4 & 99.5 & 100.6 & \multicolumn{1}{c|}{100.0} & 101.1 \\
			\multicolumn{1}{c|}{w/ GAP} & 98.8 & 98.5 & 99.8 & \multicolumn{1}{c|}{99.3} & 99.6 & 98.5 & 99.5 & 100.2 & \multicolumn{1}{c|}{99.1} & 98.0 & 99.8 & 98.8 & \multicolumn{1}{c|}{98.0} & 99.0 & 97.0 & 99.2 & \multicolumn{1}{c|}{98.9} & 99.0 \\ \midrule
			\multicolumn{1}{c|}{\textbf{ResNet-50}} & 100.0 & 100.0 & 100.0 & \multicolumn{1}{c|}{100.0} & 100.0 & 100.0 & 100.0 & 100.0 & \multicolumn{1}{c|}{100.0} & 100.0 & 100.0 & 100.0 & \multicolumn{1}{c|}{100.0} & 100.0 & 100.0 & 100.0 & \multicolumn{1}{c|}{100.0} & 100.0 \\
			\multicolumn{1}{c|}{w/o ASPP} & \textbf{105.4} & \textbf{107.4} & \textbf{106.4} & \multicolumn{1}{c|}{\textbf{107.1}} & \textbf{103.5} & \textbf{103.1} & \textbf{103.3} & \textbf{104.8} & \multicolumn{1}{c|}{\textbf{109.7}} & \textbf{109.2} & \textbf{105.8} & \textbf{109.5} & \multicolumn{1}{c|}{\textbf{109.6}} & \textbf{102.2} & \textbf{105.8} & \textbf{108.3} & \multicolumn{1}{c|}{\textbf{104.3}} & \textbf{111.0} \\
			\multicolumn{1}{c|}{w/o AC} & 99.7 & 100.6 & 100.5 & \multicolumn{1}{c|}{100.6} & 100.5 & 101.1 & 100.8 & 100.6 & \multicolumn{1}{c|}{101.4} & 101.3 & 100.9 & 101.4 & \multicolumn{1}{c|}{101.2} & 101.0 & 101.5 & 100.8 & \multicolumn{1}{c|}{100.9} & 102.3 \\
			\multicolumn{1}{c|}{w/ DPC} & 98.9 & 99.3 & 99.3 & \multicolumn{1}{c|}{99.4} & 100.8 & 100.7 & 100.9 & 100.2 & \multicolumn{1}{c|}{99.3} & 99.3 & 98.9 & 99.7 & \multicolumn{1}{c|}{99.2} & 100.1 & 100.8 & 99.2 & \multicolumn{1}{c|}{100.0} & 99.8 \\
			\multicolumn{1}{c|}{w/o LRL} & 99.6 & 99.9 & 101.0 & \multicolumn{1}{c|}{100.6} & 100.6 & 100.7 & 100.7 & 100.5 & \multicolumn{1}{c|}{101.6} & 100.1 & 98.7 & 99.9 & \multicolumn{1}{c|}{100.3} & 100.7 & 100.4 & 99.5 & \multicolumn{1}{c|}{100.3} & 101.3 \\
			\multicolumn{1}{c|}{w/ GAP} & 98.4 & 98.3 & 99.7 & \multicolumn{1}{c|}{99.2} & 98.4 & 100.0 & 98.4 & 97.7 & \multicolumn{1}{c|}{97.4} & 97.4 & 98.1 & 97.7 & \multicolumn{1}{c|}{98.1} & 98.7 & 97.9 & 97.5 & \multicolumn{1}{c|}{98.5} & 99.7 \\ \midrule
			\multicolumn{1}{c|}{\textbf{ResNet-101}} & 100.0 & 100.0 & 100.0 & \multicolumn{1}{c|}{100.0} & 100.0 & 100.0 & 100.0 & 100.0 & \multicolumn{1}{c|}{100.0} & 100.0 & 100.0 & 100.0 & \multicolumn{1}{c|}{100.0} & 100.0 & 100.0 & 100.0 & \multicolumn{1}{c|}{100.0} & 100.0 \\
			\multicolumn{1}{c|}{w/o ASPP} & \textbf{105.9} & \textbf{108.1} & \textbf{105.5} & \multicolumn{1}{c|}{\textbf{107.0}} & \textbf{102.5} & \textbf{103.2} & \textbf{102.3} & \textbf{103.8} & \multicolumn{1}{c|}{\textbf{107.5}} & \textbf{108.4} & \textbf{106.5} & \textbf{109.4} & \multicolumn{1}{c|}{\textbf{110.6}} & \textbf{103.2} & \textbf{105.7} & \textbf{109.2} & \multicolumn{1}{c|}{\textbf{103.9}} & \textbf{109.9} \\
			\multicolumn{1}{c|}{w/o AC} & 101.0 & 100.8 & 100.5 & \multicolumn{1}{c|}{101.0} & 101.2 & 100.2 & 101.2 & 101.6 & \multicolumn{1}{c|}{101.8} & 100.3 & 101.9 & 101.5 & \multicolumn{1}{c|}{101.3} & 100.9 & 100.0 & 101.4 & \multicolumn{1}{c|}{100.5} & 101.0 \\
			\multicolumn{1}{c|}{w/ DPC} & 99.3 & 99.4 & 99.6 & \multicolumn{1}{c|}{99.7} & 99.4 & 99.3 & 99.1 & 98.9 & \multicolumn{1}{c|}{100.1} & 99.3 & 100.3 & 99.6 & \multicolumn{1}{c|}{100.4} & 100.6 & 99.5 & 100.3 & \multicolumn{1}{c|}{100.1} & 100.3 \\
			\multicolumn{1}{c|}{w/o LRL} & 100.2 & 100.1 & 99.6 & \multicolumn{1}{c|}{99.9} & 98.7 & 100.0 & 98.7 & 98.8 & \multicolumn{1}{c|}{99.6} & 100.0 & 99.9 & 100.7 & \multicolumn{1}{c|}{101.9} & 100.5 & 100.9 & 99.6 & \multicolumn{1}{c|}{100.0} & 101.0 \\
			\multicolumn{1}{c|}{w/ GAP} & 98.7 & 98.6 & 99.3 & \multicolumn{1}{c|}{98.9} & 99.5 & 100.4 & 99.5 & 99.2 & \multicolumn{1}{c|}{98.3} & 97.6 & 99.1 & 97.6 & \multicolumn{1}{c|}{97.9} & 99.4 & 99.5 & 98.1 & \multicolumn{1}{c|}{98.1} & 98.0 \\ \midrule
			\multicolumn{1}{c|}{\textbf{Xception-41}} & 100.0 & 100.0 & 100.0 & \multicolumn{1}{c|}{100.0} & 100.0 & 100.0 & 100.0 & 100.0 & \multicolumn{1}{c|}{100.0} & 100.0 & 100.0 & 100.0 & \multicolumn{1}{c|}{100.0} & 100.0 & 100.0 & 100.0 & \multicolumn{1}{c|}{100.0} & 100.0 \\
			\multicolumn{1}{c|}{w/o ASPP} & \textbf{103.4} & \textbf{103.4} & \textbf{102.5} & \multicolumn{1}{c|}{\textbf{103.0}} & \textbf{103.4} & \textbf{102.3} & \textbf{103.3} & \textbf{104.1} & \multicolumn{1}{c|}{\textbf{106.9}} & \textbf{105.9} & \textbf{104.0} & \textbf{105.9} & \multicolumn{1}{c|}{\textbf{106.4}} & \textbf{102.1} & \textbf{103.1} & \textbf{106.9} & \multicolumn{1}{c|}{\textbf{103.1}} & \textbf{104.7} \\
			\multicolumn{1}{c|}{w/o AC} & 100.3 & 100.6 & 99.6 & \multicolumn{1}{c|}{100.3} & 98.7 & 99.8 & 99.2 & 99.8 & \multicolumn{1}{c|}{101.3} & 101.2 & 101.1 & 101.1 & \multicolumn{1}{c|}{100.9} & 100.9 & 100.3 & 102.8 & \multicolumn{1}{c|}{101.2} & 102.1 \\
			\multicolumn{1}{c|}{w/ DPC} & 100.3 & 100.2 & 99.9 & \multicolumn{1}{c|}{100.0} & 100.8 & 100.8 & 101.2 & 101.2 & \multicolumn{1}{c|}{100.7} & 99.1 & 100.8 & 99.8 & \multicolumn{1}{c|}{99.6} & 99.1 & 99.3 & 102.2 & \multicolumn{1}{c|}{99.4} & 99.9 \\
			\multicolumn{1}{c|}{w/o LRL} & 100.8 & 100.1 & 100.2 & \multicolumn{1}{c|}{100.2} & 100.0 & 100.7 & 100.0 & 99.7 & \multicolumn{1}{c|}{101.7} & 100.1 & 99.8 & 100.6 & \multicolumn{1}{c|}{100.6} & 99.1 & 99.4 & 101.8 & \multicolumn{1}{c|}{100.0} & 100.3 \\
			\multicolumn{1}{c|}{w/ GAP} & 99.2 & 99.6 & 99.9 & \multicolumn{1}{c|}{99.7} & 98.6 & 99.5 & 98.5 & 99.2 & \multicolumn{1}{c|}{101.0} & 100.1 & 99.1 & 100.3 & \multicolumn{1}{c|}{100.0} & 98.5 & 99.0 & 99.4 & \multicolumn{1}{c|}{99.2} & 100.4 \\ \midrule
			\multicolumn{1}{c|}{\textbf{Xception-65}} & 100.0 & 100.0 & 100.0 & \multicolumn{1}{c|}{100.0} & 100.0 & 100.0 & 100.0 & 100.0 & \multicolumn{1}{c|}{100.0} & 100.0 & 100.0 & 100.0 & \multicolumn{1}{c|}{100.0} & 100.0 & 100.0 & 100.0 & \multicolumn{1}{c|}{100.0} & 100.0 \\
			\multicolumn{1}{c|}{w/o ASPP} & \textbf{102.7} & \textbf{102.6} & \textbf{101.0} & \multicolumn{1}{c|}{\textbf{101.5}} & \textbf{103.5} & \textbf{104.0} & \textbf{103.7} & \textbf{103.2} & \multicolumn{1}{c|}{\textbf{103.6}} & \textbf{103.2} & \textbf{102.8} & \textbf{103.4} & \multicolumn{1}{c|}{\textbf{104.2}} & \textbf{103.2} & \textbf{103.3} & \textbf{103.8} & \multicolumn{1}{c|}{\textbf{103.5}} & 100.8 \\
			\multicolumn{1}{c|}{w/o AC} & 100.9 & 101.0 & 100.4 & \multicolumn{1}{c|}{100.8} & 99.4 & 98.7 & 100.1 & 100.9 & \multicolumn{1}{c|}{101.1} & 101.6 & 100.2 & 101.3 & \multicolumn{1}{c|}{101.2} & 101.0 & 100.7 & 101.3 & \multicolumn{1}{c|}{100.7} & \textbf{100.9} \\
			\multicolumn{1}{c|}{w/ DPC} & 99.6 & 100.3 & 100.7 & \multicolumn{1}{c|}{99.9} & 100.2 & 101.3 & 100.1 & 100.2 & \multicolumn{1}{c|}{99.6} & 100.6 & 100.6 & 101.1 & \multicolumn{1}{c|}{100.4} & 100.2 & 99.6 & 100.7 & \multicolumn{1}{c|}{100.6} & 99.9 \\
			\multicolumn{1}{c|}{w/o LRL} & 100.3 & 100.2 & 100.4 & \multicolumn{1}{c|}{100.0} & 98.6 & 100.1 & 98.8 & 99.2 & \multicolumn{1}{c|}{99.6} & 100.5 & 100.1 & 101.1 & \multicolumn{1}{c|}{100.6} & 99.9 & 100.2 & 101.0 & \multicolumn{1}{c|}{99.3} & 99.3 \\
			\multicolumn{1}{c|}{w/ GAP} & 99.4 & 99.4 & 99.7 & \multicolumn{1}{c|}{99.2} & 98.0 & 98.8 & 98.3 & 98.6 & \multicolumn{1}{c|}{98.7} & 99.9 & 99.6 & 99.7 & \multicolumn{1}{c|}{100.1} & 99.3 & 99.3 & 99.3 & \multicolumn{1}{c|}{97.9} & 99.5 \\ \midrule
			\multicolumn{1}{c|}{\textbf{Xception-71}} & 100.0 & 100.0 & 100.0 & \multicolumn{1}{c|}{100.0} & 100.0 & 100.0 & 100.0 & 100.0 & \multicolumn{1}{c|}{100.0} & 100.0 & 100.0 & 100.0 & \multicolumn{1}{c|}{100.0} & 100.0 & 100.0 & 100.0 & \multicolumn{1}{c|}{100.0} & 100.0 \\
			\multicolumn{1}{c|}{w/o ASPP} & \textbf{103.3} & \textbf{103.1} & \textbf{102.6} & \multicolumn{1}{c|}{\textbf{102.6}} & \textbf{104.3} & \textbf{104.6} & \textbf{104.3} & \textbf{104.6} & \multicolumn{1}{c|}{\textbf{104.5}} & \textbf{103.8} & \textbf{103.8} & \textbf{103.9} & \multicolumn{1}{c|}{\textbf{103.6}} & \textbf{103.8} & \textbf{104.4} & \textbf{103.6} & \multicolumn{1}{c|}{\textbf{104.1}} & \textbf{103.0} \\
			\multicolumn{1}{c|}{w/o AC} & 100.2 & 101.4 & 100.0 & \multicolumn{1}{c|}{100.4} & 100.0 & 100.5 & 100.6 & 101.8 & \multicolumn{1}{c|}{101.8} & 101.7 & 102.8 & 101.3 & \multicolumn{1}{c|}{101.5} & 100.5 & 101.5 & 102.1 & \multicolumn{1}{c|}{100.1} & 102.1 \\
			\multicolumn{1}{c|}{w/ DPC} & 101.4 & 100.8 & 101.4 & \multicolumn{1}{c|}{101.0} & 100.7 & 101.8 & 101.2 & 101.4 & \multicolumn{1}{c|}{100.9} & 101.1 & 101.7 & 101.3 & \multicolumn{1}{c|}{100.7} & 100.4 & 101.5 & 101.1 & \multicolumn{1}{c|}{101.5} & 101.3 \\
			\multicolumn{1}{c|}{w/o LRL} & 101.9 & 100.7 & 101.0 & \multicolumn{1}{c|}{100.6} & 103.0 & 102.0 & 103.0 & 102.9 & \multicolumn{1}{c|}{102.1} & 101.3 & 101.5 & 101.9 & \multicolumn{1}{c|}{101.1} & 101.3 & 101.0 & 101.0 & \multicolumn{1}{c|}{100.2} & 101.1 \\
			\multicolumn{1}{c|}{w/ GAP} & 100.5 & 99.7 & 100.5 & \multicolumn{1}{c|}{99.9} & 100.5 & 100.6 & 100.8 & 101.1 & \multicolumn{1}{c|}{101.5} & 100.2 & 100.4 & 100.8 & \multicolumn{1}{c|}{100.1} & 99.3 & 100.1 & 99.6 & \multicolumn{1}{c|}{98.6} & 101.3 \\ \bottomrule
		\end{tabular}
		
	\end{adjustbox}
	\caption{
		CD for corrupted variants of the validation set of ADE20K for several network backbones of the DeepLabv3+ architecture and respective architectural ablations. 
		Highest CD per corruption is bold.}
	\label{tab:cd_ade_allbackbones_and_ablations}
\end{table*}

\begin{table*}[h]
	\begin{adjustbox}{width=\textwidth}
		\begin{tabular}{@{}ccccccccccccccccccc@{}}
			\toprule
			& \multicolumn{4}{c}{\textbf{Blur}} & \multicolumn{5}{c}{\textbf{Noise}} & \multicolumn{4}{c}{\textbf{Digital}} & \multicolumn{4}{c}{\textbf{Weather}} & \textbf{} \\ \midrule
			\multicolumn{1}{c|}{\begin{tabular}[c]{@{}c@{}}Deeplab-v3+ \\ Backbone\end{tabular}} & Motion & Defocus & \begin{tabular}[c]{@{}c@{}}Frosted \\ Glass\end{tabular} & \multicolumn{1}{c|}{Gaussian} & Gaussian & Impulse & Shot & Speckle & \multicolumn{1}{c|}{Intensity} & Brightness & Contrast & Saturate & \multicolumn{1}{c|}{JPEG} & Snow & Spatter & Fog & \multicolumn{1}{c|}{Frost} & \begin{tabular}[c]{@{}c@{}}Geometric\\ Distortion\end{tabular} \\ \midrule
			\multicolumn{1}{c|}{\textbf{MobileNet-V2}} & 100.0 & \textbf{100.0} & 100.0 & \multicolumn{1}{c|}{100.0} & 100.0 & \textbf{100.0} & 100.0 & 100.0 & \multicolumn{1}{c|}{100.0} & \textbf{100.0} & 100.0 & 100.0 & \multicolumn{1}{c|}{\textbf{100.0}} & \textbf{100.0} & \textbf{100.0} & 100.0 & \multicolumn{1}{c|}{\textbf{100.0}} & 100.0 \\
			\multicolumn{1}{c|}{w/o ASPP} & 82.3 & 88.6 & 85.4 & \multicolumn{1}{c|}{86.6} & 79.5 & 77.4 & 78.9 & 80.2 & \multicolumn{1}{c|}{83.8} & 74.0 & 83.8 & 81.9 & \multicolumn{1}{c|}{82.9} & 81.0 & 74.5 & 88.0 & \multicolumn{1}{c|}{82.0} & 74.2 \\
			\multicolumn{1}{c|}{w/o AC} & \textbf{100.1} & 99.4 & 98.9 & \multicolumn{1}{c|}{99.6} & 100.0 & 96.5 & 100.2 & 102.4 & \multicolumn{1}{c|}{\textbf{105.5}} & 91.8 & 100.9 & 105.0 & \multicolumn{1}{c|}{92.8} & 97.7 & 93.3 & 100.0 & \multicolumn{1}{c|}{96.7} & 95.0 \\
			\multicolumn{1}{c|}{w/ DPC} & 97.2 & 91.7 & 99.2 & \multicolumn{1}{c|}{95.5} & 95.1 & 93.4 & 96.4 & 95.5 & \multicolumn{1}{c|}{94.9} & 88.5 & 97.8 & 93.2 & \multicolumn{1}{c|}{95.0} & 99.2 & 90.7 & 94.8 & \multicolumn{1}{c|}{98.5} & 93.2 \\
			\multicolumn{1}{c|}{w/o LRL} & 98.5 & 96.7 & 99.4 & \multicolumn{1}{c|}{97.5} & 93.6 & 96.3 & 93.8 & 94.5 & \multicolumn{1}{c|}{95.7} & 97.7 & 99.3 & \textbf{106.0} & \multicolumn{1}{c|}{97.7} & 98.6 & 93.8 & 98.1 & \multicolumn{1}{c|}{97.1} & 99.6 \\
			\multicolumn{1}{c|}{w/ GAP} & 99.0 & 97.8 & \textbf{102.8} & \multicolumn{1}{c|}{\textbf{100.9}} & \textbf{101.7} & 97.9 & \textbf{101.7} & \textbf{105.0} & \multicolumn{1}{c|}{101.4} & 94.0 & \textbf{102.8} & 99.3 & \multicolumn{1}{c|}{94.6} & 99.8 & 91.4 & \textbf{101.0} & \multicolumn{1}{c|}{99.3} & \textbf{100.9} \\ \midrule
			\multicolumn{1}{c|}{\textbf{ResNet-50}} & \textbf{100.0} & \textbf{100.0} & 100.0 & \multicolumn{1}{c|}{100.0} & 100.0 & 100.0 & 100.0 & 100.0 & \multicolumn{1}{c|}{100.0} & 100.0 & \textbf{100.0} & 100.0 & \multicolumn{1}{c|}{\textbf{100.0}} & 100.0 & 100.0 & \textbf{100.0} & \multicolumn{1}{c|}{100.0} & 100.0 \\
			\multicolumn{1}{c|}{w/o ASPP} & 83.1 & 89.8 & 88.1 & \multicolumn{1}{c|}{89.0} & 79.6 & 79.7 & 78.6 & 78.6 & \multicolumn{1}{c|}{90.9} & 77.6 & 84.5 & 85.4 & \multicolumn{1}{c|}{87.9} & 80.2 & 79.9 & 89.8 & \multicolumn{1}{c|}{83.1} & 92.4 \\
			\multicolumn{1}{c|}{w/o AC} & 94.1 & 97.4 & 97.4 & \multicolumn{1}{c|}{97.8} & 98.0 & 100.1 & 98.9 & 97.8 & \multicolumn{1}{c|}{101.1} & 100.0 & 99.0 & 100.4 & \multicolumn{1}{c|}{99.0} & 99.9 & 101.6 & 98.0 & \multicolumn{1}{c|}{99.6} & 114.1 \\
			\multicolumn{1}{c|}{w/ DPC} & 98.0 & 99.6 & 99.8 & \multicolumn{1}{c|}{100.0} & \textbf{105.0} & \textbf{104.5} & \textbf{105.5} & \textbf{103.9} & \multicolumn{1}{c|}{99.8} & \textbf{100.5} & 98.1 & \textbf{103.8} & \multicolumn{1}{c|}{98.9} & \textbf{102.3} & \textbf{107.0} & 99.4 & \multicolumn{1}{c|}{\textbf{102.1}} & 108.3 \\
			\multicolumn{1}{c|}{w/o LRL} & 94.3 & 95.1 & 100.0 & \multicolumn{1}{c|}{98.4} & 98.8 & 99.4 & 99.2 & 98.0 & \multicolumn{1}{c|}{\textbf{103.8}} & 88.5 & 90.4 & 88.3 & \multicolumn{1}{c|}{92.4} & 99.6 & 97.1 & 91.4 & \multicolumn{1}{c|}{97.6} & 102.4 \\
			\multicolumn{1}{c|}{w/ GAP} & 97.3 & 97.1 & \textbf{102.9} & \multicolumn{1}{c|}{\textbf{100.8}} & 97.6 & 103.2 & 97.6 & 94.1 & \multicolumn{1}{c|}{88.1} & 84.4 & 96.4 & 89.5 & \multicolumn{1}{c|}{93.5} & 98.9 & 94.7 & 91.8 & \multicolumn{1}{c|}{98.0} & \textbf{113.9} \\ \midrule
			\multicolumn{1}{c|}{\textbf{ResNet-101}} & 100.0 & 100.0 & 100.0 & \multicolumn{1}{c|}{100.0} & 100.0 & 100.0 & 100.0 & 100.0 & \multicolumn{1}{c|}{100.0} & \textbf{100.0} & 100.0 & 100.0 & \multicolumn{1}{c|}{100.0} & 100.0 & 100.0 & 100.0 & \multicolumn{1}{c|}{\textbf{100.0}} & 100.0 \\
			\multicolumn{1}{c|}{w/o ASPP} & 86.3 & 94.7 & 86.3 & \multicolumn{1}{c|}{90.3} & 76.9 & 80.3 & 75.8 & 74.7 & \multicolumn{1}{c|}{77.9} & 74.6 & 88.3 & 86.3 & \multicolumn{1}{c|}{98.5} & 83.9 & 80.6 & 95.7 & \multicolumn{1}{c|}{83.2} & 86.1 \\
			\multicolumn{1}{c|}{w/o AC} & 99.9 & 99.1 & 98.2 & \multicolumn{1}{c|}{99.8} & 100.8 & 97.1 & 100.8 & 102.7 & \multicolumn{1}{c|}{\textbf{104.5}} & 90.1 & \textbf{103.8} & \textbf{102.8} & \multicolumn{1}{c|}{\textbf{101.3}} & 99.9 & 94.4 & \textbf{101.4} & \multicolumn{1}{c|}{98.2} & 96.8 \\
			\multicolumn{1}{c|}{w/ DPC} & 94.8 & 94.8 & 96.0 & \multicolumn{1}{c|}{96.3} & 95.7 & 95.6 & 94.7 & 92.4 & \multicolumn{1}{c|}{96.1} & 85.5 & 98.6 & 89.1 & \multicolumn{1}{c|}{96.8} & 100.1 & 94.2 & 97.8 & \multicolumn{1}{c|}{98.3} & 95.4 \\
			\multicolumn{1}{c|}{w/o LRL} & 98.0 & 97.1 & 95.6 & \multicolumn{1}{c|}{96.4} & 92.6 & 97.6 & 92.6 & 91.3 & \multicolumn{1}{c|}{90.8} & 91.2 & 96.5 & 98.6 & \multicolumn{1}{c|}{110.6} & 99.6 & 101.1 & 93.0 & \multicolumn{1}{c|}{97.6} & \textbf{101.6} \\
			\multicolumn{1}{c|}{w/ GAP} & \textbf{100.5} & \textbf{100.0} & \textbf{102.6} & \multicolumn{1}{c|}{\textbf{101.6}} & \textbf{103.1} & \textbf{106.2} & \textbf{103.2} & \textbf{102.9} & \multicolumn{1}{c|}{99.6} & 92.8 & 102.3 & 92.9 & \multicolumn{1}{c|}{95.6} & \textbf{102.3} & \textbf{104.8} & 98.2 & \multicolumn{1}{c|}{98.1} & 96.4 \\ \midrule
			\multicolumn{1}{c|}{\textbf{Xception-41}} & 100.0 & 100.0 & 100.0 & \multicolumn{1}{c|}{100.0} & 100.0 & 100.0 & 100.0 & 100.0 & \multicolumn{1}{c|}{100.0} & \textbf{100.0} & 100.0 & 100.0 & \multicolumn{1}{c|}{100.0} & \textbf{100.0} & \textbf{100.0} & 100.0 & \multicolumn{1}{c|}{100.0} & 100.0 \\
			\multicolumn{1}{c|}{w/o ASPP} & 90.8 & 90.3 & 90.2 & \multicolumn{1}{c|}{89.8} & 91.4 & 88.7 & 91.1 & 91.9 & \multicolumn{1}{c|}{104.6} & 92.7 & 93.8 & 95.5 & \multicolumn{1}{c|}{100.7} & 90.7 & 86.6 & 105.6 & \multicolumn{1}{c|}{92.3} & 77.0 \\
			\multicolumn{1}{c|}{w/o AC} & 94.0 & 95.1 & 92.5 & \multicolumn{1}{c|}{94.1} & 87.8 & 92.9 & 89.7 & 90.1 & \multicolumn{1}{c|}{92.1} & 90.7 & 97.7 & 92.2 & \multicolumn{1}{c|}{90.2} & 98.0 & 92.5 & 106.1 & \multicolumn{1}{c|}{98.7} & 101.4 \\
			\multicolumn{1}{c|}{w/ DPC} & \textbf{104.3} & \textbf{104.2} & \textbf{102.2} & \multicolumn{1}{c|}{\textbf{102.6}} & \textbf{106.1} & \textbf{105.4} & \textbf{107.6} & \textbf{109.5} & \multicolumn{1}{c|}{\textbf{116.2}} & 98.5 & \textbf{106.0} & \textbf{105.5} & \multicolumn{1}{c|}{\textbf{103.4}} & 99.0 & 100.1 & \textbf{118.5} & \multicolumn{1}{c|}{\textbf{100.2}} & \textbf{108.8} \\
			\multicolumn{1}{c|}{w/o LRL} & 100.0 & 96.9 & 98.0 & \multicolumn{1}{c|}{97.4} & 96.5 & 99.5 & 96.5 & 94.2 & \multicolumn{1}{c|}{107.8} & 89.9 & 95.6 & 97.0 & \multicolumn{1}{c|}{97.1} & 94.7 & 92.8 & 105.9 & \multicolumn{1}{c|}{97.3} & 91.7 \\
			\multicolumn{1}{c|}{w/ GAP} & 92.8 & 94.4 & 96.6 & \multicolumn{1}{c|}{95.1} & 90.7 & 95.1 & 90.4 & 91.3 & \multicolumn{1}{c|}{100.1} & 88.6 & 92.6 & 93.8 & \multicolumn{1}{c|}{90.9} & 92.5 & 90.3 & 90.2 & \multicolumn{1}{c|}{94.0} & 92.7 \\ \midrule
			\multicolumn{1}{c|}{\textbf{Xception-65}} & 100.0 & 100.0 & \textbf{100.0} & \multicolumn{1}{c|}{\textbf{100.0}} & 100.0 & 100.0 & 100.0 & 100.0 & \multicolumn{1}{c|}{100.0} & 100.0 & 100.0 & 100.0 & \multicolumn{1}{c|}{100.0} & 100.0 & 100.0 & 100.0 & \multicolumn{1}{c|}{100.0} & \textbf{100.0} \\
			\multicolumn{1}{c|}{w/o ASPP} & \textbf{104.3} & \textbf{103.8} & 97.9 & \multicolumn{1}{c|}{99.3} & \textbf{107.8} & \textbf{108.8} & \textbf{108.5} & \textbf{107.5} & \multicolumn{1}{c|}{\textbf{117.5}} & \textbf{113.6} & \textbf{104.9} & \textbf{114.0} & \multicolumn{1}{c|}{\textbf{120.2}} & \textbf{105.3} & \textbf{108.0} & \textbf{112.2} & \multicolumn{1}{c|}{\textbf{106.6}} & 86.3 \\
			\multicolumn{1}{c|}{w/o AC} & 96.0 & 95.8 & 95.1 & \multicolumn{1}{c|}{95.8} & 89.8 & 88.7 & 92.8 & 94.7 & \multicolumn{1}{c|}{88.5} & 92.3 & 92.8 & 91.9 & \multicolumn{1}{c|}{91.8} & 98.0 & 93.3 & 95.6 & \multicolumn{1}{c|}{96.3} & 86.2 \\
			\multicolumn{1}{c|}{w/ DPC} & 95.4 & 98.3 & 100.0 & \multicolumn{1}{c|}{96.9} & 97.9 & 102.6 & 97.6 & 97.1 & \multicolumn{1}{c|}{86.7} & 97.4 & 99.8 & 102.8 & \multicolumn{1}{c|}{96.7} & 98.7 & 94.1 & 100.0 & \multicolumn{1}{c|}{99.8} & 90.3 \\
			\multicolumn{1}{c|}{w/o LRL} & 98.6 & 97.9 & 99.2 & \multicolumn{1}{c|}{97.5} & 91.4 & 98.0 & 92.2 & 92.6 & \multicolumn{1}{c|}{88.2} & 97.0 & 97.6 & 103.0 & \multicolumn{1}{c|}{98.5} & 97.9 & 97.5 & 101.6 & \multicolumn{1}{c|}{95.4} & 83.7 \\
			\multicolumn{1}{c|}{w/ GAP} & 98.6 & 98.6 & 99.7 & \multicolumn{1}{c|}{97.7} & 92.7 & 96.6 & 94.2 & 94.4 & \multicolumn{1}{c|}{89.7} & 102.8 & 99.6 & 100.4 & \multicolumn{1}{c|}{104.0} & 98.7 & 97.9 & 97.6 & \multicolumn{1}{c|}{93.5} & 97.7 \\ \midrule
			\multicolumn{1}{c|}{\textbf{Xception-71}} & 100.0 & 100.0 & 100.0 & \multicolumn{1}{c|}{100.0} & 100.0 & 100.0 & 100.0 & 100.0 & \multicolumn{1}{c|}{100.0} & 100.0 & 100.0 & 100.0 & \multicolumn{1}{c|}{100.0} & 100.0 & 100.0 & 100.0 & \multicolumn{1}{c|}{100.0} & 100.0 \\
			\multicolumn{1}{c|}{w/o ASPP} & 103.6 & 102.7 & 101.1 & \multicolumn{1}{c|}{100.8} & 107.6 & \textbf{108.3} & 107.8 & 109.9 & \multicolumn{1}{c|}{118.3} & \textbf{110.7} & 106.0 & 110.5 & \multicolumn{1}{c|}{\textbf{106.5}} & \textbf{105.3} & \textbf{109.8} & 105.8 & \multicolumn{1}{c|}{\textbf{106.6}} & 101.7 \\
			\multicolumn{1}{c|}{w/o AC} & 97.1 & 102.1 & 97.0 & \multicolumn{1}{c|}{98.0} & 96.2 & 98.7 & 98.9 & 104.1 & \multicolumn{1}{c|}{108.0} & 107.2 & \textbf{108.3} & 102.6 & \multicolumn{1}{c|}{103.7} & 99.2 & 103.0 & 107.0 & \multicolumn{1}{c|}{97.3} & 112.2 \\
			\multicolumn{1}{c|}{w/ DPC} & 106.0 & \textbf{103.9} & \textbf{105.1} & \multicolumn{1}{c|}{\textbf{104.4}} & 103.2 & 106.7 & 105.1 & 107.0 & \multicolumn{1}{c|}{110.6} & 113.7 & 107.4 & 112.5 & \multicolumn{1}{c|}{106.1} & 101.4 & 108.2 & \textbf{107.2} & \multicolumn{1}{c|}{105.5} & \textbf{115.9} \\
			\multicolumn{1}{c|}{w/o LRL} & \textbf{106.5} & 101.7 & 102.3 & \multicolumn{1}{c|}{101.0} & \textbf{110.7} & 105.9 & \textbf{110.4} & \textbf{112.4} & \multicolumn{1}{c|}{\textbf{118.5}} & 110.3 & 104.8 & \textbf{114.1} & \multicolumn{1}{c|}{105.5} & 103.0 & 103.5 & 103.7 & \multicolumn{1}{c|}{99.4} & 107.3 \\
			\multicolumn{1}{c|}{w/ GAP} & 99.7 & 96.1 & 100.0 & \multicolumn{1}{c|}{97.3} & 100.0 & 100.3 & 100.9 & 102.3 & \multicolumn{1}{c|}{109.8} & 94.5 & 99.5 & 101.6 & \multicolumn{1}{c|}{95.4} & 96.1 & 97.3 & 93.8 & \multicolumn{1}{c|}{93.2} & 107.4 \\ \bottomrule
		\end{tabular}
	\end{adjustbox}
	\caption{
		Relative CD for corrupted variants of the validation set of ADE20K for several network backbones of the DeepLabv3+ architecture and respective architectural ablations. 
		Highest rCD per corruption is bold.}
	\label{tab:rcd_ade_allbackbones_and_ablations}
\end{table*}

\end{document}